\documentclass[twoside,11pt]{article}
\RequirePackage{luatex85}

\usepackage[natbibapa]{apacite} 
\usepackage{jair, rawfonts}

\ShortHeadings{A Review of Pseudo-Labeling for Computer Vision}
{Kage, Rothenberger, Andreadis, \& Diochnos}

\usepackage{xspace}
\usepackage{graphicx}
\usepackage[usenames,dvipsnames]{xcolor} 
\definecolor{RoyalBlue}{cmyk}{1, 0.50, 0, 0}
\definecolor{BeamerBlue}{rgb}{0.2, 0.202, 0.698}
\definecolor{ForestGreen}{cmyk}{0.864, 0.0, 0.429, 0.396}
\definecolor{DarkGreen}{rgb}{0.076, 0.379, 0.306}
\definecolor{Brown}{cmyk}{0.0,0.692,0.925,0.529}
\definecolor{MyGreen}{cmyk}{0.95, 0.05, 0.95, 0.05}
\definecolor{MyStylishGreen}{rgb}{0.328,0.601,0.169}
\definecolor{LightPurple}{rgb}{0.550,0.394,0.664}
\definecolor{DarkPurple}{rgb}{0.433,0.172,0.569}
\definecolor{MyPurple1}{rgb}{0.45,0.353,0.963}
\definecolor{MyPurple2}{rgb}{0.63,0.4,0.63}
\definecolor{Orange}{rgb}{0.93,0.478,0.121}
\definecolor{MyYellow}{rgb}{0.901,0.547,0.0}
\newif\ifcomments
\commentstrue

\usepackage[utf8]{inputenc} 
\usepackage[T1]{fontenc}    
\usepackage[
   colorlinks%
   ,plainpages=false
   ,hypertexnames=true
   ,naturalnames
   ,hyperindex
   ,backref=page
   ,citecolor=RoyalBlue
   ,urlcolor=blue
   ,pdfauthor={Kage, Rothenberger, Andreadis, Diochnos}
   ,pdftitle={A Review of Pseudo Labeling for Semi-Supervised Learning in Computer Vision}
   ,pdfsubject={Survey}
]{hyperref}
\usepackage{url}            
\usepackage{booktabs}       
\usepackage{amsfonts}       
\usepackage{nicefrac}       
\usepackage{microtype}      
\usepackage{lipsum}
\usepackage{amsmath}
\usepackage{bm}             
\usepackage{bbm}            
\usepackage{graphicx}
\usepackage{xcolor}
\usepackage{framed}
\usepackage{tikz}
\usepackage{comment}
\usepackage{makecell}
\usepackage{array, tabularx, caption, boldline}
\usepackage{xltabular}
\usepackage{cellspace}
\usepackage{upgreek}

\usetikzlibrary{arrows,arrows.meta,positioning,fit,decorations.pathmorphing,shapes.geometric}

\usepackage{subfigure}
\usepackage{mathtools}
\usepackage{amsmath,amsthm}
\newcounter{myeg}
\newtheorem{example}[myeg]{Example}
\newcounter{myrem}
\newtheorem{remark}[myrem]{Remark}

\newcommand{\abs}[1]{\ensuremath{\left\lvert#1\right\rvert}}

\newcommand{\set}[1]{\ensuremath{\left\{#1\right\}}}
\newcommand{\tuple}[1]{\ensuremath{\left(#1\right)}}

\newcommand{\proj}[2]{\ensuremath{\mathrm{proj}_{#2}{\left(#1\right)}}\xspace}
\newcommand{\charfunc}[1]{\ensuremath{\mathbbm{1}\left\{#1\right\}}\xspace}
\newcommand{\XX}{\ensuremath{\mathcal{X}}\xspace}
\newcommand{\YY}{\ensuremath{\mathcal{Y}}\xspace}

\newcommand{\RR}{\ensuremath{\mathbb{R}}\xspace}
\newcommand{\FF}{\ensuremath{\mathcal{F}}\xspace}

\newboolean{draftenabled}
\setboolean{draftenabled}{true} 

\ifthenelse{\boolean{draftenabled}}
{
    \newenvironment{draft}{
       
       \MakeFramed{\advance\hsize-\width\FrameRestore}}
     {\endMakeFramed}

    \newcommand{\draftinline}[1]{\colorbox{pink}{#1}}
}{
    \newcommand{\draftinline}[1]{}

}

\def\func#1{#1(\,\cdot\,)}
\def\mDU{\ensuremath{\mathcal{D}_\text{U}}\xspace}
\def\mDL{\ensuremath{\mathcal{D}_\text{L}}\xspace}

\def\bx{\ensuremath{\mathbf{x}}\xspace}
\def\bz{\ensuremath{\mathbf{z}}\xspace}

\def\argmax{\text{argmax\,}}
\def\argmin{\text{argmin\,}}

\newtheorem{definition}{Definition}
\begin{document}

\title{A Review of Pseudo-Labeling for Computer Vision}

\def\thefootnote{\fnsymbol{footnote}}

\author{\name Patrick Kage\footnotemark[1] \email p.kage@ed.ac.uk \\
       \addr School of Informatics, The University of Edinburgh, Edinburgh, UK
       \AND
       \name Jay C.~Rothenberger\footnotemark[1] \email jay.c.rothenberger@ou.edu \\
       \addr School of Computer Science, The University of Oklahoma, Norman OK, USA
       \AND
       \name Pavlos Andreadis \email pavlos.andreadis@ed.ac.uk \\
       \addr School of Informatics, The University of Edinburgh, Edinburgh, UK
       \AND
       \name Dimitrios I.~Diochnos \email diochnos@ou.edu \\
       \addr School of Computer Science, The University of Oklahoma, Norman OK, USA
    }


\maketitle

\footnotetext[1]{Equal contribution.}

\def\thefootnote{\arabic{footnote}}

\begin{abstract}

    Deep neural models have achieved state of the art performance on a wide range of problems in computer science, especially in computer vision. However, deep neural networks often require large datasets of 
    labeled samples to generalize effectively, and an important area of active research is \textit{semi-supervised learning}, which attempts to instead utilize large quantities of (easily acquired) unlabeled samples. One family of methods in this space is
    \textit{pseudo-labeling}, a class of algorithms that
    use model outputs to assign labels to unlabeled samples which are then used as
    labeled samples during training. Such assigned labels, called \textit{pseudo-labels}, are most commonly associated with the field of semi-supervised learning. In this work we explore a broader interpretation of pseudo-labels within both self-supervised and unsupervised methods. By drawing the connection
    between these areas we identify 
    new directions when advancements
    in one area would likely benefit others, such as curriculum learning and self-supervised regularization.
\end{abstract}

\section{Introduction}\label{sec:intro}
\begin{figure*}[!ht]
    \centering
    \includegraphics[width=1.0\textwidth]{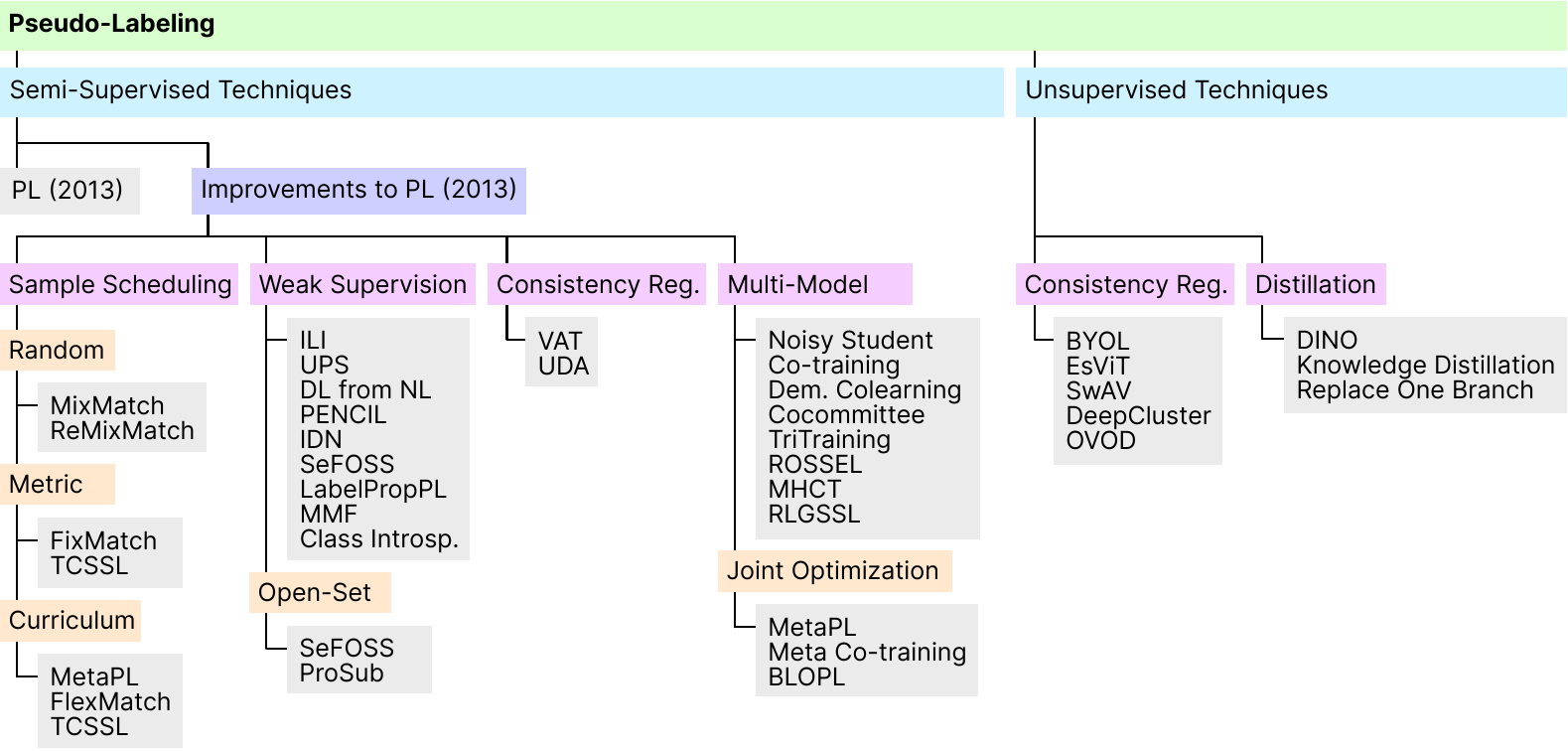}
    \caption{
       Family tree of pseudo-labeling methods.  Please see Appendix~\ref{appendix:tables} 
       for a table with section links and references.  In Table \ref{tab:pl_benchmarks} we give a performance comparison of relevant methods.
    }\label{fig:family-tree}
\end{figure*}

Deep neural networks have emerged as transformative tools especially in natural language processing and computer
vision due to their impressive performance on a broad spectrum of tasks
requiring generalization. However, a significant limitation of these systems is
their requirement for a large set of labeled data for training. This becomes
particularly challenging in niche domains such as scientific fields where human
annotation requires domain experts, making the dataset curation process
laborious and often prohibitively expensive. This review will focus on a specific methodology \textit{pseudo-labeling} (PL) within computer vision.  In Figure~\ref{fig:family-tree} we 
provide a 
taxonomy of PL, and in Table~\ref{tab:pl_benchmarks} we give a performance comparison for various methods of PL. 

\begin{table}[ht]
    \centering
    \resizebox{\columnwidth}{!}{\begin{tabular}{l|l|r|r}
        \toprule
         Method & Reference & CIFAR-10-4k & ImageNet-10\%\\
        
        \cmidrule(r){1-4}
        \multicolumn{4}{c}{Semi-supervised} \\
        \midrule
        BLOPL & \cite{BLOPL} & 96.88 & \\
        RLGSSL & \cite{RLGSSL} & 96.48 & \\
        MHCT & \cite{MHCotraining} & 96.16 & \\
        MPL & \cite{MetaPL} & 96.11 & 73.89\\
        FlexMatch & \cite{FlexMatch} & 95.81 & \\
        FixMatch & \cite{FixMatch} & 95.74 & 71.5\\
        ReMixMatch & \cite{ReMixMatch} & 94.86 & \\
        Meta-Semi & \cite{Meta-Semi} & 93.90 & \\
        InstanceDependentNoise & \cite{InstanceDependentNoise} & *93.81 \\
        UPS & \cite{InDefenseOfPL} & 93.61 & \\
        PENCIL & \cite{PENCIL} & *93.28 & \\
        MixMatch & \cite{MixMatch} & 92.71 & \\
        ICT & \cite{ICT} & 92.71 & \\
        Co-Training & \cite{deepcotraining} & 91.65 & 53.5\\
        \(\Pi-Model\) & \cite{PiModel}& 87.84 & \\
        Mean Teacher & \cite{MeanTeachers} & 84.13 & \\
        MCT & \cite{MetaCotraining} &  & 85.8\\

        \cmidrule(r){1-4}
        \multicolumn{4}{c}{Unsupervised/Self-Supervised} \\
        \midrule
        TCSSL & \cite{TCSSL} & 94.97 & \\
        UDA & \cite{UDA} & 94.53 & 68.07\\
        VAT & \cite{VAT} & 86.87 & \\
        BYOL & \cite{BYOL} &  & *89.0\\
        CLIP & \cite{CLIP} &  & 84.7\\
        DINOv2 & \cite{DINO} &  & 82.9\\
        EsViT & \cite{EsViT} &  & *74.4\\
        SimCLR & \cite{SimCLR} &  & *71.7\\
        SwAV & \cite{SwAV} &  & 70.2\\

        \bottomrule
    \end{tabular}}
    \caption{Performance of traditional pseudo-labeling for semi-supervised learning approaches and related methods in unsupervised/self-supervised learning.  Unsupervised and self-supervised methods are frequently leveraged to learn representations from unlabeled data and then those representations enable learning from few examples.  This combination of methods is semi-supervised, and that semi-supervised performance is what is shown in the second part of the table.  Models trained may differ between algorithms.  Nearly all CIFAR-10-4k models were trained using a WideResNet-28-2.  Nearly all ImageNet-10\% models were trained with ViT-L.  Approaches with different settings are marked with an (*) and the reader is encouraged to see the corresponding paper for implementation details.}
    \label{tab:pl_benchmarks}
\end{table}

\textit{Semi-supervised learning}~(SSL)
aims to address the missing-labeling problem by introducing strategies to
incorporate unlabeled samples into the training process, allowing for training
with only a small percentage of labeled samples~\citep{van2020survey,chapelle_semi-supervised_2006,zhu_introduction_2009,ASurveyDeepSemi, prakash2014survey, pise2008survey,ouali_overview_2020,ASurveyDeepSemi, SelfTrainingASurvey}. 
In recent years so-called ``scaling laws'' indicate that more data and larger models lead to better performance, thus driven by a demand for data to train ever-larger models SSL has become
an important focus of research within the machine learning research community.

\textit{Pseudo-labels} (also known as \emph{proxy labels})  
have primarily been considered in the context of 

SSL. 
However, 
pseudo-labeling can also be seen  
across sub-fields of \emph{unsupervised learning}~(UL), particularly \textit{self-supervised} learning~\citep{SwAV} and \textit{knowledge distillation}~\citep{KD-hinton}.  We formalize a definition of pseudo-labels which unifies these different areas.

In this analysis, we
investigate the ways in which
PL techniques across 
 
SSL and UL
compare and contrast with each other. 
We 
show that PL techniques largely share one or more of
the following characteristics: 
\begin{itemize}
\item \emph{self-sharpening}, where a single model is used to create the pseudo-label from a single view of a set of data points, 
\item \emph{multi-view learning}, where single samples are represented in multiple ways to promote learning generalizable patterns, 
and
\item \emph{multi-model learning}, where multiple models are used during the learning process (e.g.,~in a student-teacher setting). 
\end{itemize}
We 
outline directions for future work that lie at the intersection of 
these
areas, 
and directions illuminated in one area by considering established methods in the other.

\paragraph{Pseudo-labels within semi-supervised learning.}
Pseudo-labeling was first applied to deep learning for computer vision in~\citet{PL}.

Most techniques 
within SSL 
improve on this paper; see Figure~\ref{fig:family-tree}. In~\citet{PL}, it is argued
that PL is
equivalent to \emph{entropy minimization} (EM), maximizing
the margin of the decision boundary in a classification problem by ensuring
examples with similar labels lie close to each other on the underlying data
manifold, and far from other examples. Though not explicitly mentioned in~\citet{PL}, this is an effect of two of the three assumptions inherent to SSL
regimes: the \textit{low density separation assumption} that samples are embedded in high-density regions separated
by low-density regions and the \textit{cluster
assumption} that points within high density regions share the
same class label~\citep{chapelle_semi-supervised_2006}.\footnote{The third
assumption is the \textit{manifold assumption}, which states that
high-dimensional data lies on a low(er)-dimensional manifold embedded within
that space~\citep{chapelle_semi-supervised_2006}. 
This is 
equivalent
to saying the dimensionality of the data can be meaningfully reduced.} 

Subsequently, the
literature 
~\citep{CL4PL,FixMatch,MixMatch,ReMixMatch,TCSSL,UDA,VAT,MMF,LabelPropPL}
includes the notion of \textit{consistency regularization} (CR), in which
examples that have been slightly perturbed (e.g.,~by an augmentation or
adversarial additive perturbation) should lie close together in any embedding
space or output distribution space produced by the model.
CR and EM form strong theoretical arguments for PL with techniques using the average of predicted probabilities for augmentations to assign pseudo-labels to achieve \emph{sample scheduling} or \emph{curriculum learning} (CL) and most approaches choosing to assign ``hard'' (one-hot) pseudo-labels to examples to encourage EM~\citep{MixMatch,FixMatch,ReMixMatch,TCSSL, FlexMatch}.  When it can be assumed that examples form clusters with high internal consistency and low density separation, then it may be appropriate to perform \emph{label propagation} with methods such as \citet{LabelPropPL, LabelPropagation, MMF}.  It is natural to think of the pseudo-label assigned by some model as \emph{weak supervision}, and there are a variety of methods that leverage techniques for utilizing unreliable labels \citep{ILI, DLfromNL, PENCIL, InstanceDependentNoise, SeFOSS, ProSub}.  Furthermore, there are methods that utilize multiple models in the hopes that each model will fail independently when predicting pseudo-labels for a given instance and that this independence of failure can be leveraged to produce high-quality pseudo-labels~\citep{MetaPL, NST, cotraining, democraticcolearning, TriTraining, ROSSEL, MHCotraining, MetaCotraining, BLOPL, RLGSSL}.
Finally, we note that some of the above-mentioned lines of work combine multiple ideas. However, in Figure~\ref{fig:family-tree} we place these lines of work in the branch that we identify as most fitting for their contribution, given the taxonomy that we provide.

\paragraph{Pseudo-labels within unsupervised learning.}

Pseudo-labels have been applied in 
sub-fields 
of 
unsupervised learning.  
Self-supervised learning~\citep{DINO, EsViT, SwAV, SimCLR, CLIP, SigLIP, ALIGN, balestriero_cookbook_2023} is typically some form of CR optionally combined with contrastive learning. Knowledge distillation~\citep{KD-hinton, KD-RoB, KD-self-distill, DataFreeKD, PreparingKD} also makes use of pseudo labels to distill the knowledge of one model into another.  These sub-fields are not traditionally considered pseudo-labeling, but they all make use of one or more neural networks which provide some notion of supervision.  In some cases, this supervision corresponds to an approximation of a ground truth classification function~\citep{KD-hinton, VAT, UDA, KD-self-distill, PreparingKD}, in some cases this supervision serves only to transfer knowledge from one network (sometimes referred to as the teacher) to another (student) network~\citep{KD-hinton, KD-RoB, KD-self-distill, DataFreeKD, PreparingKD, DINO, SigLIP}, and in some cases two models provide labels to a batch of samples which are optimized to have certain properties.

\paragraph{Pseudo-labeling techniques are tolerant to label noise.}
The process of providing pseudo-labels is ultimately noisy.  PL approaches are inspired by techniques known to be effective under label noise, or new mechanisms that are tolerant to label noise which are developed in this context.

For example, the work of \citet{PAC:Noise:Angluin} on random classification noise has inspired new learning algorithms in SSL for more than two decades now; e.g.,~\citet{Cotraining:GoldmanZhu} and \citet{TriTraining}.
Similarly, other lines of work integrate recent developments on label noise, in a broader framework that uses pseudo-labels, and in particular in contexts that utilize deep neural networks; e.g.,~\citet{Noise:Label:Inspired:Recent,Noise:Label:CurriculumNet,Noise:Label:InstanceDependent:UPM,Noise:Label:New:Anchors,Noise:Label:Co-Teaching,Noise:Label:AdaptationLayer}. 
Along these lines there is work that has been done to address labeling errors by human data annotators of varying expertise (and thus of varying label quality); e.g.,~\citet{Noise:Label:New:Annotators}.
Also, quite close to this context is the use of adversarial training as a mechanism that creates more robust models by using pseudo-labels; e.g.,~\citet{VAT, UDA}.

Techniques that deal with label noise are deeply embedded to PL methods and one can find noise-tolerant mechanisms for PL approaches in all the categories shown in Figure~\ref{fig:family-tree}. 
While the investigation of label noise and methods that mitigate label noise is fascinating in its own right, it is nevertheless outside the scope of the current survey.  The interested reader may find more information in the references cited above and the references therein regarding the various approaches that are being followed towards noise mitigation.
There are also excellent surveys on the topic; e.g.,~\citet{Noise:Surveys:Label-DNN,Noise:Surveys:Representation,Noise:Label:Causality} to name a few.

\paragraph{Outline.}

The following sections give an overview of
PL approaches organized by supervision regime, mechanism of regularization used, and by model architectures (see Figure~\ref{fig:family-tree}). 
In particular, in Section~\ref{sec:prelims} we provide preliminaries so that we can clarify terms.  Of particular importance are the notions of fuzzy sets, fuzzy partitions, and stochastic labels which allow us to define the central notion of this review, that of \emph{pseudo-labels}.
Once this common language is established, we proceed in Section~\ref{sec:base-pl} with a review of the methods that belong to semi-supervised learning.
In Section~\ref{sec:pre-training} we discuss unsupervised methods, including methods of self-supervision.
In Section~\ref{sec:commonalities}
we discuss
commonalities
between the techniques that are presented in Sections~\ref{sec:base-pl} and~\ref{sec:pre-training}. 
In Section~\ref{sec:future} we discuss some directions for future work.
Finally, we conclude our review in Section~\ref{sec:conclusion}.


\section{Preliminaries}\label{sec:prelims}


Before we proceed with our presentation we give some definitions and background that can make the rest of the presentation easier and clearer.



\subsection{Fuzzy Partitions}\label{sec:prelims:fuzzy}
Fuzzy sets have the property that elements of the universe of discourse $\Omega$ belong to sets with some 
\emph{degree of membership}
that is quantified by some real number 
in the interval 
$[0, 1]$.
For this reason a function $m \colon \Omega \rightarrow [0, 1]$ is used so that $m(\omega)$ expresses the \emph{grade} (or, \emph{extent}) to which an $\omega\in\Omega$ belongs to a particular set. 
The function $m = \mu_{\mathcal A}$ is called the \emph{membership function} of 
the fuzzy set 
$\mathcal{A}$. 
Such a fuzzy set $\mathcal{A}$ is usually defined as $\mathcal{A} = (\Omega, \mu_{\mathcal{A}})$ and is denoted as $\mathcal{A} = \set{\mu_{\mathcal{A}}(\omega)/\omega \mid \omega\in\Omega}$.

\begin{example}[Illustration of a Fuzzy Set]\label{ex:fuzzy-set}
Let \XX be the universe of discourse with just four elements; 
i.e.,
$\XX = \set{a, b, c, d}$. 
Then, $\mathcal{S}_1 = \set{0.4/a, 0.1/b, 0.0/c, 1.0/d}$ is a \emph{fuzzy set} on \XX.
\end{example}

The intention is to treat the ``belongingness'' numbers as probabilities that the various elements have for being members of particular sets. 
In our case we will assume that there is a predetermined number of fuzzy sets $\mathcal{S}_1, \ldots, \mathcal{S}_k$, for some $k\ge 2$ and these will correspond to the $k$ hard labels that are available in a classification problem (more below).
Along these lines, an important notion is that of a \emph{fuzzy partition}, which we will use to define pseudo-labeling precisely.  

\begin{definition}[Fuzzy Partition]\label{def:fuzzy-parition}
Let $k \ge 2$. Let $\mathcal{I} = \set{1, \ldots, k}$.
Let $\mathcal{Q} = \left(\mathcal{S}_1, \ldots, \mathcal{S}_k\right)$ be a finite family where for each $i\in\mathcal{I}$,  
$\mathcal{S}_i$ is a fuzzy set and $\mu_{\mathcal{S}_i}$ is its corresponding membership function.
Then, $\mathcal{Q}$ is a \emph{fuzzy partition} if and only if it holds that $(\forall x\in\XX) \left[\sum_{i\in\mathcal{I}} \mu_{\mathcal{S}_i}(x) = 1\right]$.
\end{definition}

\begin{example}[Illustration of a Fuzzy Partition]\label{ex:fuzzy-partition}
Let $\mathcal{I} = \set{1, 2, 3}$.
Let $\XX = \set{a, b, c, d}$ and consider the following fuzzy sets
\begin{displaymath}
\left\{
\begin{array}{rcl}
    \mathcal{S}_1 &=& \set{ \ 0.4/a, \ 0.1/b, \ 0.0/c, \ 1.0/d \ }  \\
    \mathcal{S}_2 &=& \set{ \ 0.3/a, \ 0.6/b, \ 0.6/c, \ 0.0/d \ }  \\
    \mathcal{S}_3 &=& \set{ \ 0.3/a, \ 0.3/b, \ 0.4/c, \ 0.0/d \ }  \\
\end{array}
\right.
\end{displaymath}
The above family of fuzzy sets $\mathcal{S}_1, \mathcal{S}_2, \mathcal{S}_3$ together with the corresponding membership functions $\mu_{\mathcal{S}_1}, \mu_{\mathcal{S}_2}, \mu_{\mathcal{S}_3}$ whose values are shown in the equation on display, provides a fuzzy partition since, for every $x\in\XX$, it holds that $\sum_{i\in\mathcal{I}} \mu_{\mathcal{S}_i}(x) = 1$.
\end{example}

More information on fuzzy sets is available at~\citet{Fuzzy:Zadeh,fuzzypart}.

\subsection{Basic Machine Learning Notation}
We use \XX to denote the \emph{instance space} (or, \emph{sample space})
and \YY the \emph{label space}. 
We care about classification problems in which case it holds that $\abs{\YY} = k$; that is there are $k$ \emph{labels} (\emph{categories}).
Instances will have \emph{dimension} $n$. 
An \emph{example} $(\bx, y) \in \XX\times\YY$ is a \emph{labeled instance}.
We denote an \emph{unlabeled dataset} with \mDU; e.g., $\mDU = \left(\bx_1, \ldots, \bx_m\right)$ is an unlabeled dataset composed of $m$ instances.
If in addition, we have the true labels of all these instances, then we are working with a \emph{labeled dataset} denoted with \mDL; e.g., $\mDL = \left((\bx_1, y_1), \ldots, (\bx_m, y_m)\right)$ is a labeled dataset with $m$ examples.

Given access to datasets \mDU and \mDL, the goal of semi-supervised learning is to develop a \emph{model} $f$ that approximates well some underlying function that maps the instance space \XX to the label space \YY. 
Examples of this behavior that we want to learn are contained in \mDL and additional instances are available in \mDU. 
The model that we want to learn has trainable parameters $\mathbf{\theta}$ that govern its predictions and for this reason we may write the model as $f_{\mathbf{\theta}}$ so that this dependence on $\mathbf{\theta}$ is explicit. 
The space of all possible functions $f_{\mathbf{\theta}}$ that correspond to all the possible parameterizations $\mathbf{\theta}$ is called the \emph{model space} and is denoted by $\FF$.
With all the above definitions, in our context, a \emph{supervised learning problem} is specified by the tuple $(\XX, \YY, \FF, \mDL)$, whereas a \emph{semi-supervised learning problem} is specified by the tuple $(\XX, \YY, \FF, \mDL, \mDU)$.

We assume familiarity with artificial neural networks. 
However, the interested reader can find more information in a plethora of resources, including, e.g.,~\citep{Book:Understanding,Book:Mitchell,Book:StatLearning:Python,Book:HandsOn}.
\label{subsec:PL-ex}
To illustrate the relevance of Definition \ref{def:fuzzy-parition}, we can see how it applies to the first deep learning application of pseudo-labeling \citet{PL}.  
Lee utilizes a neural network to perform image classification, which we will call $f_\theta$.  The neural network defines a fuzzy partition.  For each classification problem there is an input space of images $\mathcal{X}$ and a number of target classes $k$. For $x \in \mathcal{X}$ we have $f_\theta (x) \in \Delta^{k}$,  where $\Delta^k$ is the probability simplex of dimension~$k$. Thus, each position of the vector $f_\theta (x)$ represents a class probability, or equivalently membership in a fuzzy set $\mathcal{S}_i$.  Note that as training proceeds this network might be updated, but the network will continue to provide pseudo-supervision.  

\subsubsection{Deeper Discussion on Labels}
Below we provide more information on labels and pseudo-labels.
In addition, $\Delta^k$ will continue to be the probability simplex of dimension $k$.

\begin{definition}[Stochastic Labels]\label{def:labels:stochastic}
Let $\abs{\YY} = k$.
\emph{Stochastic labels} are vectors $y \in \Delta^{k}$ where the positions of the vector defines a probability distribution over the classes $\YY_i\in\YY$ e.g., $y = \tuple{p_1, \dots, p_k}$. 

\end{definition}
The idea of stochastic labels is very natural as the output of deep neural networks almost always has the form shown above.  
In order to make sure that the role of fuzzy sets and fuzzy partitions is clear to the reader, and what the relationship of these terms is to stochastic labels, below we revisit Example~\ref{ex:fuzzy-partition} and make some additional comments.

\begin{remark}[Stochastic Labels from Fuzzy Partitions]
In Example~\ref{ex:fuzzy-partition}, the fuzzy sets $\mathcal{S}_1$, $\mathcal{S}_2$, and $\mathcal{S}_3$ correspond to three different labels. Furthermore, by looking at vertical cuts of these three fuzzy sets, for each one of the four instances $a, b, c, d\in\XX$ one can obtain the stochastic label that is associated with the respective instance. For example, the stochastic label of instance $b$ is $(0.1, 0.6, 0.3)$.
\end{remark}

In addition, we note that the notion of stochastic labels also makes sense even without considering predictions of a model, but rather allowing instances to have fuzzy membership in different classes; see, e.g.,~\citep{MixUp}.

\begin{definition}[Pseudo-Labels]\label{def:labels:pseudo}
Given a fuzzy partition $\mathcal{Q}$ over $\mathcal{X}$, pseudo-labels are vectors $y \in \Delta^{k}$ where each position of the vector defines the probability of membership in a unique fuzzy set in $\mathcal{Q}$ (corresponding to a label), but which were not specified as part of the learning problem. For some instance $x \in \XX$ we have $y = \tuple{\mu_{\mathcal{S}_1}(x), \ldots, \mu_{\mathcal{S}_k}(x)}$.  We write $y = \mathcal{Q}(x)$.
\end{definition}

Using the notation introduced earlier, for some instance $x$, the pseudo-label takes the form $f_\theta (x)$. In other words, a pseudo-label is a stochastic label by definition.  However, the crucial point is that a pseudo-label is obtained as a result of a predictive model $f_{\theta}(x)$ applied on the instance $x$, indicating the probability that $x$ has belonging to different classes according to $f_{\theta}$.
Ideally, generated pseudo-labels are similar to ground-truth labels for the various instances $x\in\XX$. Good pseudo-labels are similar to the ground truth to the extent that the ground truth is not significantly different from the labeled dataset $\mDL$ (e.g.,~because of noise). 
Because each element of the pseudo-label vector indicates membership in a particular class, the pseudo-label can also be one-hot (e.g.,~see instance $d$ in Example~\ref{ex:fuzzy-partition}).
The difference between the ground truth label and the pseudo-label in that case is that the latter was generated by the fuzzy partition using  
the instance.

\begin{definition}[Pseudo-Labeling]
    Utilizing pseudo-labels generated by at least one fuzzy partition $\mathcal{Q}$ of the input space $\mathcal{X}$ inferred from only instances and ground truth labels to provide supervision during the training of a machine learning algorithm is called \emph{pseudo-labeling}.
\end{definition}

In other words, pseudo-labelling is the process of generating/inferring stochastic labels. In our example the fuzzy class partition defined by $f_\theta$ is used to assign pseudo-labels to unseen instances which are then used to optimize the network further.

\begin{definition}[Pseudo-Examples]
    Pseudo-labeled examples are tuples $\tuple{x, \mathcal{Q}(x)}$.
\end{definition}

The examples that are formed by the tuples of instances and pseudo-labels are the pseudo-examples which are used during retraining.

\subsubsection{Data Augmentation} 
Data augmentation is a common technique across multiple branches of machine learning disciplines, broadly attempting to generate a set of derivative samples \(\tilde{\bx}\) from a single sample \(\bx\) by modifying the sample in some way. Augmenting samples can help normalize skewed sample class counts by artificially inflating a rare subclass~\citep{DataAugSurvey}, and are used within many unsupervised learning methods to generate distinct samples that are known to be similar for consistency regularization (discussed in Section~\ref{sec:pre-training}). These augmentations vary by implementation, but generally fall into one of the following categories~\citep{DataAugSurvey}: \textit{flipping} (flips the image 180 degrees along the horizontal or vertical axes), \textit{color manipulation} (alters the color information of the image), \textit{cropping} (randomly crops the image), \textit{affine transformations} (rotates, translates, or shears the image), \textit{noise injection} (add random noise to the image), \textit{kernel filters} (applies e.g.,~a Sobel filter over the image), or random masking (blanks out a section of the image).

\section{Semi-Supervised Regimes}\label{sec:base-pl}

The classic formulation of 
PL
techniques as applied to deep neural networks is defined in~\citet{PL}, 
where PL is presented as a fine-tuning stage in addition to normal training. 
At a high level, the model $f_\theta$ 
is initially trained 
over samples 
from the labeled set \(\mDL\), and after a fixed number of epochs $\theta$ is frozen and
pseudo-examples derived from $f_\theta$ and \(\mDU\) 
are added into the training steps, 
yielding an effective 
SSL 
technique.  We use the symbol $f_{\theta'}$ to denote the model trained using the pseudo-examples. 

More specifically, 
this 
formulation of 
PL
derives labels for samples drawn from
\(\mDU\) 
by ``sharpening'' the predictions of the partially-trained network, 
effectively reinforcing the partially-trained model's ``best guess'' for a particular sample. The loss function \(\mathcal L\) being optimized 
has
a supervised component \(\ell_\text{S}\) 
and a nearly-identical unsupervised component \(\mathcal \ell_\text{U}\); 
in every case cross-entropy loss is used to penalize the model's predictions against, respectively, the true labels, or the pseudo-labels obtained by the sharpening process:

\begin{equation}\label{eq:pl-loss}
        \mathcal L = \underset{\text{supervised}}{\ell_\text{S}(y, f(\bx_\text{S}))} + \underset{\text{unsupervised}}{\alpha(t) \ell_\text{U}(f_{\theta}(\bx_\text{U}), f_{\theta'}(\bx_\text{U}))}
\end{equation}

Crucially, 
the unsupervised segment is moderated by \(\alpha(t)\) an epoch-determined annealment parameter. 
This starts at zero and then ramps linearly up to a ceiling value. 
According to~\citet{PL}
careful consideration is required in choosing the scheduling of $\alpha$, 
as setting it too high will disturb training of labeled data and too low will not have an impact on the training process at all.

This is the basic foundation upon which most fine-tuning 
PL
papers are built. The remainder of this section presents 
modifications and improvements to this formulation.  It should be clear from the previous sections how this fits into our definition for pseudo-labeling: the unsupervised portion of the loss is computed through the use of an inferred fuzzy partition (defined by the weights of the model trained so far).  In the case of hard pseudo-labels the membership probabilities are one-hot, and in the case of soft pseudo-labels they may not be. \citet{PL} 
evaluates adding a denoising variational autoencoder in order to simplify the representations of the samples in both datasets 
$\mDL$ and $\mDU$
to boost the performance of the 
PL steps. For example, assigning sharpened labels 
to samples drawn 
from \(\mDU\) can 
make the training objective 
unstable when pseudo-labeling 
some unlabeled samples changes dramatically from one weight update to the next~\citep{PLandCB}. Consequently, one of the primary focuses of work in this direction is how to best choose the pseudo-labels or examples such that training progresses advantageously.

In this section we will provide an overview of pseudo-labeling methods in computer vision for semi-supervised learning.  We will discuss methods based on scheduling unlabeled instances (Section~\ref{sec:sample-scheduling}), methods based on performing well under weak supervision (Section~\ref{sec:weak-supervision}), methods of consistency regularization(Section~\ref{sec:ssl:consistency-regularization}), and approaches that use multiple models (Section~\ref{sec:multimodel}).

\subsection{Sample Scheduling}\label{sec:sample-scheduling}

An inherent prerequisite to solving the problem of generating pseudo-labels for
samples is determining for \textit{which} samples to generate pseudo-labels.
The precise sample schedule can greatly impact the performance of the final
pseudo-labeling algorithm~\citep{PLandCB}. Across the literature, a variety of
techniques have been proposed as improvements to~\cite{PL} by altering the manner in which unlabeled instances are selected and how pseudo-labels are assigned to them.

\subsubsection{Random Selection}\label{sec:random-selection}
The simplest way to choose unlabeled samples to label is to sample them
randomly from the unlabeled set. 
With this method of selection it is assumed that the unlabeled data follow the same distribution as the labeled data, which may not be realistic in the real world thus leading to poor generalization ability of the final model.

Pseudo-labeling 
techniques are vulnerable to \emph{confirmation bias}, where
initially-wrong predictions are 
reinforced by the PL 
process~\citep{PLandCB}. 
One 
solution is to reduce the confidence of the network
predictions overall~\citep{PLandCB}. This is a feature of \emph{MixMatch}~\citep{MixMatch}, a hybrid 
PL 
technique combining several dominant methods within the SSL space with the goal of preventing overconfident predictions. 
MixMatch consists of several steps: a \textit{data augmentation} step, a \textit{label guessing} step (the 
PL
component), and a \textit{sample mixing} step. First, a batch of samples is drawn from both the labeled set \(\mathcal X \subset \mDL\) and the unlabeled set \(\mathcal U \subset \mDU\). Samples from \(\mathcal X\) are then replaced with stochastically-augmented versions to form \(\hat{\mathcal X}\), and samples from \(\mathcal U\) are replaced by \(K\) augmentations of the each sample to form \(\hat{\mathcal U}\). Samples from \(\mathcal U\) are 
pseudo-labeled 
based on the sharpened average of the model's predictions for all \(K\) augmentations of each sample. 
Finally, these samples are then shuffled and mixed using MixUp~\citep{MixMatch,MixUp} to yield two mixed distributions \(\mathcal{X}^\prime,\mathcal{U}^\prime\) containing labeled and unlabeled samples.

\emph{ReMixMatch}~\citep{ReMixMatch} was proposed with two additional concepts on top of MixMatch: \textit{distribution alignment} and \textit{augmentation anchoring}, and a modification of the loss function to improve stability. Distribution anchoring is the scaling of each prediction's output probabilities element-wise by the ratio of the output to the average of the outputs over a number of previous batches---replacing the sharpening function in MixMatch and serving as an 
EM 
term. The other contribution was augmentation anchoring, which replaced MixMatch's one-step stochastic augmentation of a true sample with a two step process: first weakly augmenting a sample, and then using that sample/label pair as a base for a series of \(K\) strong augmentations (using their proposed CTAugment, a version of AutoAugment~\citep{AutoAug}). This functions as a consistency regularization technique, and is similar to methods discussed in Section~\ref{sec:pre-training} such as SimCLR~\citep{SimCLR}. 
Additionally, ReMixMatch's loss function incorporates two additional targets: a \textit{pre-MixUp unlabeled cross-entropy loss} component (comparing unmixed unlabeled samples with their guessed labels) and a \textit{rotation loss} component (rotating a sample randomly in \(90\deg\) increments and predicting the rotation of the sample). These changes yield a significant improvement in the label efficiency of the model, with the authors showing a \(5 - 16\times\) reduction in required labels for the same accuracy.

Sample scheduling approaches, such as the *-Match methods, rely on a carefully chosen recipe of data augmentation strategies.  These strategies are essential to effective consistency regularization, and to some extent entropy minimization as well.  For some well-known dataset there exist popular data augmentation policies like those of AutoAugment on CIFAR10 and CIFAR100, however for novel domains it may require extensive experimentation to construct such a policy.  Since constructing an effective strong data augmentation strategy is a prerequisite to applying the above semi-supervised methods, this may pose a limitation to applying them to some data domains.

\subsubsection{Confidence-Based Selection}\label{sec:label-metric-based}
An alternative approach to random sampling is metric-based sampling, where the sample to pseudo-label is selected from the unlabeled pool \(\mDU\) via a non-random metric such as confidence. One such method is \emph{FixMatch}~\citep{FixMatch}; 
a simplification of the ReMixMatch process. FixMatch again has two loss terms.
The labeled loss is cross entropy computed over labels given weakly augmented (flipped and cropped) versions of supervised examples. The unlabeled loss is computed only for confident predictions on weakly augmented versions of the images from the unlabeled set. This is equivalent to filtering unlabeled data based on confidence. Pseudo-labels are again computed from the weakly augmented versions of unlabeled samples, but in FixMatch, hard labels are assigned. FixMatch includes the labeled data (without labels) in the unlabeled set during training. 

While most methods use confidence as the metric by which to select,
\textit{Time-Consistent Self-Supervision for Semi-Supervised Learning}
(TCSSL)~\citep{TCSSL} uses a different metric. For each gradient update step, TCSSL
maintains a ``time consistency'' metric which is the exponential moving average
of the divergence of the discrete output distribution of the model for each
sample. They show this metric is positively correlated with accuracy,
especially at later epochs, and that the more commonly used confidence is not.
In addition to PL, this approach includes jointly in its objective
representation learning losses.

Selecting pseudo-labels based on a model's confidence may not be effective if the model's uncertainty is poorly calibrated.  In general it is known that training of neural networks via gradient descent methods does not necessarily produce calibrated uncertainty estimates.  

\subsubsection{Curriculum Learning}\label{sec:curriculum-learning}
One of the most natural solutions to choosing a suitable set of examples to
pseudo-label is \textit{curriculum learning} (CL)~\citep{Curriculum}. In this paradigm, the
model is first allowed to learn on ``easy'' instances of a class or concept and
then trained on progressively ``harder'' instances.  Instances which are ``easy'' are ones which will be given correct pseudo-labels (with high probability), and thus we have extended our labeled set reliably.  Through training in this way the hope is that harder examples become easier and we can continue to expand our labeled set. Unfortunately designing a curriculum is not easy, and often requires handcrafting a heuristic for determining what is easy and what is hard.

One such CL approach is~\citet{CL4PL}. Functionally, it is a self-training framework where the within each epoch unlabeled
data is sorted based on the confidence of prediction, and only the top \(k\%\)
most confident predictions are given pseudo-labels at each iteration of self
training until all labels are exhausted. Curriculum learning is also the foundation of FlexMatch~\citep{FlexMatch}, an extension to the *-Match family of methods (see 
Section~\ref{sec:random-selection}). 
FlexMatch extends FixMatch's pseudo-label thresholding by introducing dynamically-selected thresholds per-class (dubbed \emph{Curriculum Pseudo-Labeling} or CPL), and these are scaled to the ``difficulty'' of learning a particular class' label. This is achieved through two assumptions: the first that the dataset is class-balanced and the second that the difficulty of learning a particular class is indicated by the number of instances with a prediction confidence above a given threshold \(\tau\). During the training process, \(\tau\) is adjusted down for ``hard'' classes to encourage learning on more pseudo-labels, and up for ``easy'' classes in order to avoid confirmation bias~\citep{FlexMatch}. 
This 
improves 
SSL performance 
without much additional cost, as there are no extra forward passes and no extra learnable parameters~\citep{FlexMatch}.

As we discussed previously, neural networks do not necessarily produce calibrated uncertainty estimates and choosing an effective data augmentation strategy is not always trivial.  Both of these limitations apply to the above methods with respect to choosing a curriculum for pseudo-labeling.  Additional effort may have to be invested in calibrating model outputs or developing an effective data augmentation strategy to apply methods of curriculum-learning-based pseudo-labeling.

\subsection{Weak Supervision}\label{sec:metric-based}\label{sec:weak-supervision}

Subtly different from metric-based selection of unlabeled examples, we can also
imagine assigning a label based on a metric such as using a
\(k\)-nearest-neighbors (\(k\)NN) classifier for PL. This updated task brings
pseudo-labeling to \textit{weak supervision}, a class of problem similar to SSL
where the labels in \(\mDL\) are treated as inaccurate, and are updated
along with the pseudo-labels at training time.

An early approach in this space was~\citet{LabelPropagation}, which used
\(k\)NN to assign a label based on its geodesic distance to other samples in
the embedding space. This approach was extended to deep neural networks
in~\citet{LabelPropPL}. Under the deep learning regime, a CNN is trained on the
labeled data and the representation output by the penultimate layer is used as
an approximation of the underlying data manifold. Then, an iterative algorithm
assigns soft pseudo-labels to the unlabeled data based on density and geodesic
distance to labeled examples. The CNN is retrained on these new targets in
addition to the original labels. This process is iterated until some stopping
criterion is satisfied.~\citet{MMF} improves upon this with a regularization
they call ``Min-Max Feature,'' this is a loss which encourages all examples of
one class assignment to be embedded far from those of other classes, and
examples within an assignment to be drawn together. During model optimization,
this is performed during a separate step subsequent to the optimization of the
cross entropy loss over the union of the labeled and pseudo-labeled sets, and
thus the method is included in semi-supervised section. At the beginning of
each self-training iteration hard pseudo-labels are assigned as the argmax of
the output of the CNN.

\textit{Iterative Label Improvement} (ILI)~\citep{ILI} recognizes the potential inaccuracies in
supervised dataset labels, emphasizing the inherent noise in pseudo-labels. ILI
employs a self-training method where the model, post-initial training, may
adjust any sample's label based on confidence levels, without calibration.
Additionally, ILI extends through \emph{Iterative Cross Learning} (ICL)~\citep{ICL}, where models
trained on dataset partitions assign pseudo-labels to other segments.

It is known that without special care the outputs of neural networks can be
highly uncalibrated~\citep{Calibration}, that is to say they are very over- or
under-confident. This is a serious issue when confidence is the metric by
which unlabeled examples are chosen to be given pseudo-labels.
\textit{Uncertainty-Aware Pseudo-Label Selection} (UPS)~\citep{InDefenseOfPL} seeks to
combat this by training a calibrated network for confidence filtering on the
unlabeled set. Popular ways to train uncertainty-aware networks include
Bayesian Neural
Networks~\citep{PracticalVariationalInference,WeightUncertainty,GaussianPosteriors},
Deep Ensembles~\citep{SSUncertainty}, and MC Dropout~\citep{MCDropout}, but these
all come at the expense of extra computation. UPS uses a function of the model
confidence and uncertainty estimation to filter unlabeled data. 
This approach empirically
works quite well, but is limited because it requires the training of a deep
ensemble for each dataset for uncertainty estimation. Pseudo-labels generated
by this method are hard, and only computed after each iteration of
self-training.

Continuing in the theme of directly estimating the label noise,~\citep{DLfromNL} adds a probabilistic model which predicts the clean label for an instance. This approach splits noise in labels into two categories: ``confusing noise'' which are due to poor image/label matching (though still conditional on the label), and ``pure random noise'' which is unconditional and simply wrong. The model represents the image/noisy label relationship by treating true label/noise types as latent variables. This latent variable, $\mathbf z$, is a one-hot encoded 3-element vector with the following properties:

\begin{enumerate}
    \item
        $\mathbf z_1 = 1$: Label is noise free\vspace{-.5em}
    \item
        $\mathbf z_2 = 1$: Label is unconditionally noisy (uniform random distribution)\vspace{-.5em}
    \item
        $\mathbf z_3 = 1$: Label is conditionally noisy (moderated by learnable matrix $\mathbf C$)
\end{enumerate}

Two CNNs are then trained to estimate the label and noise type respectively: $p(\mathbf y\mid\mathbf x)$ and $p(\mathbf z \mid \mathbf x)$. The CNNs are pretrained using strongly supervised data, and then noisy data is mixed into clean data on an annealing schedule. In benchmarks, this approach yielded a $78.24\%$ test accuracy, a modest improvement over treating the noisy dataset as unlabeled and using pseudo-labeling ($73.04\%$ test accuracy) and treating the noisy labels as ground truth ($75.30\%$ test accuracy).

More recently, PENCIL~\citep{PENCIL} handles uncertainty by modeling the label for an image as a distribution among all possible labels
and jointly updates the label distribution and network parameters during training. PENCIL extends DLDL~\citep{DLDL} by using the inverse-KL divergence (\(\text{KL}(f(\mathbf x_i;\mathbf\theta\Vert  \mathbf y^d_i )) \) rather than \(  \text{KL}(\mathbf y^d_i \Vert f(\mathbf x_i;\,\mathbf\theta)) \)) to moderate the label noise learning, as KL divergence is not symmetric and this swap gives a better gradient (\textit{i.e.,} not close to zero) when the label is incorrect. Additionally, two regularization terms are added to control the learning---a \textit{compatibility loss} $\mathcal L_o$ and an \textit{entropy loss} $\mathcal L_e$. The compatibility loss term ensures that the noisy labels and estimated labels don't drift too far apart---given that the noisy labels are mostly correct---and is a standard cross-entropy between the two terms. The entropy loss $\mathcal L_e$ serves to ensure that the network does not just directly learn the noised weights, but rather encourages a one-hot distribution.  These one-hot labels are correct for classification problems and keeps training from stalling. The final loss is a sum of the classification loss and hyperparameter-moderated compatibility/entropy losses. Similar to other pseudo-labeling techniques, PENCIL pre-trains the backbone network assuming that the existing labeling is correct. Subsequently, the compatibility and entropy losses are brought in for several epochs before a final fine-tuning is done over the learned new labels. This approach makes PENCIL useful for both semi-supervised and weakly supervised regimes.

Another approach in this space is by \citet{InstanceDependentNoise}, which directly models label noise by relating noisy labels to instances. Here, a DNN classifier is trained to identify an estimate of the probability $P( \text{confusing}\mid x)$, and this classifier is optimized using a variation of the expectation-maximization algorithm. This is shown to be slightly more effective than PENCIL, with the added benefit that the confusing instance classifier gives an explainable view of which instances are confusing.

An additional strategy for label selection is \textit{Class Introspection}~\citep{ClassIntrospection}, which uses an explainability method over a classifier in order to embed the inputs into a space separated by their learned explanations. This space is then hierarchically clustered in order generate a new (pseudo-) labeling, with the intuition that points which were included in a single class but are explained differently may belong in a separate, new labeling.

\subsubsection{Open-set Recognition}\label{sec:open-set}

Open-set recognition (OSR) and pseudo-labeling, while distinct, share a
connection in weakly-supervised tasks. OSR aims to identify known classes while
detecting and rejecting unknowns (out-of-distribution
instances)~\citep{OSR_Survey}, a challenge that becomes more complex without
strong labels. In the context of weakly-supervised OSR, pseudo-labeling can be
employed to create initial, albeit noisy, labels for the known classes in
unlabeled data. These pseudo-labels help define boundaries for known classes,
which is crucial for OSR to distinguish them from unknowns. Thus,
pseudo-labeling provides a form of implicit supervision, enabling OSR models to
learn and differentiate classes even with limited or absent explicit labels.

An approach in this space is SeFOSS~\citep{SeFOSS}, which aims to learn from all
data (in and out of distribution) in \(\mDU\) and uses a consistency
regularization objective to pseudo-label inliers and outliers for the dataset.
SeFOSS utilizes the \textit{free energy score} of the dataset (as defined
in~\citep{FreeEnergy}) rather than a softmax objective to filter \(x \in \mDU\)
into three buckets: \textit{pseudo-inliers}, \textit{uncertain data}, and
\textit{pseudo-inliers}. Similar to FixMatch~\citep{FixMatch}, these are sorted
by thresholding, however because 
free energy is based on the raw logits rather than a normalized probability
distribution SeFOSS delegates these (scalar) thresholds to be assigned as
hyperparameters.

Free energy is not the only metric used for in-/out-of-distribution assignment.
ProSub~\citep{ProSub} uses the angle \(\theta\) between clusters in the
penultimate layer activations of a neural network to assign instances to
the in- or out-distribution, and avoids having to manually set thresholds by
estimating the probability of being in-distribution as a Beta distribution.
This Beta distribution is computed using a variant of the standard EM
algorithm, and the resulting ID/OOD prediction is used to weight the instance in the
loss function. ProSub finds that this achieved SOTA results on benchmarks,
showing the efficacy of this method.

\subsection{Consistency Regularization}\label{sec:ssl:consistency-regularization}
A common problem with models trained on few labeled examples that of overfitting.  There are two representative techniques which inject consistency regularization in a way which specifically addresses the case where pairs of instances that differ only slightly receive differing model predictions.  These are Unsupervised Data Augmentation (UDA) by~\citet{UDA} and Virtual Adversarial Training (VAT) by~\citet{VAT}.  It is possible to apply these methods without any supervision at all, and in that case they are still pseudo-labeling, but they would be unsupervised rather than semi-supervised.  UDA is even explicitly called \emph{unsupervised} data augmentation, but because historically these two works are specifically for semi-supervised applications we include them in this section rather than the next.

Unsupervised data augmentation~\citep{UDA} is a method for enforcing consistency regularization by penalizing a model when it predicts differently on an augmented instance. UDA employs a consistency loss between unaugmented and an augmented version of a pseudo-labeled instance.  If the two predictions differ, then the pseudo-label on the unaugmented instance is assumed to be correct and the model is trained to reproduce that label for the augmented instance.  In this way the model learns to be invariant to the augmentation transformations.  For its simplicity UDA is remarkably effective, and it bears strong resemblance to many of the CR methods of self-supervised learning in the next section.  UDA does, however, require an effective data augmentation strategy and is typically achieved with a strategy that is known to be effective already on a particular dataset.

\textit{Virtual adversarial training} (VAT) is a CR technique which augments the training set by creating new instances that result from original samples with small adversarial perturbations applied to them.  A consistency loss is applied between the new adversarial instance and the original instance.  If the model's prediction differs then the pseudo-label for the original instance is assumed to be correct.  The goal of this process is to increase the robustness and generalization of the final model. Rather than enforcing consistency by encouraging the model to be invariant to combinations of pre-defined transformations, VAT makes the assumption that examples within a ball of small radius around each instance should have the same label.  

In both of the above cases the pseudo-label of an augmented instances is inferred by its close proximity to the original sample.  This distance is either a p-norm in the case of VAT or a semantic notion of distance as in UDA.  The fuzzy partition is thus constituted of fuzzy sets which are the union of balls of a chosen radius around points of the same class, or which are sets of augmented versions of an instance.  In the case of VAT the labels that are assigned are one-hot, but this is a subset of the admissible fuzzy partitions. Adversarial training is a special case of consistency regularization in which the consistency is enforced locally for each instance.   Adversarial training is an effective method of consistency regularization, but it incurs the additional cost of a gradient update to the input sample.

\subsection{Multi-Model Approaches}\label{sec:multimodel}
In 
multi-model approaches
more than
one network is used for training 
one singular model, combining
the strengths of the individual component networks. 
Meta Pseudo Label (MPL) by~\citet{MetaPL}, 
discussed in 
Section~\ref{sec:joint-optimization} 
of joint-optimization selection,
is an example where 
a teacher
model generates labels and a student model 
processes labeled data and
jointly optimizes both.

Noisy Student Training~\citep{NST} is an effective technique in this category where a student model is initially trained on labeled and pseudo-labeled data. This student then becomes the teacher, and a new, possibly larger, ``noised'' student model takes its place, undergoing augmentation and stochastic depth adjustments. Training involves using a mix of supervised and pseudo-labeled data, with the teacher model filtering pseudo-labeled examples based on confidence. The process typically involves a few iterations of self-training, after which the final student model is selected. This approach differs from model distillation in that the student model is often larger, subjected to noise, and utilizes unlabeled data, which is not the case in distillation. The addition of noise is believed to make the student model mimic a larger ensemble's behavior.

However, the student-teacher model is not the only multi-model architecture
that relies on pseudo-labels for SSL.  A representative
technique in this area is \textit{co-training}~\citep{cotraining} 
in which two models are trained 
on disjoint feature sets which are both sufficient for
classification. In co-training, pseudo-labels are iteratively incorporated into the 
training set just as with self-training, except labels added to one model's training set
are generated by the other model. The models need not have the same number of parameters,
the same architecture, or even be of the same type. One key problem in co-training is 
that of constructing the views when they are not available~\citep{viewconstruction, RASCO}. It
is relatively uncommon to have a dataset in which there are two independent and sufficient
views for classification.  
Furthermore, it is possible that the views we have are independent, but not sufficient.  
Multiple works 
attempt to address the issue of insufficiency. 
\citet{insufficientviews} shows that
this is still an acceptable setting in which to perform co-training provided that
the views are ``diverse.'' 

To address the problem of the lack of independent views, some algorithms
propose using a diverse group of classifiers on the same view instead of
constructing or otherwise acquiring multiple sufficient and independent
views~\citep{democraticcolearning, cocomittee}. These approaches are
ensembles where each classifier issues a vote on the class of unlabeled
samples, and the majority vote is assigned as the pseudo-label and the
instance is added to the training set.  In this setting, it is reasonable to
assume that the single view is sufficient, as in most supervised learning
scenarios we are simply attempting to train the best classifier we can with the
data we are given. Independence is far more dubious, as we rely on the
diversity of the ensemble to ensure our classifiers do not all fail for the
same reasons on points which they classify incorrectly.

In the same vein, \textit{Tri-Training} 
employs three
classifiers in order to democratically select pseudo-labels for \(\mDU\) after being trained on \(\mDL\)~\citep{TriTraining}.
While this is a 
simple solution, it comes with a \(3\times\) increase in compute cost. An
improvement 
is 
\emph{Robust Semi-Supervised Ensemble
Learning}~\citep{ROSSEL}, which formulates
the 
PL 
process as an ensemble of low-cost supervisors whose
predictions are 
aggregated together with a weighted SVM-based solver.

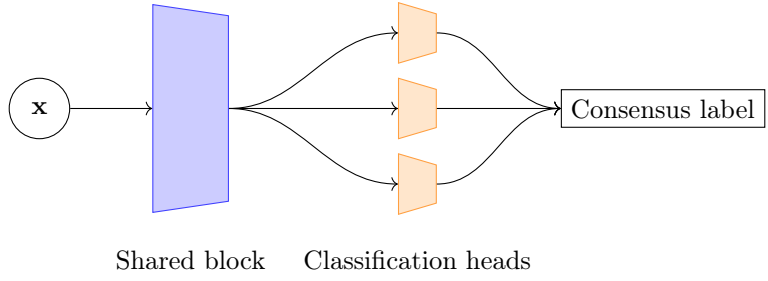
\begin{figure}[ht]
    \centering
    \begin{tikzpicture}
        [
            every node/.style={execute at begin node=\small},
            nnlayer/.style={trapezium, fill=#1!20, draw=#1!75, text=black, shape border rotate=270, trapezium stretches body},
            sample/.style={circle,draw=black,minimum size=8mm},
            consensus/.style={rectangle,draw=black},
            scale=1
        ]

        \node [sample](input) at (-2,0) {\(\bx\)};

        \node [nnlayer=blue, minimum width=2.75cm, minimum height=1cm, trapezium stretches body](nnshared) at (0,0) {};

        \begin{scope}[shift={(3,0)}]
54            \node [nnlayer=orange, minimum width=0.8cm, minimum height=0.5cm](nnhead1) at (0,-1) {};
            \node [nnlayer=orange, minimum width=0.8cm, minimum height=0.5cm](nnhead2) at (0,0) {};
            \node [nnlayer=orange, minimum width=0.8cm, minimum height=0.5cm](nnhead3) at (0,1) {};
            \node at (0,-2) {Classification heads};
        \end{scope}

        \begin{scope}[shift={(6.25,0)}]
            \node [consensus](consensus) at (0,0) {Consensus label};
        \end{scope}

        \node at (0,-2) {Shared block};


        \draw [->] (input.east) -- (nnshared.west);

        \draw [->] (nnshared.east) to [out=0,in=180] (nnhead1.west);
        \draw [->] (nnshared.east) to [out=0,in=180] (nnhead2.west);
        \draw [->] (nnshared.east) to [out=0,in=180] (nnhead3.west);

        \draw [->] (nnhead1.east)  to [out=0,in=180] (consensus.west);
        \draw [->] (nnhead2.east)  to [out=0,in=180] (consensus.west);
        \draw [->] (nnhead3.east)  to [out=0,in=180] (consensus.west);

    \end{tikzpicture}
    \caption{
        A simplified diagram of Multi-Head Co-Training's architecture. Adapted
        from~\citet{MHCotraining}.
    }\label{fig:mhct-arch}
\end{figure}

\emph{Multi-Head Co-Training} (MHCT)~\citep{MHCotraining} 
uses a shared network block
to create an initial representation which is then evaluated with multiple
classification heads (see Figure~\ref{fig:mhct-arch}). With labeled data, these
classification heads (and the shared block) are then all updated as normal.
However, when an unlabeled sample is selected, the most confident pseudo-label is
assigned based on a consensus between a majority of the heads. 
For better generalization 
MHCT uses data augmentation.

Methods that require training multiple models have the obvious drawback of the additional time and memory resources required to train those models.  Often these methods are employing iterative re-training to learn from the pseudo-labels they produce, so the computational cost of that re-training is multiplied by the number of models used.  Additionally, multi-model methods require that each of the models has captured some independent information or pattern from the data.  If all of the models trained are more or less identical, then it is not useful at all to have trained multiple models.  In cases where single model approaches are already very performant it can be challenging to find or train multiple models with competitive performance that capture independent information.

\subsubsection{Joint-Optimization Selection}\label{sec:joint-optimization}
Unlike other methods of pseudo-labeling, \emph{Meta Pseudo Label} (MPL) by~\citet{MetaPL} treats pseudo-label
assignment as an optimization problem, with the specific assignment of pseudo-labels jointly optimized with the  model
itself. 
The framework consists of two models, a student and a teacher, which share an architecture. 
The teacher receives unlabeled examples and
assigns pseudo-labels to them. The student receives the example pseudo-label
pair, and performs a gradient update. 
The student then is evaluated on a
labeled sample. The teacher receives the student performance on the labeled
sample, and performs a weight update to make better pseudo-labels.  
Note that
the student is \textit{only} trained on pseudo-labels and the teacher is
\textit{only} trained on the student's response on labeled data. In this way,
the teacher learns an approximation of the optimal PL policy with respect to
optimizing the student performance on the labeled set, and the student learns to
approximate the PL function described by the teacher.  

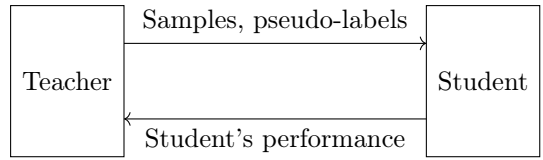
\begin{figure}[!ht]
    \centering
    \vspace{0.5em}
    \begin{tikzpicture}
        [repr/.style={draw=black,minimum size=8mm},
         every node/.style={execute at begin node=\small}]

        \draw[repr] (-0.5,-1) rectangle node[midway]{Teacher} (1, 1);

        \begin{scope}[shift={(5,0)}]
            \draw[repr] (0,-1) rectangle node[midway]{Student} (1.5, 1);
        \end{scope}

        \draw [->] (1, 0.5) -- node[midway,above]{ Samples, pseudo-labels } (5, 0.5);
        \draw [<-] (1,-0.5) -- node[midway,below]{ Student's performance } (5,-0.5);
    \end{tikzpicture}
    \vspace{0.5em}
    \caption{
        A figure describing Meta Pseudo Label's overall architecture. Adapted
        from~\citet{MetaPL}.
    }\label{fig:mpl-arch}
\end{figure}

A simplified overview of the MPL method is shown in Figure~\ref{fig:mpl-arch}.  First, the teacher $f_{\theta_T}$ is fixed and the student is trying to learn a better model by aligning its predictions to pseudo-labeled batches $(X_u, f_{\theta_T}(X_u))$.
On a pseudo-labeled batch, the student optimizes its parameters using 
\begin{equation}
\theta_{S}^{\text{PL}} \in 
\argmin_{\theta_S}\ell\big(\argmax_{i} f_{\theta_T}(X_u)_i, f_{\theta_S}(X_u)\big)\,.
\end{equation}

The hope is that updated student parameters $\theta_S^{\text{PL}}$ will behave well on the labeled data $L = (X_L, Y_L) = \left\{(x_i, y_i)\right\}_{i=1}^{m}$; i.e., the loss 
$\mathcal{L}_L\left(\theta_{S}^{\text{PL}}\left(\theta_T\right)\right) = 
\ell\left( Y_L, f_{\theta_S^{\text{PL}}}(X_L)\right)$. $\mathcal{L}\left(\theta_{S}^{\text{PL}}\left(\theta_T\right)\right)$ is defined such that the dependence between $\theta_S$ and $\theta_T$ is explicit.  The loss that the student suffers on $\mathcal{L}_L\left(\theta_{S}^{\text{PL}}\left(\theta_T\right)\right)$ is a function of $\theta_{T}$.   Exploiting this dependence between $\theta_S$ and $\theta_T$.
and making a similar notational convention for the unlabeled batch $\mathcal{L}_u\left(\theta_T, \theta_S\right) = \ell\left(\argmax_{i} f_{\theta_T}(X_u)_i, f_{\theta_S}(X_u)\right)$,
then one can further optimize $\theta_T$ as a function of the performance of $\theta_S$:

\begin{equation}
\begin{aligned}
\text{min}_{\theta_T} & \mathcal{L}_L\left(\theta_{S}^{\text{PL}}\left(\theta_T\right)\right)\,, & \\
& \textit{where} & \\
&\theta_{S}^{\text{PL}}(\theta_T) \in \argmin_{\theta_S} \mathcal{L}_u\left(\theta_T, \theta_S\right)\,.\\
\end{aligned}
\end{equation}

However, because the dependency between $\theta_T$ and $\theta_S$ is complicated, a practical approximation
is obtained via 
$\theta_{S}^{\text{PL}} \approx \theta_S - \eta_S\cdot\nabla_{\theta_S} \mathcal{L}_u\left(\theta_T, \theta_S\right)$ 
which leads to the practical teacher objective:
\begin{equation}\label{eq:MPL:teacher}
\min_{\theta_T} \mathcal{L}_L\left(\theta_S - \eta_S\cdot\nabla_{\theta_S} \mathcal{L}_u\left(\theta_T, \theta_S\right)\right)\,.
\end{equation}

Whereas MPL optimizes a labeler to provide good labels to unlabeled points, Meta-Semi~\citep{Meta-Semi} optimizes which pseudo-labels to use for training.  Particularly, if inclusion of a pseudo-label causes a decrease in loss for the student on a labeled batch then the pseudo-labeled example is included for training and otherwise it is omitted.  Similarly, Bi-Level Optimization for Pseudo-Labeling Based Semi-Supervised Learning (BLOPL)~\citep{BLOPL} considers pseudo-labels as latent variables of which the learner's weights are a function.  Rather than optimizing the presence or absence of pseudo-labels BLOPL optimizes soft pseudo-labels which are always included to result in a good learned model.  Reinforcement Learning Guided Semi-Supervised Learning~\citep{RLGSSL} poses this optimization of pseudo labels as a reinforcement learning problem and learns a policy to provide good soft pseudo-labels to the learner.  

\emph{Meta Co-Training} (MCT)~\citep{MetaCotraining} combines the ideas of Meta Pseudo Labels~\citep{MetaPL} and Co-Training~\citep{cotraining}.  Meta Co-Training introduces a novel bi-level optimization over multiple views to determine optimal assignment of pseudo labels for co-training.  At the lower level of the optimization the student parameters $\theta_S$ are optimized as a function of the teacher parameters $\theta_T$:
\begin{equation}\label{eq:LU}
\mathcal{L}_u\left(\theta_T, \theta_S\right) = \ell\left(\argmax_{\xi} f_{\theta_T}(X_u)|_{\xi}, f_{\theta_S}(X_u)\right)
\end{equation}
\begin{equation}\label{eq:SPL}
\theta_S' = \theta_{S}^{\text{PL}}(\theta_T) \in 
\argmin_{\theta_S}\mathcal{L}_u\left(\theta_T, \theta_S\right)\,.
\end{equation}

At the upper level (Equation \ref{eq:MCTTprime}) the teacher parameters are optimized to improve the student:
\begin{equation}\label{eq:LL}
\mathcal{L}_L\left(\theta_S'\right) = \ell\left( Y_L, f_{\theta_S'}(X_L)\right)
\end{equation}
\begin{equation}\label{eq:MCTTprime}
\theta_T' = \argmin_{\theta_T} \displaystyle\mathcal{L}_L\left(\theta_S'\right)\,.
\end{equation}

All together the objective of MCT from the perspective of the current view model as the teacher is
\begin{equation}\label{eq:MCT:loss}
\min_{\theta_T} \mathcal{L}_u\left(\theta_T, \theta_S\right) + \mathcal{L}_L\left(\theta_S'\right)\,.
\end{equation}

The full objective of MCT is then:

\begin{equation}\label{eq:MCT:joint}
    \min_{\theta_1, \theta_2} \mathcal{L}_u\left(\theta_1, \theta_2\right) + \mathcal{L}_L\left(\theta_2'\right)\ + \mathcal{L}_u\left(\theta_2, \theta_1\right) + \mathcal{L}_L\left(\theta_1'\right)\,
\end{equation}
In this formulation each view is utilized by a single model which serves as both the student for that view and the teacher for the other view.  The student is optimized to replicate the pseudo-labels assigned by its teacher given the corresponding instance as input.  The teacher is optimized to produce pseudo-labels that improve student performance on a held-out labeled set.  This formulation is a natural extension of the ideas of Meta Pseudo Labels to two views.  They further propose using pre-trained models to construct two views from a single instance in the dataset.  This method of view construction compresses the input significantly in the case of image classification which mitigates the impact of the expensive computational nature of multi-model semi-supervised approaches, though it requires the existence of such models.

Each of~\citet{MetaPL, MetaCotraining, BLOPL, RLGSSL, Meta-Semi} pose a bi-level optimization in which the way in which pseudo-labels are provided to the learner is optimized to create a performant learner.  Any approach based on a bi-level optimization has the potential to be expensive to compute.  In the case of MPL it takes a million optimization steps on the CIFAR-10 dataset to produce results only marginally better than Unsupervised Data Augmentation~\citep{UDA}.  One of the major drawbacks of MPL and similar methods are their long training times.

\section{Unsupervised and Self-supervised Regimes}\label{sec:pre-training}
While pseudo-labeling is usually presented in the context of
semi-supervised training regimes, there are very close analogues to PL in the
domain of unsupervised learning. Specifically, unsupervised consistency
regularization approaches show significant similarities with pseudo-labeling
approaches as they tend to assign labels to points even if these labels have no relation to any ground truth.
For example, EsViT~\citep{EsViT}, DINO~\citep{DINO}, BYOL~\citep{BYOL}, and SwAV~\citep{SwAV} all use some form of label assignment as a representation learning \textit{pretext task}, i.e.,~a task which prepares
a model for transfer learning to a downstream task~\citep{SSSurvey}. Within this section, we will discuss descriminative self-supervised learning which operate as CR methods (Section~\ref{sec:consistency-regularization}), and response-based knowledge distillation (Section~\ref{sec:knowledge-distillation}) both of which are forms of pseudo-labeling.

\subsection{Consistency Regularization}\label{sec:consistency-regularization}

It is perhaps surprising to consider pseudo-labeling as a definition for unsupervised learning settings, where traditional, fixed class labels are not available. However, such probability vectors are used in the training of particularly descriminative self-supervised learning.  These approaches such as DINO~\citep{DINO} and are incorporated into the loss function of
SWaV~\citep{SwAV}, EsViT~\citep{EsViT}, and Deep Cluster~\citep{DeepCluster} and their effectiveness is explained as a form of consistency regularization. 

These approaches all result in assigning instances to a particular cluster or class which has no defined meaning, but is rather inferred from the data in order to accomplish the minimization of the self-supervised loss.  The goal of the clustering or classification is clear: to place images which share an augmentation relationship into the same cluster or class, and optionally a different class from all other images.  The partition of the latent space however is inferred, and it is also fuzzy as membership probabilities may not be one-hot.  Thus, these techniques clearly satisfy our definition.  

\begin{figure}[ht]
    \centering
    \begin{tikzpicture}
        [repr/.style={circle,draw=black,minimum size=8mm}]

        \node[repr] (original) at (0,0) {\(\bx\)};

        \node[repr] (aug-1) at (2,  1) {\(\tilde\bx_1\)};
        \node[repr] (aug-2) at (2, -1) {\(\tilde\bx_2\)};

        \node[repr] (latent-1) at (4,  1) {\(\bz_1\)};
        \node[repr] (latent-2) at (4, -1) {\(\bz_2\)};


        \draw [->] (original.north) to[out= 90,in=180] node[swap,auto]  {\small{\(t \sim \mathcal T\)}} (aug-1.west);
        \draw [->] (original.south) to[out=270,in=180] node[above,auto] {\small{\(t \sim \mathcal T\)}} (aug-2.west);

        \draw [->,blue,decorate,decoration={snake,amplitude=.5mm}] (aug-1) -- node[midway,below]{\small\(\func{f}\)} (latent-1);
        \draw [->,blue,decorate,decoration={snake,amplitude=.5mm}] (aug-2) -- node[midway,above]{\small\(\func{f}\)} (latent-2);
        

        \draw [|-|,orange] (latent-1) -- (latent-2)
            node [right,align=center,midway] {\small{Comparison}};
    \end{tikzpicture}
    \caption{
        A chart showing 
        how general pairwise 
        consistency regularization 
        works.
        Generally, samples are drawn from \(\bx \sim \mathcal D\), where a
        series of transformations \(t\sim\mathcal T\) transforms \(\bx\) into a
        pair of unequally augmented samples \(\tilde\bx_1,\tilde\bx_2\). This
        pair of samples is then projected into a latent representation by some
        projection function \(\func{f}\) resulting in latent vectors
        \(\bz_1,\bz_2\). Typically during consistency regularization, the loss
        function will be structured to minimize the distance from
        \(\bz_1,\bz_2\) from the same sample \(\bx\) and maximize the distance
        from other sample representations within a minibatch.
    }\label{fig:unsupervised-cr}
\end{figure}
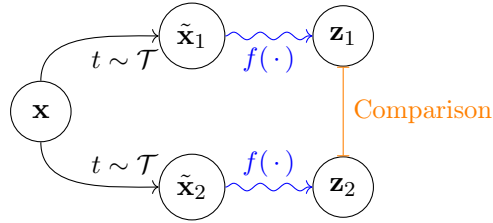

All these frameworks are ``unsupervised'' in the sense that they do not rely on
ground-truth labels, but they truly belong to the category of 
SSL
as they are trained on pretext tasks and are typically applied to semi-supervised tasks.
A common adaptation (and indeed the standard metric by
which representational performance is judged) is training a linear layer on top
of the learned representations~\citep{kolesnikov_revisiting_2019,SimCLR}, with
the intent of performing well in settings where few labels are available. A
common theme in solutions in this space is stochastically augmenting a pair of
samples and ensuring that the learned representation of those samples is
similar to each other and different to all other samples in the batch, for
example as in
Figure~\ref{fig:unsupervised-cr}; see, e.g.,~\citet{SwAV,EsViT,DeepCluster,UDA}. 
In this
sense, these are self-supervised learning techniques that technically use
pseudo-labels.  Such self-supervised learning approaches are referred to as ``discriminative''
as they are designed to discriminate between different ground truth signals by assigning different (pseudo) labels to them. 
Because they do not predict labels that correspond to any specific important function assignment, they are functionally indistinguishable from
techniques like SimCLR~\citep{SimCLR}, and BYOL~\citep{BYOL}, which do not make
use of any sort of class assignment (pseudo or otherwise).

In a stricter sense, there are approaches such as 
\textit{Scaling
Open-Vocabulary Object Detection}~\citep{OVOD} which make use of actual
predicted class labels that are relevant to a future task. This is not
a classification task but rather an object detection task, however the
CLIP 
model is used to give weak supervision (pseudo-labels) for
bounding boxes and this weak supervision is used as a pre-training step. This
kind of weak supervision is common in open-vocabulary settings such as that
used to train CLIP, however in the case of CLIP training that weak supervision
came from humans not any model outputs.

In order to perform consistency regularization the training algorithm has to define between which instances model predictions should be consistent.  Typically, consistency is enforced between augmented versions of the same image, or between different representations of the same data point.  Effective consistency regularization relies on being able to define transformations to which the content of an image is invariant.  This may pose a limitation to specialized data domains for which standard data augmentation techniques to not apply.  Another possible failure for consistency regularization is representation collapse.  It is possible that a model learns to predict the same output for every input which leads to very good consistency but very poor representation quality.

\subsection{Knowledge Distillation}\label{sec:knowledge-distillation}

Knowledge distillation~\citep{KD-hinton} describes the process of transferring the knowledge of a larger network into a smaller one.  The significant attention brought to this technique has yielded a variety of techniques~\citep{KD-Survey}.  In this assessment we are concerned with \textit{response-based} knowledge distillation; i.e., techniques for knowledge distillation that transfer knowledge through pseudo-labels assigned by the larger model.  For a more comprehensive treatment of knowledge distillation 
see~\citep{KD-Survey}.  

The 
formulation in~\citet{KD-hinton} 
is a response-based framework which is semi-supervised.  The original larger model is trained in a supervised way on a classification dataset.  The process of distilling knowledge into the smaller model is semi-supervised in general, although if only the model and instances are present without labels the distillation can still occur with only self-supervision.  The model is trained to minimize a loss function that is composed of a supervised term measuring the disagreement with the true labels of the examples, and another loss term which measures the disagreement with the prediction made by the larger model. This second term is clearly making use of PL.

As mentioned 
in~\citet{DINO}, the DINO framework for self-supervised learning is interpreted as a framework for knowledge distillation between a student model and a mean teacher (a teacher model which is an average of the student model's weights).  This is self-distillation with \textit{no labels} which is entirely based on PL.  DINO can be seen as a special case of the \textit{replace one branch}~\citep{KD-RoB} approach to knowledge distillation with a slight modification of including a mean teacher.  In fact, this is exactly how the smaller published DINO models are created.  Rather than training the models from scratch, the smaller architectures' parameters are trained by holding the parameters of the larger model constant and distilling their knowledge from them into the smaller model using the same DINO framework which was used to train the larger model.  This approach is broadly applicable to the above self-supervised frameworks and others which may not utilize PL but have a similar two-branch training structure.

In all cases a fuzzy partition parameterized by a neural network is inferred from data and available at the beginning of the distillation process.  This is the teacher model, or the model which is being distilled.  In the case of Hinton's original formulation this fuzzy partition clearly corresponds to division of the input space into classes informed by real labels.  It is also possible to incorporate real labels in the distillation process in that formulation, however pseudo-labels are still used in the training process.  in fact, the labels generated by the model and given to the student model are identical to those utilized in traditional pseudo-labeling approaches such as~\citet{PL}.  In the case of DINO and RoB the labels do not necessarily correspond to any ground truth quantity, but the loss function is identical.  DINO distillation and RoB also define fuzzy partitions of the input space into categories and the goal of learning in the knowledge distillation regime is the same: minimize the average cross entropy over a set of unlabeled instances and their associated pseudo-labels between the student and the teacher.

\section{Commonalities Between Discussed Methods}\label{sec:commonalities}

We have covered a variety of ways in which pseudo-labeling is applied in computer vision algorithms.  We have connected these ideas with our  interpretation of pseudo-labels.  In this section we identify commonalities between the methods that we have discussed so far. This will then illuminate the directions for future work, which we present in the next section.

\subsection{Filtering the Data}
One of the main arguments presented in CLIP~\citep{CLIP} is the extensive dataset filtering methodology, and even subsequent works such as ALIGN~\citep{ALIGN} still perform some dataset filtering.  Such rigorous dataset construction seems to be the norm now with the LAION dataset using the CLIP pre-trained model to filter out noisy image-caption pairs~\citep{LAION5B}, and~\citet{demystifyingCLIP} argue the dataset curation was an essential aspect of CLIP's success.  Proprietary pre-filtered datasets such as JFT and IG-1B serve as an integral component in training state-of-the-art models' self-supervised and semi-supervised learning~\citep{MetaPL, CLIP, DINO, NST}. In knowledge distillation, maintaining the full model's performance requires some access to the training data; without it, we must resort to techniques like {data-free} knowledge distillation~\citep{DataFreeKD} or reconstructing a similar dataset from open data such as LAION-5B~\citep{LAION5B}. Thus, strategic dataset filtering emerges as a key enabler for effective pseudo-labeling.

\subsection{Establishing a Curriculum}

Curriculum learning~\citep{Curriculum} is an approach that was originally
proposed for supervised learning, but it is often used as the justification for
metric-based selection procedures for PL (as discussed in Section~\ref{sec:curriculum-learning}). 
Outside SSL, curriculum approaches have received a small amount of
attention recently in knowledge distillation literature with curricula
inspired by the training trajectory of the heavier
model~\citep{RouteConstrainedKD}, and by the temperature used during
training~\citep{CurriculumTemperatureKD}.  Curriculum learning has also been
applied in self-supervised learning to learn representations that are robust to
spurious correlations~\citep{FairSSL}. The curricula proposed for knowledge
distillation and self-supervised learning are practically interchangeable when
using an appropriate framework such as RoB~\citep{KD-RoB} and DINO~\citep{DINO}
respectively.  Though the connection is perhaps looser, the choice of a
curriculum is very important for good self-training and co-training
performance.  There is similarly no practical reason the curricula proposed for
self-training could not be applied to knowledge distillation or self-training
and vice-versa.

\subsection{Data Augmentation Strategies}

Data augmentations typically perturb the input instances without altering their
ground truth assignments, and strategies for this are varied and
domain-dependent. All forms of data augmentation can be seen as providing
pseudo-labels, as the original sample label is assigned to a new (augmented)
sample. Rather than communicating the prediction of some machine learning
model, these pseudo-labels instill an inductive bias. Outside of limited cases
such as horizontal flips of most images (which can be known to not change the
underlying class), augmentations such as MixUp~\citep{MixUp},
the *-Match family~\citep{MixMatch, FixMatch, ReMixMatch, FlexMatch}, or even RandAugment~\citep{RandAug} can
meaningfully change or obscure class information. This is critical as they define the invariants the model is to learn and makes the most of limited supervised data.

\subsection{Soft vs.~Hard Pseudo-Labels}

The dominant strategy is in line with the spirit of EM, where choosing hard
labels extends the margin of the decision boundary. Some works report better
performance when using soft labels~\cite{UDA}, 
but other works take precautions to avoid
trivial solutions when using soft labels~\cite{PENCIL}; not to mention the challenges of using soft labels (and thus weaker
gradients) for larger model architectures. Conversely, hard labels can be sampled from
a discrete distribution described by the output vector of a
model yielding an assignment less aligned with the ground truth, but which
strikes a
desirable learning process exploration / exploitation trade-off (e.g.,~\citet{MetaPL}).
Interestingly, this hard-label choice is not nearly as popular in the self-supervised or knowledge distillation areas where the argument usually follows from an argument centered around smoothing a distribution rather than performing EM.  Despite this, such techniques typically make use of a large temperature parameter to ``harden'' the label output by the teacher~\citep{SimCLR, DINO, EsViT, KD-hinton, PreparingKD, CurriculumTemperatureKD}.

\subsection{Updating the Teacher}

There are a variety of strategies that are employed to update the teacher model from which knowledge is being transferred.  In self-supervised settings a teacher is usually a model updated as the EMA of the student's weights from the last several steps~\citep{BYOL, DINO, EsViT}. Conversely, in semi-supervised learning the teacher's weights are usually the product of a separate optimization---either through meta-learning~\citep{MetaPL} or a previous weakly-supervised training iteration~\citep{NST, SelfTrain, cotraining, cocomittee, BetterTriTraining, MHCotraining, TCSSL}.  Acknowledging the weakness of the teacher's labels within the knowledge distillation framework, one can even imagine altering teacher predictions to produce a better student~\citep{PreparingKD}.  In all pseudo-labeling cases it is useful to consider how (if at all) the teacher should be updated.

\subsection{Self-Supervised Regularization}

In practice, there are far more unlabeled samples than labeled samples: for example, open source NLP/CV datasets contain billions of un- or weakly-curated samples~\citep{LAION5B, redpajama}. 
Typically, representations are learned first on unsupervised data and then are transferred to a downstream task.  In the case of full-parameter finetuning, however, it is possible for a network to `forget' valuable information learned during pre-training while it is being finetuned. An important type
of regularization may be to include a semi-supervised objective in the loss
such as NT-Xent~\citep{SimCLR} or the BYOL~\citep{BYOL} loss function; preventing
a representation space from collapsing or to allow learning a
representation space jointly with the classification objective.

\section{Open Directions}\label{sec:future}

Having identified the ways in which these different algorithms which use pseudo-labels are similar, we are now prepared to discuss unexplored directions.  Below we outline what we believe to be some of the most promising open directions 
when one tries to exploit these commonalities between the different forms of pseudo-labeling.

\subsection{Dataset Filtering}

There is significant focus on dataset filtering in the self-supervised learning literature as we discussed in the previous section, however this remains relatively unexplored in semi-supervised learning and knowledge distillation.  We believe dataset subsetting has the potential to expedite knowledge distillation, and that it has the potential to benefit open-set semi-supervised learning algorithms which are vulnerable to class imbalance.  

In semi-supervised learning the typical method of evaluating the performance of algorithms is to hold out a subset of the labels in a dataset, but this yields an idealized unlabeled distribution.  Semi-supervised learning algorithms are also vulnerable to differences in distribution between the unlabeled and labeled sets, and they are vulnerable to class imbalance in the unlabeled data.  Both of these cases seem likely when using unlabeled data collected from the wild.  It seems likely that if some method of intelligent filtering were applied to a large unlabeled and uncurated dataset a more realistic evaluation scenario and improved model performance would result.

\subsection{Establishing a Useful Curriculum}

Establishing an effective curriculum is an open direction for all of these pseudo-labeling methods.  It seems like some curriculum is more useful than none at all, but it is unclear what properties an ideal curriculum has.  In Section~\ref{sec:curriculum-learning} we identified the fact that neural networks generally produce uncalibrated uncertainty estimates which is a barrier to curricula based on model confidence.  We see the application of uncertainty quantification or calibration methods to the construction of such curricula as a promising future direction.

\subsection{Direct Initialization of Models}

While we can initialize models naively from some initialization distribution, a large number of pre-trained ``foundation'' models for vision have been produced which can be compressed using low-rank compression~\cite{lowrank, HALOC}, neuron clustering~\cite{k-meansNeuronClustering}, or neuron selection~\cite{directInit}.  Such techniques have the potential to expedite training of the student, or lower the cost of assigning a pseudo-label with the teacher.

\subsection{Self-Supervised Regularization of Semi-Supervised Learning}

In semi-supervised and knowledge distillation there is a risk of representation collapse of the student.  To prevent this collapse we can include self-supervised loss terms in the optimization to prevent this collapse.  One recent method that does this is \textit{self-supervised semi-supervised learning}~\citep{ReMixMatch} in which a model is trained to jointly optimize a supervised and unsupervised loss.  Despite the simplicity they are able to improve on supervised baselines. Their approach could easily be extended by optimizing their (relatively simple) losses in a self-training setting, in a knowledge distillation setting, or by simply utilizing more modern self-supervised methods.

\subsection{Meta Learning to Determine Pseudo-Label Assignment}

In Section~\ref{sec:multimodel} we identified several of a class of methods for semi-supervised learning which  proposed formulating the pseudo-label assignment as a meta-learning problem.  To our knowledge meta-learning has not been applied in this way to either knowledge distillation or self-supervised learning.  Furthermore, one of those works~\citep{RLGSSL} proposed a reinforcement learning formulation of the pseudo-labeling problem.  Meta learning response-based distillation targets or discriminative self-supervised targets and reinforcement learning pseudo-label assignment for semi-supervised learning remain exciting directions.

\section{Conclusion}\label{sec:conclusion}
In Section~\ref{sec:intro} we gave an overview of theoretical motivations for pseudo-labeling.  We provided a taxonomy of methods for pseudo-labeling in these areas and discussed their relation to the important motivating ideas of sample scheduling, label propagation, weak supervision, consistency regularization, multi-model learning, and knowledge distillation.  This taxonomy was shown in Figure~\ref{fig:family-tree}. We provided a comparison of methods applied to semi-supervised image classification; these comparative results were shown in Table~\ref{tab:pl_benchmarks}.  In Section~\ref{sec:prelims} we introduced a definition for pseudo-labels.  We showed how this definition allows for a unified interpretation of methods from semi-supervised learning and self-supervised learning.  In Section~\ref{sec:base-pl} we presented many different methods for semi-supervised learning that make use of pseudo-labels.  In Section~\ref{sec:pre-training} we showed how pseudo-labels are used in unsupervised and self-supervised ways.  We discussed commonalities that we observed between different types of pseudo-labeling algorithms in Section~\ref{sec:commonalities}.  In Section~\ref{sec:future} we discussed open directions that were illuminated by our definition of pseudo-labels.  
Finally, in Table~\ref{tab:family-map} in Appendix~\ref{appendix:tables} we provide a glossary of the different techniques that are shown in the taxonomy of Figure~\ref{fig:family-tree}. 
Table~\ref{tab:family-map} provides the paper that proposed each technique and the section in which they appear in this review.

We find that there are many exciting new opportunities for research.  The prospect of applying dataset subsetting methods on data collected from the wild is particularly exciting, and so is the application of reinforcement learning algorithms/bi-level optimization frameworks and uncertainty quantification.  We are excited to see how new applications of pseudo-labels continue to shape the computer vision landscape.

\section*{Acknowledgments}
This material is based upon work supported by the U.S.~National Science Foundation under Grant No.~RISE-2019758. The authors would like to thank Dr.~Kerri Cahoy for providing valuable input on an earlier version of this work.

\appendix

\section{Reference Table for the Literature Shown in Figure~\ref{fig:family-tree}}~\label{appendix:tables}
Table~\ref{tab:family-map} has a mapping between the techniques that appear in Figure~\ref{fig:family-tree} (first column), with the appropriate references that are associated with the corresponding techniques (second column), and finally the section in this work where that technique is discussed.

  \begin{xltabular}{\textwidth}{SlV{2}l|X}
  \toprule
    \textbf{Label} & \textbf{Reference} & \textbf{Section} \\
    \midrule

      BLOPL & \cite{BLOPL} & \ref{sec:multimodel} \\
    BYOL   & \cite{BYOL}   & \ref{sec:consistency-regularization} \\
    Class Introspection & \cite{ClassIntrospection} & \ref{sec:sample-scheduling} \\
    Co-training & \cite{cotraining} & \ref{sec:multimodel} \\
    Cocommittee & \cite{cocomittee} & \ref{sec:multimodel} \\
    DINO & \cite{DINO} & \ref{sec:knowledge-distillation} \\
    DeepCluster & \cite{DeepCluster}  & \ref{sec:consistency-regularization} \\
    Dem.~Colearning & \cite{democraticcolearning} & \ref{sec:multimodel} \\
    DL from NL & \cite{DLfromNL} & \ref{sec:sample-scheduling} \\
    EsViT   & \cite{EsViT}   & \ref{sec:consistency-regularization} \\
    FixMatch & \cite{FixMatch} & 
    \ref{sec:sample-scheduling} \\
    FlexMatch & \cite{FlexMatch} & \ref{sec:sample-scheduling} \\
    ILI & \cite{ILI} & 
   \ref{sec:sample-scheduling}\\ 
    IDN & \cite{InstanceDependentNoise} & \ref{sec:sample-scheduling} \\
    Knowledge Distillation & \cite{KD-hinton} & \ref{sec:knowledge-distillation} \\
    LabelPropPL & \cite{LabelPropPL} & \ref{sec:metric-based} \\
    MHCT & \cite{MHCotraining} & \ref{sec:multimodel} \\
    MMF & \cite{MMF} & \ref{sec:metric-based} \\
    Meta Co-training & \cite{MetaCotraining} & \ref{sec:multimodel} \\
    MetaPL & \cite{MetaPL} & \ref{sec:sample-scheduling} \\ 
    MixMatch & \cite{MixMatch} & 
    \ref{sec:sample-scheduling} \\ 
    Noisy Student & \cite{NST} & \ref{sec:multimodel} \\
    OVOD & \cite{OVOD}  & \ref{sec:consistency-regularization} \\
    PENCIL & \cite{PENCIL} & \ref{sec:sample-scheduling} \\
    PL (2013) & \cite{PL} & \ref{sec:base-pl} \\
    ProSub & \cite{ProSub} & \ref{sec:sample-scheduling} \\
    ROSSEL & \cite{ROSSEL} & \ref{sec:multimodel} \\
    RLGSSL & \cite{RLGSSL} & \ref{sec:multimodel} \\
    ReMixMatch & \cite{ReMixMatch} & 
    \ref{sec:sample-scheduling} \\
    Replace One Branch & \cite{KD-RoB} & \ref{sec:knowledge-distillation} \\
    SeFOSS & \cite{SeFOSS} & \ref{sec:sample-scheduling} \\
    SwAV & \cite{SwAV}  & \ref{sec:consistency-regularization} \\
    TCSSL & \cite{TCSSL} & 
    \ref{sec:sample-scheduling}\\
    TriTraining & \cite{TriTraining} & \ref{sec:multimodel} \\
    UDA & \cite{UDA}  & \ref{sec:consistency-regularization} \\
    UPS & \cite{InDefenseOfPL} & \ref{sec:sample-scheduling} \\
    VAT   & \cite{VAT}   & \ref{sec:consistency-regularization} \\

    \bottomrule
    
    \caption{Figure~\ref{fig:family-tree} Label Map}\label{tab:family-map}\\
  \end{xltabular}


\vskip 0.2in
\bibliography{references}

\begin{thebibliography}{}

\bibitem [\protect \citeauthoryear {%
Amini%
, Feofanov%
, Pauletto%
, Devijver%
\BCBL {}\ \BBA {} Maximov%
}{%
Amini%
\ \protect \BOthers {.}}{%
{\protect \APACyear {2022}}%
}]{%
SelfTrainingASurvey}
\APACinsertmetastar {%
SelfTrainingASurvey}%
\begin{APACrefauthors}%
Amini, M.%
, Feofanov, V.%
, Pauletto, L.%
, Devijver, E.%
\BCBL {}\ \BBA {} Maximov, Y.%
\end{APACrefauthors}%
\unskip\
\newblock
\APACrefYearMonthDay{2022}{}{}.
\newblock
{\BBOQ}\APACrefatitle {{Self-Training: {A} Survey}} {{Self-Training: {A} Survey}}.{\BBCQ}
\newblock
\APACjournalVolNumPages{CoRR}{abs/2202.12040}{}{}.
\newblock
\begin{APACrefURL} \url{https://arxiv.org/abs/2202.12040} \end{APACrefURL}
\PrintBackRefs{\CurrentBib}

\bibitem [\protect \citeauthoryear {%
Angluin%
\ \BBA {} Laird%
}{%
Angluin%
\ \BBA {} Laird%
}{%
{\protect \APACyear {1987}}%
}]{%
PAC:Noise:Angluin}
\APACinsertmetastar {%
PAC:Noise:Angluin}%
\begin{APACrefauthors}%
Angluin, D.%
\BCBT {}\ \BBA {} Laird, P\BPBI D.%
\end{APACrefauthors}%
\unskip\
\newblock
\APACrefYearMonthDay{1987}{}{}.
\newblock
{\BBOQ}\APACrefatitle {{Learning From Noisy Examples}} {{Learning From Noisy Examples}}.{\BBCQ}
\newblock
\APACjournalVolNumPages{Mach. Learn.}{2}{4}{343--370}.
\newblock
\begin{APACrefURL} \url{https://doi.org/10.1007/BF00116829} \end{APACrefURL}
\newblock
\begin{APACrefDOI} \doi{10.1007/BF00116829} \end{APACrefDOI}
\PrintBackRefs{\CurrentBib}

\bibitem [\protect \citeauthoryear {%
Arazo%
, Ortego%
, Albert%
, O'Connor%
\BCBL {}\ \BBA {} McGuinness%
}{%
Arazo%
\ \protect \BOthers {.}}{%
{\protect \APACyear {2020}}%
}]{%
PLandCB}
\APACinsertmetastar {%
PLandCB}%
\begin{APACrefauthors}%
Arazo, E.%
, Ortego, D.%
, Albert, P.%
, O'Connor, N\BPBI E.%
\BCBL {}\ \BBA {} McGuinness, K.%
\end{APACrefauthors}%
\unskip\
\newblock
\APACrefYearMonthDay{2020}{}{}.
\newblock
{\BBOQ}\APACrefatitle {{Pseudo-Labeling and Confirmation Bias in Deep Semi-Supervised Learning}} {{Pseudo-Labeling and Confirmation Bias in Deep Semi-Supervised Learning}}.{\BBCQ}
\newblock
\BIn{} \APACrefbtitle {2020 International Joint Conference on Neural Networks, {IJCNN} 2020, Glasgow, United Kingdom, July 19-24, 2020} {2020 international joint conference on neural networks, {IJCNN} 2020, glasgow, united kingdom, july 19-24, 2020}\ (\BPGS\ 1--8).
\newblock
\APACaddressPublisher{}{{IEEE}}.
\newblock
\begin{APACrefURL} \url{https://doi.org/10.1109/IJCNN48605.2020.9207304} \end{APACrefURL}
\newblock
\begin{APACrefDOI} \doi{10.1109/IJCNN48605.2020.9207304} \end{APACrefDOI}
\PrintBackRefs{\CurrentBib}

\bibitem [\protect \citeauthoryear {%
Balestriero%
\ \protect \BOthers {.}}{%
Balestriero%
\ \protect \BOthers {.}}{%
{\protect \APACyear {2023}}%
}]{%
balestriero_cookbook_2023}
\APACinsertmetastar {%
balestriero_cookbook_2023}%
\begin{APACrefauthors}%
Balestriero, R.%
, Ibrahim, M.%
, Sobal, V.%
, Morcos, A.%
, Shekhar, S.%
, Goldstein, T.%
\BDBL {}Goldblum, M.%
\end{APACrefauthors}%
\unskip\
\newblock
\APACrefYearMonthDay{2023}{}{}.
\newblock
{\BBOQ}\APACrefatitle {{A Cookbook of Self-Supervised Learning}} {{A Cookbook of Self-Supervised Learning}}.{\BBCQ}
\newblock
\APACjournalVolNumPages{CoRR}{abs/2304.12210}{}{}.
\newblock
\begin{APACrefURL} \url{https://doi.org/10.48550/arXiv.2304.12210} \end{APACrefURL}
\newblock
\begin{APACrefDOI} \doi{10.48550/ARXIV.2304.12210} \end{APACrefDOI}
\PrintBackRefs{\CurrentBib}

\bibitem [\protect \citeauthoryear {%
Barcina-Blanco%
, Lobo%
, Garcia-Bringas%
\BCBL {}\ \BBA {} Ser%
}{%
Barcina-Blanco%
\ \protect \BOthers {.}}{%
{\protect \APACyear {2024}}%
}]{%
OSR_Survey}
\APACinsertmetastar {%
OSR_Survey}%
\begin{APACrefauthors}%
Barcina-Blanco, M.%
, Lobo, J\BPBI L.%
, Garcia-Bringas, P.%
\BCBL {}\ \BBA {} Ser, J\BPBI D.%
\end{APACrefauthors}%
\unskip\
\newblock
\APACrefYearMonthDay{2024}{}{}.
\newblock
\APACrefbtitle {Managing the unknown: a survey on Open Set Recognition and tangential areas.} {Managing the unknown: a survey on open set recognition and tangential areas.}
\newblock
\begin{APACrefURL} \url{https://arxiv.org/abs/2312.08785} \end{APACrefURL}
\PrintBackRefs{\CurrentBib}

\bibitem [\protect \citeauthoryear {%
Bengio%
, Louradour%
, Collobert%
\BCBL {}\ \BBA {} Weston%
}{%
Bengio%
\ \protect \BOthers {.}}{%
{\protect \APACyear {2009}}%
}]{%
Curriculum}
\APACinsertmetastar {%
Curriculum}%
\begin{APACrefauthors}%
Bengio, Y.%
, Louradour, J.%
, Collobert, R.%
\BCBL {}\ \BBA {} Weston, J.%
\end{APACrefauthors}%
\unskip\
\newblock
\APACrefYearMonthDay{2009}{}{}.
\newblock
{\BBOQ}\APACrefatitle {{Curriculum learning}} {{Curriculum learning}}.{\BBCQ}
\newblock
\BIn{} A\BPBI P.~Danyluk, L.~Bottou\BCBL {}\ \BBA {} M\BPBI L.~Littman\ (\BEDS), \APACrefbtitle {Proceedings of the 26th Annual International Conference on Machine Learning, {ICML} 2009, Montreal, Quebec, Canada, June 14-18, 2009} {Proceedings of the 26th annual international conference on machine learning, {ICML} 2009, montreal, quebec, canada, june 14-18, 2009}\ (\BVOL~382, \BPGS\ 41--48).
\newblock
\APACaddressPublisher{}{{ACM}}.
\newblock
\begin{APACrefURL} \url{https://doi.org/10.1145/1553374.1553380} \end{APACrefURL}
\newblock
\begin{APACrefDOI} \doi{10.1145/1553374.1553380} \end{APACrefDOI}
\PrintBackRefs{\CurrentBib}

\bibitem [\protect \citeauthoryear {%
Berthelot%
, Carlini%
, Cubuk%
\BCBL {}\ \protect \BOthers {.}}{%
Berthelot%
, Carlini%
, Cubuk%
\BCBL {}\ \protect \BOthers {.}}{%
{\protect \APACyear {2019}}%
}]{%
ReMixMatch}
\APACinsertmetastar {%
ReMixMatch}%
\begin{APACrefauthors}%
Berthelot, D.%
, Carlini, N.%
, Cubuk, E\BPBI D.%
, Kurakin, A.%
, Sohn, K.%
, Zhang, H.%
\BCBL {}\ \BBA {} Raffel, C.%
\end{APACrefauthors}%
\unskip\
\newblock
\APACrefYearMonthDay{2019}{}{}.
\newblock
{\BBOQ}\APACrefatitle {{ReMixMatch: Semi-Supervised Learning with Distribution Alignment and Augmentation Anchoring}} {{ReMixMatch: Semi-Supervised Learning with Distribution Alignment and Augmentation Anchoring}}.{\BBCQ}
\newblock
\APACjournalVolNumPages{CoRR}{abs/1911.09785}{}{}.
\newblock
\begin{APACrefURL} \url{http://arxiv.org/abs/1911.09785} \end{APACrefURL}
\PrintBackRefs{\CurrentBib}

\bibitem [\protect \citeauthoryear {%
Berthelot%
, Carlini%
, Goodfellow%
\BCBL {}\ \protect \BOthers {.}}{%
Berthelot%
, Carlini%
, Goodfellow%
\BCBL {}\ \protect \BOthers {.}}{%
{\protect \APACyear {2019}}%
}]{%
MixMatch}
\APACinsertmetastar {%
MixMatch}%
\begin{APACrefauthors}%
Berthelot, D.%
, Carlini, N.%
, Goodfellow, I\BPBI J.%
, Papernot, N.%
, Oliver, A.%
\BCBL {}\ \BBA {} Raffel, C.%
\end{APACrefauthors}%
\unskip\
\newblock
\APACrefYearMonthDay{2019}{}{}.
\newblock
{\BBOQ}\APACrefatitle {{MixMatch: {A} Holistic Approach to Semi-Supervised Learning}} {{MixMatch: {A} Holistic Approach to Semi-Supervised Learning}}.{\BBCQ}
\newblock
\BIn{} H\BPBI M.~Wallach, H.~Larochelle, A.~Beygelzimer, F.~d'Alch{\'{e}}{-}Buc, E\BPBI B.~Fox\BCBL {}\ \BBA {} R.~Garnett\ (\BEDS), \APACrefbtitle {Advances in Neural Information Processing Systems 32: Annual Conference on Neural Information Processing Systems 2019, NeurIPS 2019, December 8-14, 2019, Vancouver, BC, Canada} {Advances in neural information processing systems 32: Annual conference on neural information processing systems 2019, neurips 2019, december 8-14, 2019, vancouver, bc, canada}\ (\BPGS\ 5050--5060).
\newblock
\begin{APACrefURL} \url{https://proceedings.neurips.cc/paper/2019/hash/1cd138d0499a68f4bb72bee04bbec2d7-Abstract.html} \end{APACrefURL}
\PrintBackRefs{\CurrentBib}

\bibitem [\protect \citeauthoryear {%
Blum%
\ \BBA {} Mitchell%
}{%
Blum%
\ \BBA {} Mitchell%
}{%
{\protect \APACyear {1998}}%
}]{%
cotraining}
\APACinsertmetastar {%
cotraining}%
\begin{APACrefauthors}%
Blum, A.%
\BCBT {}\ \BBA {} Mitchell, T\BPBI M.%
\end{APACrefauthors}%
\unskip\
\newblock
\APACrefYearMonthDay{1998}{}{}.
\newblock
{\BBOQ}\APACrefatitle {{Combining Labeled and Unlabeled Data with Co-Training}} {{Combining Labeled and Unlabeled Data with Co-Training}}.{\BBCQ}
\newblock
\BIn{} P\BPBI L.~Bartlett\ \BBA {} Y.~Mansour\ (\BEDS), \APACrefbtitle {Proceedings of the Eleventh Annual Conference on Computational Learning Theory, {COLT} 1998, Madison, Wisconsin, USA, July 24-26, 1998} {Proceedings of the eleventh annual conference on computational learning theory, {COLT} 1998, madison, wisconsin, usa, july 24-26, 1998}\ (\BPGS\ 92--100).
\newblock
\APACaddressPublisher{}{{ACM}}.
\newblock
\begin{APACrefURL} \url{https://doi.org/10.1145/279943.279962} \end{APACrefURL}
\newblock
\begin{APACrefDOI} \doi{10.1145/279943.279962} \end{APACrefDOI}
\PrintBackRefs{\CurrentBib}

\bibitem [\protect \citeauthoryear {%
Blundell%
, Cornebise%
, Kavukcuoglu%
\BCBL {}\ \BBA {} Wierstra%
}{%
Blundell%
\ \protect \BOthers {.}}{%
{\protect \APACyear {2015}}%
}]{%
WeightUncertainty}
\APACinsertmetastar {%
WeightUncertainty}%
\begin{APACrefauthors}%
Blundell, C.%
, Cornebise, J.%
, Kavukcuoglu, K.%
\BCBL {}\ \BBA {} Wierstra, D.%
\end{APACrefauthors}%
\unskip\
\newblock
\APACrefYearMonthDay{2015}{}{}.
\newblock
{\BBOQ}\APACrefatitle {{Weight Uncertainty in Neural Network}} {{Weight Uncertainty in Neural Network}}.{\BBCQ}
\newblock
\BIn{} F\BPBI R.~Bach\ \BBA {} D\BPBI M.~Blei\ (\BEDS), \APACrefbtitle {Proceedings of the 32nd International Conference on Machine Learning, {ICML} 2015, Lille, France, 6-11 July 2015} {Proceedings of the 32nd international conference on machine learning, {ICML} 2015, lille, france, 6-11 july 2015}\ (\BVOL~37, \BPGS\ 1613--1622).
\newblock
\APACaddressPublisher{}{JMLR.org}.
\newblock
\begin{APACrefURL} \url{http://proceedings.mlr.press/v37/blundell15.html} \end{APACrefURL}
\PrintBackRefs{\CurrentBib}

\bibitem [\protect \citeauthoryear {%
Caron%
, Bojanowski%
, Joulin%
\BCBL {}\ \BBA {} Douze%
}{%
Caron%
\ \protect \BOthers {.}}{%
{\protect \APACyear {2018}}%
}]{%
DeepCluster}
\APACinsertmetastar {%
DeepCluster}%
\begin{APACrefauthors}%
Caron, M.%
, Bojanowski, P.%
, Joulin, A.%
\BCBL {}\ \BBA {} Douze, M.%
\end{APACrefauthors}%
\unskip\
\newblock
\APACrefYearMonthDay{2018}{}{}.
\newblock
{\BBOQ}\APACrefatitle {{Deep Clustering for Unsupervised Learning of Visual Features}} {{Deep Clustering for Unsupervised Learning of Visual Features}}.{\BBCQ}
\newblock
\BIn{} V.~Ferrari, M.~Hebert, C.~Sminchisescu\BCBL {}\ \BBA {} Y.~Weiss\ (\BEDS), \APACrefbtitle {Computer Vision - {ECCV} 2018 - 15th European Conference, Munich, Germany, September 8-14, 2018, Proceedings, Part {XIV}} {Computer vision - {ECCV} 2018 - 15th european conference, munich, germany, september 8-14, 2018, proceedings, part {XIV}}\ (\BVOL\ 11218, \BPGS\ 139--156).
\newblock
\APACaddressPublisher{}{Springer}.
\newblock
\begin{APACrefURL} \url{https://doi.org/10.1007/978-3-030-01264-9\_9} \end{APACrefURL}
\newblock
\begin{APACrefDOI} \doi{10.1007/978-3-030-01264-9\_9} \end{APACrefDOI}
\PrintBackRefs{\CurrentBib}

\bibitem [\protect \citeauthoryear {%
Caron%
\ \protect \BOthers {.}}{%
Caron%
\ \protect \BOthers {.}}{%
{\protect \APACyear {2020}}%
}]{%
SwAV}
\APACinsertmetastar {%
SwAV}%
\begin{APACrefauthors}%
Caron, M.%
, Misra, I.%
, Mairal, J.%
, Goyal, P.%
, Bojanowski, P.%
\BCBL {}\ \BBA {} Joulin, A.%
\end{APACrefauthors}%
\unskip\
\newblock
\APACrefYearMonthDay{2020}{}{}.
\newblock
{\BBOQ}\APACrefatitle {{Unsupervised Learning of Visual Features by Contrasting Cluster Assignments}} {{Unsupervised Learning of Visual Features by Contrasting Cluster Assignments}}.{\BBCQ}
\newblock
\BIn{} H.~Larochelle, M.~Ranzato, R.~Hadsell, M.~Balcan\BCBL {}\ \BBA {} H.~Lin\ (\BEDS), \APACrefbtitle {Advances in Neural Information Processing Systems 33: Annual Conference on Neural Information Processing Systems 2020, NeurIPS 2020, December 6-12, 2020, virtual.} {Advances in neural information processing systems 33: Annual conference on neural information processing systems 2020, neurips 2020, december 6-12, 2020, virtual.}
\newblock
\begin{APACrefURL} \url{https://proceedings.neurips.cc/paper/2020/hash/70feb62b69f16e0238f741fab228fec2-Abstract.html} \end{APACrefURL}
\PrintBackRefs{\CurrentBib}

\bibitem [\protect \citeauthoryear {%
Caron%
\ \protect \BOthers {.}}{%
Caron%
\ \protect \BOthers {.}}{%
{\protect \APACyear {2021}}%
}]{%
DINO}
\APACinsertmetastar {%
DINO}%
\begin{APACrefauthors}%
Caron, M.%
, Touvron, H.%
, Misra, I.%
, J{\'{e}}gou, H.%
, Mairal, J.%
, Bojanowski, P.%
\BCBL {}\ \BBA {} Joulin, A.%
\end{APACrefauthors}%
\unskip\
\newblock
\APACrefYearMonthDay{2021}{}{}.
\newblock
{\BBOQ}\APACrefatitle {{Emerging Properties in Self-Supervised Vision Transformers}} {{Emerging Properties in Self-Supervised Vision Transformers}}.{\BBCQ}
\newblock
\BIn{} \APACrefbtitle {2021 {IEEE/CVF} International Conference on Computer Vision, {ICCV} 2021, Montreal, QC, Canada, October 10-17, 2021} {2021 {IEEE/CVF} international conference on computer vision, {ICCV} 2021, montreal, qc, canada, october 10-17, 2021}\ (\BPGS\ 9630--9640).
\newblock
\APACaddressPublisher{}{{IEEE}}.
\newblock
\begin{APACrefURL} \url{https://doi.org/10.1109/ICCV48922.2021.00951} \end{APACrefURL}
\newblock
\begin{APACrefDOI} \doi{10.1109/ICCV48922.2021.00951} \end{APACrefDOI}
\PrintBackRefs{\CurrentBib}

\bibitem [\protect \citeauthoryear {%
Cascante{-}Bonilla%
, Tan%
, Qi%
\BCBL {}\ \BBA {} Ordonez%
}{%
Cascante{-}Bonilla%
\ \protect \BOthers {.}}{%
{\protect \APACyear {2020}}%
}]{%
CL4PL}
\APACinsertmetastar {%
CL4PL}%
\begin{APACrefauthors}%
Cascante{-}Bonilla, P.%
, Tan, F.%
, Qi, Y.%
\BCBL {}\ \BBA {} Ordonez, V.%
\end{APACrefauthors}%
\unskip\
\newblock
\APACrefYearMonthDay{2020}{}{}.
\newblock
{\BBOQ}\APACrefatitle {{Curriculum Labeling: Self-paced Pseudo-Labeling for Semi-Supervised Learning}} {{Curriculum Labeling: Self-paced Pseudo-Labeling for Semi-Supervised Learning}}.{\BBCQ}
\newblock
\APACjournalVolNumPages{CoRR}{abs/2001.06001}{}{}.
\newblock
\begin{APACrefURL} \url{https://arxiv.org/abs/2001.06001} \end{APACrefURL}
\PrintBackRefs{\CurrentBib}

\bibitem [\protect \citeauthoryear {%
Chapelle%
, Sch{\"{o}}lkopf%
\BCBL {}\ \BBA {} Zien%
}{%
Chapelle%
\ \protect \BOthers {.}}{%
{\protect \APACyear {2006}}%
}]{%
chapelle_semi-supervised_2006}
\APACinsertmetastar {%
chapelle_semi-supervised_2006}%
\begin{APACrefauthors}%
Chapelle, O.%
, Sch{\"{o}}lkopf, B.%
\BCBL {}\ \BBA {} Zien, A.%
\end{APACrefauthors}%
\ (\BEDS).
\unskip\
\newblock
\APACrefYear{2006}.
\newblock
\APACrefbtitle {{Semi-Supervised Learning}} {{Semi-Supervised Learning}}.
\newblock
\APACaddressPublisher{}{The {MIT} Press}.
\newblock
\begin{APACrefURL} \url{https://doi.org/10.7551/mitpress/9780262033589.001.0001} \end{APACrefURL}
\newblock
\begin{APACrefDOI} \doi{10.7551/MITPRESS/9780262033589.001.0001} \end{APACrefDOI}
\PrintBackRefs{\CurrentBib}

\bibitem [\protect \citeauthoryear {%
M.~Chen%
, Du%
, Zhang%
, Qian%
\BCBL {}\ \BBA {} Wang%
}{%
M.~Chen%
\ \protect \BOthers {.}}{%
{\protect \APACyear {2022}}%
}]{%
MHCotraining}
\APACinsertmetastar {%
MHCotraining}%
\begin{APACrefauthors}%
Chen, M.%
, Du, Y.%
, Zhang, Y.%
, Qian, S.%
\BCBL {}\ \BBA {} Wang, C.%
\end{APACrefauthors}%
\unskip\
\newblock
\APACrefYearMonthDay{2022}{}{}.
\newblock
{\BBOQ}\APACrefatitle {{Semi-supervised Learning with Multi-Head Co-Training}} {{Semi-supervised Learning with Multi-Head Co-Training}}.{\BBCQ}
\newblock
\BIn{} \APACrefbtitle {Thirty-Sixth {AAAI} Conference on Artificial Intelligence, {AAAI} 2022, Thirty-Fourth Conference on Innovative Applications of Artificial Intelligence, {IAAI} 2022, The Twelveth Symposium on Educational Advances in Artificial Intelligence, {EAAI} 2022 Virtual Event, February 22 - March 1, 2022} {Thirty-sixth {AAAI} conference on artificial intelligence, {AAAI} 2022, thirty-fourth conference on innovative applications of artificial intelligence, {IAAI} 2022, the twelveth symposium on educational advances in artificial intelligence, {EAAI} 2022 virtual event, february 22 - march 1, 2022}\ (\BPGS\ 6278--6286).
\newblock
\APACaddressPublisher{}{{AAAI} Press}.
\newblock
\begin{APACrefURL} \url{https://doi.org/10.1609/aaai.v36i6.20577} \end{APACrefURL}
\newblock
\begin{APACrefDOI} \doi{10.1609/AAAI.V36I6.20577} \end{APACrefDOI}
\PrintBackRefs{\CurrentBib}

\bibitem [\protect \citeauthoryear {%
T.~Chen%
, Kornblith%
, Norouzi%
\BCBL {}\ \BBA {} Hinton%
}{%
T.~Chen%
\ \protect \BOthers {.}}{%
{\protect \APACyear {2020}}%
}]{%
SimCLR}
\APACinsertmetastar {%
SimCLR}%
\begin{APACrefauthors}%
Chen, T.%
, Kornblith, S.%
, Norouzi, M.%
\BCBL {}\ \BBA {} Hinton, G\BPBI E.%
\end{APACrefauthors}%
\unskip\
\newblock
\APACrefYearMonthDay{2020}{}{}.
\newblock
{\BBOQ}\APACrefatitle {{A Simple Framework for Contrastive Learning of Visual Representations}} {{A Simple Framework for Contrastive Learning of Visual Representations}}.{\BBCQ}
\newblock
\BIn{} \APACrefbtitle {Proceedings of the 37th International Conference on Machine Learning, {ICML} 2020, 13-18 July 2020, Virtual Event} {Proceedings of the 37th international conference on machine learning, {ICML} 2020, 13-18 july 2020, virtual event}\ (\BVOL~119, \BPGS\ 1597--1607).
\newblock
\APACaddressPublisher{}{{PMLR}}.
\newblock
\begin{APACrefURL} \url{http://proceedings.mlr.press/v119/chen20j.html} \end{APACrefURL}
\PrintBackRefs{\CurrentBib}

\bibitem [\protect \citeauthoryear {%
Cho%
, Alizadeh{-}Vahid%
, Adya%
\BCBL {}\ \BBA {} Rastegari%
}{%
Cho%
\ \protect \BOthers {.}}{%
{\protect \APACyear {2022}}%
}]{%
k-meansNeuronClustering}
\APACinsertmetastar {%
k-meansNeuronClustering}%
\begin{APACrefauthors}%
Cho, M.%
, Alizadeh{-}Vahid, K.%
, Adya, S.%
\BCBL {}\ \BBA {} Rastegari, M.%
\end{APACrefauthors}%
\unskip\
\newblock
\APACrefYearMonthDay{2022}{}{}.
\newblock
{\BBOQ}\APACrefatitle {{{DKM:} Differentiable k-Means Clustering Layer for Neural Network Compression}} {{{DKM:} Differentiable k-Means Clustering Layer for Neural Network Compression}}.{\BBCQ}
\newblock
\BIn{} \APACrefbtitle {The Tenth International Conference on Learning Representations, {ICLR} 2022, Virtual Event, April 25-29, 2022.} {The tenth international conference on learning representations, {ICLR} 2022, virtual event, april 25-29, 2022.}
\newblock
\APACaddressPublisher{}{OpenReview.net}.
\newblock
\begin{APACrefURL} \url{https://openreview.net/forum?id=J\_F\_qqCE3Z5} \end{APACrefURL}
\PrintBackRefs{\CurrentBib}

\bibitem [\protect \citeauthoryear {%
Computer%
}{%
Computer%
}{%
{\protect \APACyear {2023}}%
}]{%
redpajama}
\APACinsertmetastar {%
redpajama}%
\begin{APACrefauthors}%
Computer, T.%
\end{APACrefauthors}%
\unskip\
\newblock
\APACrefYearMonthDay{2023}{October}{}.
\newblock
\APACrefbtitle {{RedPajama: an Open Dataset for Training Large Language Models}.} {{RedPajama: an Open Dataset for Training Large Language Models}.}
\newblock
\APAChowpublished {\url{https://github.com/togethercomputer/RedPajama-Data}}.
\PrintBackRefs{\CurrentBib}

\bibitem [\protect \citeauthoryear {%
Cubuk%
, Zoph%
, Man{\'{e}}%
, Vasudevan%
\BCBL {}\ \BBA {} Le%
}{%
Cubuk%
\ \protect \BOthers {.}}{%
{\protect \APACyear {2018}}%
}]{%
AutoAug}
\APACinsertmetastar {%
AutoAug}%
\begin{APACrefauthors}%
Cubuk, E\BPBI D.%
, Zoph, B.%
, Man{\'{e}}, D.%
, Vasudevan, V.%
\BCBL {}\ \BBA {} Le, Q\BPBI V.%
\end{APACrefauthors}%
\unskip\
\newblock
\APACrefYearMonthDay{2018}{}{}.
\newblock
{\BBOQ}\APACrefatitle {{AutoAugment: Learning Augmentation Policies from Data}} {{AutoAugment: Learning Augmentation Policies from Data}}.{\BBCQ}
\newblock
\APACjournalVolNumPages{CoRR}{abs/1805.09501}{}{}.
\newblock
\begin{APACrefURL} \url{http://arxiv.org/abs/1805.09501} \end{APACrefURL}
\PrintBackRefs{\CurrentBib}

\bibitem [\protect \citeauthoryear {%
Cubuk%
, Zoph%
, Shlens%
\BCBL {}\ \BBA {} Le%
}{%
Cubuk%
\ \protect \BOthers {.}}{%
{\protect \APACyear {2019}}%
}]{%
RandAug}
\APACinsertmetastar {%
RandAug}%
\begin{APACrefauthors}%
Cubuk, E\BPBI D.%
, Zoph, B.%
, Shlens, J.%
\BCBL {}\ \BBA {} Le, Q\BPBI V.%
\end{APACrefauthors}%
\unskip\
\newblock
\APACrefYearMonthDay{2019}{}{}.
\newblock
{\BBOQ}\APACrefatitle {{RandAugment: Practical data augmentation with no separate search}} {{RandAugment: Practical data augmentation with no separate search}}.{\BBCQ}
\newblock
\APACjournalVolNumPages{CoRR}{abs/1909.13719}{}{}.
\newblock
\begin{APACrefURL} \url{http://arxiv.org/abs/1909.13719} \end{APACrefURL}
\PrintBackRefs{\CurrentBib}

\bibitem [\protect \citeauthoryear {%
Duval%
, Misra%
\BCBL {}\ \BBA {} Ballas%
}{%
Duval%
\ \protect \BOthers {.}}{%
{\protect \APACyear {2023}}%
}]{%
KD-RoB}
\APACinsertmetastar {%
KD-RoB}%
\begin{APACrefauthors}%
Duval, Q.%
, Misra, I.%
\BCBL {}\ \BBA {} Ballas, N.%
\end{APACrefauthors}%
\unskip\
\newblock
\APACrefYearMonthDay{2023}{}{}.
\newblock
{\BBOQ}\APACrefatitle {{A Simple Recipe for Competitive Low-compute Self supervised Vision Models}} {{A Simple Recipe for Competitive Low-compute Self supervised Vision Models}}.{\BBCQ}
\newblock
\APACjournalVolNumPages{CoRR}{abs/2301.09451}{}{}.
\newblock
\begin{APACrefURL} \url{https://doi.org/10.48550/arXiv.2301.09451} \end{APACrefURL}
\newblock
\begin{APACrefDOI} \doi{10.48550/ARXIV.2301.09451} \end{APACrefDOI}
\PrintBackRefs{\CurrentBib}

\bibitem [\protect \citeauthoryear {%
Fr{\'{e}}nay%
\ \BBA {} Verleysen%
}{%
Fr{\'{e}}nay%
\ \BBA {} Verleysen%
}{%
{\protect \APACyear {2014}}%
}]{%
Noise:Label:Causality}
\APACinsertmetastar {%
Noise:Label:Causality}%
\begin{APACrefauthors}%
Fr{\'{e}}nay, B.%
\BCBT {}\ \BBA {} Verleysen, M.%
\end{APACrefauthors}%
\unskip\
\newblock
\APACrefYearMonthDay{2014}{}{}.
\newblock
{\BBOQ}\APACrefatitle {{Classification in the Presence of Label Noise: {A} Survey}} {{Classification in the Presence of Label Noise: {A} Survey}}.{\BBCQ}
\newblock
\APACjournalVolNumPages{{IEEE} Trans. Neural Networks Learn. Syst.}{25}{5}{845--869}.
\newblock
\begin{APACrefURL} \url{https://doi.org/10.1109/TNNLS.2013.2292894} \end{APACrefURL}
\newblock
\begin{APACrefDOI} \doi{10.1109/TNNLS.2013.2292894} \end{APACrefDOI}
\PrintBackRefs{\CurrentBib}

\bibitem [\protect \citeauthoryear {%
Gal%
\ \BBA {} Ghahramani%
}{%
Gal%
\ \BBA {} Ghahramani%
}{%
{\protect \APACyear {2016}}%
}]{%
MCDropout}
\APACinsertmetastar {%
MCDropout}%
\begin{APACrefauthors}%
Gal, Y.%
\BCBT {}\ \BBA {} Ghahramani, Z.%
\end{APACrefauthors}%
\unskip\
\newblock
\APACrefYearMonthDay{2016}{}{}.
\newblock
{\BBOQ}\APACrefatitle {{Dropout as a Bayesian Approximation: Representing Model Uncertainty in Deep Learning}} {{Dropout as a Bayesian Approximation: Representing Model Uncertainty in Deep Learning}}.{\BBCQ}
\newblock
\BIn{} M.~Balcan\ \BBA {} K\BPBI Q.~Weinberger\ (\BEDS), \APACrefbtitle {Proceedings of the 33nd International Conference on Machine Learning, {ICML} 2016, New York City, NY, USA, June 19-24, 2016} {Proceedings of the 33nd international conference on machine learning, {ICML} 2016, new york city, ny, usa, june 19-24, 2016}\ (\BVOL~48, \BPGS\ 1050--1059).
\newblock
\APACaddressPublisher{}{JMLR.org}.
\newblock
\begin{APACrefURL} \url{http://proceedings.mlr.press/v48/gal16.html} \end{APACrefURL}
\PrintBackRefs{\CurrentBib}

\bibitem [\protect \citeauthoryear {%
Gao%
, Xing%
, Xie%
, Wu%
\BCBL {}\ \BBA {} Geng%
}{%
Gao%
\ \protect \BOthers {.}}{%
{\protect \APACyear {2017}}%
}]{%
DLDL}
\APACinsertmetastar {%
DLDL}%
\begin{APACrefauthors}%
Gao, B\BHBI B.%
, Xing, C.%
, Xie, C\BHBI W.%
, Wu, J.%
\BCBL {}\ \BBA {} Geng, X.%
\end{APACrefauthors}%
\unskip\
\newblock
\APACrefYearMonthDay{2017}{{\APACmonth{06}}}{}.
\newblock
{\BBOQ}\APACrefatitle {Deep {{Label Distribution Learning}} with {{Label Ambiguity}}} {Deep {{Label Distribution Learning}} with {{Label Ambiguity}}}.{\BBCQ}
\newblock
\APACjournalVolNumPages{IEEE Transactions on Image Processing}{26}{6}{2825--2838}.
\newblock
\begin{APACrefDOI} \doi{10.1109/TIP.2017.2689998} \end{APACrefDOI}
\PrintBackRefs{\CurrentBib}

\bibitem [\protect \citeauthoryear {%
G{\'e}ron%
}{%
G{\'e}ron%
}{%
{\protect \APACyear {2022}}%
}]{%
Book:HandsOn}
\APACinsertmetastar {%
Book:HandsOn}%
\begin{APACrefauthors}%
G{\'e}ron, A.%
\end{APACrefauthors}%
\unskip\
\newblock
\APACrefYear{2022}.
\newblock
\APACrefbtitle {{Hands-on machine learning with Scikit-Learn, Keras, and TensorFlow}} {{Hands-on machine learning with Scikit-Learn, Keras, and TensorFlow}}.
\newblock
\APACaddressPublisher{}{" O'Reilly Media, Inc."}.
\PrintBackRefs{\CurrentBib}

\bibitem [\protect \citeauthoryear {%
Goldberger%
\ \BBA {} Ben{-}Reuven%
}{%
Goldberger%
\ \BBA {} Ben{-}Reuven%
}{%
{\protect \APACyear {2017}}%
}]{%
Noise:Label:AdaptationLayer}
\APACinsertmetastar {%
Noise:Label:AdaptationLayer}%
\begin{APACrefauthors}%
Goldberger, J.%
\BCBT {}\ \BBA {} Ben{-}Reuven, E.%
\end{APACrefauthors}%
\unskip\
\newblock
\APACrefYearMonthDay{2017}{}{}.
\newblock
{\BBOQ}\APACrefatitle {{Training deep neural-networks using a noise adaptation layer}} {{Training deep neural-networks using a noise adaptation layer}}.{\BBCQ}
\newblock
\BIn{} \APACrefbtitle {5th International Conference on Learning Representations, {ICLR} 2017, Toulon, France, April 24-26, 2017, Conference Track Proceedings.} {5th international conference on learning representations, {ICLR} 2017, toulon, france, april 24-26, 2017, conference track proceedings.}
\newblock
\APACaddressPublisher{}{OpenReview.net}.
\newblock
\begin{APACrefURL} \url{https://openreview.net/forum?id=H12GRgcxg} \end{APACrefURL}
\PrintBackRefs{\CurrentBib}

\bibitem [\protect \citeauthoryear {%
Goldman%
\ \BBA {} Zhou%
}{%
Goldman%
\ \BBA {} Zhou%
}{%
{\protect \APACyear {2000}}%
}]{%
Cotraining:GoldmanZhu}
\APACinsertmetastar {%
Cotraining:GoldmanZhu}%
\begin{APACrefauthors}%
Goldman, S\BPBI A.%
\BCBT {}\ \BBA {} Zhou, Y.%
\end{APACrefauthors}%
\unskip\
\newblock
\APACrefYearMonthDay{2000}{}{}.
\newblock
{\BBOQ}\APACrefatitle {{Enhancing Supervised Learning with Unlabeled Data}} {{Enhancing Supervised Learning with Unlabeled Data}}.{\BBCQ}
\newblock
\BIn{} P.~Langley\ (\BED), \APACrefbtitle {Proceedings of the Seventeenth International Conference on Machine Learning {(ICML} 2000), Stanford University, Stanford, CA, USA, June 29 - July 2, 2000} {Proceedings of the seventeenth international conference on machine learning {(ICML} 2000), stanford university, stanford, ca, usa, june 29 - july 2, 2000}\ (\BPGS\ 327--334).
\newblock
\APACaddressPublisher{}{Morgan Kaufmann}.
\PrintBackRefs{\CurrentBib}

\bibitem [\protect \citeauthoryear {%
Gou%
, Yu%
, Maybank%
\BCBL {}\ \BBA {} Tao%
}{%
Gou%
\ \protect \BOthers {.}}{%
{\protect \APACyear {2021}}%
}]{%
KD-Survey}
\APACinsertmetastar {%
KD-Survey}%
\begin{APACrefauthors}%
Gou, J.%
, Yu, B.%
, Maybank, S\BPBI J.%
\BCBL {}\ \BBA {} Tao, D.%
\end{APACrefauthors}%
\unskip\
\newblock
\APACrefYearMonthDay{2021}{}{}.
\newblock
{\BBOQ}\APACrefatitle {{Knowledge Distillation: {A} Survey}} {{Knowledge Distillation: {A} Survey}}.{\BBCQ}
\newblock
\APACjournalVolNumPages{Int. J. Comput. Vis.}{129}{6}{1789--1819}.
\newblock
\begin{APACrefURL} \url{https://doi.org/10.1007/s11263-021-01453-z} \end{APACrefURL}
\newblock
\begin{APACrefDOI} \doi{10.1007/S11263-021-01453-Z} \end{APACrefDOI}
\PrintBackRefs{\CurrentBib}

\bibitem [\protect \citeauthoryear {%
Graves%
}{%
Graves%
}{%
{\protect \APACyear {2011}}%
}]{%
PracticalVariationalInference}
\APACinsertmetastar {%
PracticalVariationalInference}%
\begin{APACrefauthors}%
Graves, A.%
\end{APACrefauthors}%
\unskip\
\newblock
\APACrefYearMonthDay{2011}{}{}.
\newblock
{\BBOQ}\APACrefatitle {{Practical Variational Inference for Neural Networks}} {{Practical Variational Inference for Neural Networks}}.{\BBCQ}
\newblock
\BIn{} J.~Shawe{-}Taylor, R\BPBI S.~Zemel, P\BPBI L.~Bartlett, F\BPBI C\BPBI N.~Pereira\BCBL {}\ \BBA {} K\BPBI Q.~Weinberger\ (\BEDS), \APACrefbtitle {Advances in Neural Information Processing Systems 24: 25th Annual Conference on Neural Information Processing Systems 2011. Proceedings of a meeting held 12-14 December 2011, Granada, Spain} {Advances in neural information processing systems 24: 25th annual conference on neural information processing systems 2011. proceedings of a meeting held 12-14 december 2011, granada, spain}\ (\BPGS\ 2348--2356).
\newblock
\begin{APACrefURL} \url{https://proceedings.neurips.cc/paper/2011/hash/7eb3c8be3d411e8ebfab08eba5f49632-Abstract.html} \end{APACrefURL}
\PrintBackRefs{\CurrentBib}

\bibitem [\protect \citeauthoryear {%
Grill%
\ \protect \BOthers {.}}{%
Grill%
\ \protect \BOthers {.}}{%
{\protect \APACyear {2020}}%
}]{%
BYOL}
\APACinsertmetastar {%
BYOL}%
\begin{APACrefauthors}%
Grill, J.%
, Strub, F.%
, Altch{\'{e}}, F.%
, Tallec, C.%
, Richemond, P\BPBI H.%
, Buchatskaya, E.%
\BDBL {}Valko, M.%
\end{APACrefauthors}%
\unskip\
\newblock
\APACrefYearMonthDay{2020}{}{}.
\newblock
{\BBOQ}\APACrefatitle {{Bootstrap Your Own Latent - {A} New Approach to Self-Supervised Learning}} {{Bootstrap Your Own Latent - {A} New Approach to Self-Supervised Learning}}.{\BBCQ}
\newblock
\BIn{} H.~Larochelle, M.~Ranzato, R.~Hadsell, M.~Balcan\BCBL {}\ \BBA {} H.~Lin\ (\BEDS), \APACrefbtitle {Advances in Neural Information Processing Systems 33: Annual Conference on Neural Information Processing Systems 2020, NeurIPS 2020, December 6-12, 2020, virtual.} {Advances in neural information processing systems 33: Annual conference on neural information processing systems 2020, neurips 2020, december 6-12, 2020, virtual.}
\newblock
\begin{APACrefURL} \url{https://proceedings.neurips.cc/paper/2020/hash/f3ada80d5c4ee70142b17b8192b2958e-Abstract.html} \end{APACrefURL}
\PrintBackRefs{\CurrentBib}

\bibitem [\protect \citeauthoryear {%
C.~Guo%
, Pleiss%
, Sun%
\BCBL {}\ \BBA {} Weinberger%
}{%
C.~Guo%
\ \protect \BOthers {.}}{%
{\protect \APACyear {2017}}%
}]{%
Calibration}
\APACinsertmetastar {%
Calibration}%
\begin{APACrefauthors}%
Guo, C.%
, Pleiss, G.%
, Sun, Y.%
\BCBL {}\ \BBA {} Weinberger, K\BPBI Q.%
\end{APACrefauthors}%
\unskip\
\newblock
\APACrefYearMonthDay{2017}{}{}.
\newblock
{\BBOQ}\APACrefatitle {{On Calibration of Modern Neural Networks}} {{On Calibration of Modern Neural Networks}}.{\BBCQ}
\newblock
\BIn{} D.~Precup\ \BBA {} Y\BPBI W.~Teh\ (\BEDS), \APACrefbtitle {{Proceedings of the 34th International Conference on Machine Learning, {ICML} 2017, Sydney, NSW, Australia, 6-11 August 2017}} {{Proceedings of the 34th International Conference on Machine Learning, {ICML} 2017, Sydney, NSW, Australia, 6-11 August 2017}}\ (\BVOL~70, \BPGS\ 1321--1330).
\newblock
\APACaddressPublisher{}{{PMLR}}.
\newblock
\begin{APACrefURL} \url{http://proceedings.mlr.press/v70/guo17a.html} \end{APACrefURL}
\PrintBackRefs{\CurrentBib}

\bibitem [\protect \citeauthoryear {%
S.~Guo%
\ \protect \BOthers {.}}{%
S.~Guo%
\ \protect \BOthers {.}}{%
{\protect \APACyear {2018}}%
}]{%
Noise:Label:CurriculumNet}
\APACinsertmetastar {%
Noise:Label:CurriculumNet}%
\begin{APACrefauthors}%
Guo, S.%
, Huang, W.%
, Zhang, H.%
, Zhuang, C.%
, Dong, D.%
, Scott, M\BPBI R.%
\BCBL {}\ \BBA {} Huang, D.%
\end{APACrefauthors}%
\unskip\
\newblock
\APACrefYearMonthDay{2018}{}{}.
\newblock
{\BBOQ}\APACrefatitle {{CurriculumNet: Weakly Supervised Learning from Large-Scale Web Images}} {{CurriculumNet: Weakly Supervised Learning from Large-Scale Web Images}}.{\BBCQ}
\newblock
\BIn{} V.~Ferrari, M.~Hebert, C.~Sminchisescu\BCBL {}\ \BBA {} Y.~Weiss\ (\BEDS), \APACrefbtitle {{Computer Vision - {ECCV} 2018 - 15th European Conference, Munich, Germany, September 8-14, 2018, Proceedings, Part {X}}} {{Computer Vision - {ECCV} 2018 - 15th European Conference, Munich, Germany, September 8-14, 2018, Proceedings, Part {X}}}\ (\BVOL\ 11214, \BPGS\ 139--154).
\newblock
\APACaddressPublisher{}{Springer}.
\newblock
\begin{APACrefURL} \url{https://doi.org/10.1007/978-3-030-01249-6\_9} \end{APACrefURL}
\newblock
\begin{APACrefDOI} \doi{10.1007/978-3-030-01249-6\_9} \end{APACrefDOI}
\PrintBackRefs{\CurrentBib}

\bibitem [\protect \citeauthoryear {%
Haase{-}Sch{\"{u}}tz%
, Stal%
, Hertlein%
\BCBL {}\ \BBA {} Sick%
}{%
Haase{-}Sch{\"{u}}tz%
\ \protect \BOthers {.}}{%
{\protect \APACyear {2020}}%
}]{%
ILI}
\APACinsertmetastar {%
ILI}%
\begin{APACrefauthors}%
Haase{-}Sch{\"{u}}tz, C.%
, Stal, R.%
, Hertlein, H.%
\BCBL {}\ \BBA {} Sick, B.%
\end{APACrefauthors}%
\unskip\
\newblock
\APACrefYearMonthDay{2020}{}{}.
\newblock
{\BBOQ}\APACrefatitle {{Iterative Label Improvement: Robust Training by Confidence Based Filtering and Dataset Partitioning}} {{Iterative Label Improvement: Robust Training by Confidence Based Filtering and Dataset Partitioning}}.{\BBCQ}
\newblock
\BIn{} \APACrefbtitle {25th International Conference on Pattern Recognition, {ICPR} 2020, Virtual Event / Milan, Italy, January 10-15, 2021} {25th international conference on pattern recognition, {ICPR} 2020, virtual event / milan, italy, january 10-15, 2021}\ (\BPGS\ 9483--9490).
\newblock
\APACaddressPublisher{}{{IEEE}}.
\newblock
\begin{APACrefURL} \url{https://doi.org/10.1109/ICPR48806.2021.9411918} \end{APACrefURL}
\newblock
\begin{APACrefDOI} \doi{10.1109/ICPR48806.2021.9411918} \end{APACrefDOI}
\PrintBackRefs{\CurrentBib}

\bibitem [\protect \citeauthoryear {%
Hady%
\ \BBA {} Schwenker%
}{%
Hady%
\ \BBA {} Schwenker%
}{%
{\protect \APACyear {2008}}%
}]{%
cocomittee}
\APACinsertmetastar {%
cocomittee}%
\begin{APACrefauthors}%
Hady, M\BPBI F\BPBI A.%
\BCBT {}\ \BBA {} Schwenker, F.%
\end{APACrefauthors}%
\unskip\
\newblock
\APACrefYearMonthDay{2008}{}{}.
\newblock
{\BBOQ}\APACrefatitle {{Co-training by Committee: {A} New Semi-supervised Learning Framework}} {{Co-training by Committee: {A} New Semi-supervised Learning Framework}}.{\BBCQ}
\newblock
\BIn{} \APACrefbtitle {Workshops Proceedings of the 8th {IEEE} International Conference on Data Mining {(ICDM} 2008), December 15-19, 2008, Pisa, Italy} {Workshops proceedings of the 8th {IEEE} international conference on data mining {(ICDM} 2008), december 15-19, 2008, pisa, italy}\ (\BPGS\ 563--572).
\newblock
\APACaddressPublisher{}{{IEEE} Computer Society}.
\newblock
\begin{APACrefURL} \url{https://doi.org/10.1109/ICDMW.2008.27} \end{APACrefURL}
\newblock
\begin{APACrefDOI} \doi{10.1109/ICDMW.2008.27} \end{APACrefDOI}
\PrintBackRefs{\CurrentBib}

\bibitem [\protect \citeauthoryear {%
Han%
\ \protect \BOthers {.}}{%
Han%
\ \protect \BOthers {.}}{%
{\protect \APACyear {2020}}%
}]{%
Noise:Surveys:Representation}
\APACinsertmetastar {%
Noise:Surveys:Representation}%
\begin{APACrefauthors}%
Han, B.%
, Yao, Q.%
, Liu, T.%
, Niu, G.%
, Tsang, I\BPBI W.%
, Kwok, J\BPBI T.%
\BCBL {}\ \BBA {} Sugiyama, M.%
\end{APACrefauthors}%
\unskip\
\newblock
\APACrefYearMonthDay{2020}{}{}.
\newblock
{\BBOQ}\APACrefatitle {A Survey of Label-noise Representation Learning: Past, Present and Future} {A survey of label-noise representation learning: Past, present and future}.{\BBCQ}
\newblock
\APACjournalVolNumPages{CoRR}{abs/2011.04406}{}{}.
\newblock
\begin{APACrefURL} \url{https://arxiv.org/abs/2011.04406} \end{APACrefURL}
\PrintBackRefs{\CurrentBib}

\bibitem [\protect \citeauthoryear {%
Han%
\ \protect \BOthers {.}}{%
Han%
\ \protect \BOthers {.}}{%
{\protect \APACyear {2018}}%
}]{%
Noise:Label:Co-Teaching}
\APACinsertmetastar {%
Noise:Label:Co-Teaching}%
\begin{APACrefauthors}%
Han, B.%
, Yao, Q.%
, Yu, X.%
, Niu, G.%
, Xu, M.%
, Hu, W.%
\BDBL {}Sugiyama, M.%
\end{APACrefauthors}%
\unskip\
\newblock
\APACrefYearMonthDay{2018}{}{}.
\newblock
{\BBOQ}\APACrefatitle {{Co-teaching: Robust training of deep neural networks with extremely noisy labels}} {{Co-teaching: Robust training of deep neural networks with extremely noisy labels}}.{\BBCQ}
\newblock
\BIn{} S.~Bengio, H\BPBI M.~Wallach, H.~Larochelle, K.~Grauman, N.~Cesa{-}Bianchi\BCBL {}\ \BBA {} R.~Garnett\ (\BEDS), \APACrefbtitle {{Advances in Neural Information Processing Systems 31: Annual Conference on Neural Information Processing Systems 2018, NeurIPS 2018, December 3-8, 2018, Montr{\'{e}}al, Canada}} {{Advances in Neural Information Processing Systems 31: Annual Conference on Neural Information Processing Systems 2018, NeurIPS 2018, December 3-8, 2018, Montr{\'{e}}al, Canada}}\ (\BPGS\ 8536--8546).
\newblock
\begin{APACrefURL} \url{https://proceedings.neurips.cc/paper/2018/hash/a19744e268754fb0148b017647355b7b-Abstract.html} \end{APACrefURL}
\PrintBackRefs{\CurrentBib}

\bibitem [\protect \citeauthoryear {%
Heidari%
\ \BBA {} Guo%
}{%
Heidari%
\ \BBA {} Guo%
}{%
{\protect \APACyear {2025}}%
}]{%
BLOPL}
\APACinsertmetastar {%
BLOPL}%
\begin{APACrefauthors}%
Heidari, M.%
\BCBT {}\ \BBA {} Guo, Y.%
\end{APACrefauthors}%
\unskip\
\newblock
\APACrefYearMonthDay{2025}{}{}.
\newblock
\APACrefbtitle {Bi-Level Optimization for Pseudo-Labeling Based Semi-Supervised Learning.} {Bi-level optimization for pseudo-labeling based semi-supervised learning.}
\newblock
\begin{APACrefURL} \url{https://openreview.net/forum?id=AEi2wyAMyb} \end{APACrefURL}
\PrintBackRefs{\CurrentBib}

\bibitem [\protect \citeauthoryear {%
Heidari%
, Zhang%
\BCBL {}\ \BBA {} Guo%
}{%
Heidari%
\ \protect \BOthers {.}}{%
{\protect \APACyear {2024}}%
}]{%
RLGSSL}
\APACinsertmetastar {%
RLGSSL}%
\begin{APACrefauthors}%
Heidari, M.%
, Zhang, H.%
\BCBL {}\ \BBA {} Guo, Y.%
\end{APACrefauthors}%
\unskip\
\newblock
\APACrefYearMonthDay{2024}{}{}.
\newblock
\APACrefbtitle {Reinforcement Learning-Guided Semi-Supervised Learning.} {Reinforcement learning-guided semi-supervised learning.}
\newblock
\begin{APACrefURL} \url{https://arxiv.org/abs/2405.01760} \end{APACrefURL}
\PrintBackRefs{\CurrentBib}

\bibitem [\protect \citeauthoryear {%
Hinton%
, Vinyals%
\BCBL {}\ \BBA {} Dean%
}{%
Hinton%
\ \protect \BOthers {.}}{%
{\protect \APACyear {2015}}%
}]{%
KD-hinton}
\APACinsertmetastar {%
KD-hinton}%
\begin{APACrefauthors}%
Hinton, G\BPBI E.%
, Vinyals, O.%
\BCBL {}\ \BBA {} Dean, J.%
\end{APACrefauthors}%
\unskip\
\newblock
\APACrefYearMonthDay{2015}{}{}.
\newblock
{\BBOQ}\APACrefatitle {{Distilling the Knowledge in a Neural Network}} {{Distilling the Knowledge in a Neural Network}}.{\BBCQ}
\newblock
\APACjournalVolNumPages{CoRR}{abs/1503.02531}{}{}.
\newblock
\begin{APACrefURL} \url{http://arxiv.org/abs/1503.02531} \end{APACrefURL}
\PrintBackRefs{\CurrentBib}

\bibitem [\protect \citeauthoryear {%
Iscen%
, Tolias%
, Avrithis%
\BCBL {}\ \BBA {} Chum%
}{%
Iscen%
\ \protect \BOthers {.}}{%
{\protect \APACyear {2019}}%
}]{%
LabelPropPL}
\APACinsertmetastar {%
LabelPropPL}%
\begin{APACrefauthors}%
Iscen, A.%
, Tolias, G.%
, Avrithis, Y.%
\BCBL {}\ \BBA {} Chum, O.%
\end{APACrefauthors}%
\unskip\
\newblock
\APACrefYearMonthDay{2019}{}{}.
\newblock
{\BBOQ}\APACrefatitle {{Label Propagation for Deep Semi-Supervised Learning}} {{Label Propagation for Deep Semi-Supervised Learning}}.{\BBCQ}
\newblock
\BIn{} \APACrefbtitle {{IEEE} Conference on Computer Vision and Pattern Recognition, {CVPR} 2019, Long Beach, CA, USA, June 16-20, 2019} {{IEEE} conference on computer vision and pattern recognition, {CVPR} 2019, long beach, ca, usa, june 16-20, 2019}\ (\BPGS\ 5070--5079).
\newblock
\APACaddressPublisher{}{Computer Vision Foundation / {IEEE}}.
\newblock
\begin{APACrefURL} \url{http://openaccess.thecvf.com/content\_CVPR\_2019/html/Iscen\_Label\_Propagation\_for\_Deep\_Semi-Supervised\_Learning\_CVPR\_2019\_paper.html} \end{APACrefURL}
\newblock
\begin{APACrefDOI} \doi{10.1109/CVPR.2019.00521} \end{APACrefDOI}
\PrintBackRefs{\CurrentBib}

\bibitem [\protect \citeauthoryear {%
James%
, Witten%
, Hastie%
, Tibshirani%
\BCBL {}\ \BBA {} Taylor%
}{%
James%
\ \protect \BOthers {.}}{%
{\protect \APACyear {2023}}%
}]{%
Book:StatLearning:Python}
\APACinsertmetastar {%
Book:StatLearning:Python}%
\begin{APACrefauthors}%
James, G.%
, Witten, D.%
, Hastie, T.%
, Tibshirani, R.%
\BCBL {}\ \BBA {} Taylor, J.%
\end{APACrefauthors}%
\unskip\
\newblock
\APACrefYear{2023}.
\newblock
\APACrefbtitle {{An introduction to statistical learning: With applications in python}} {{An introduction to statistical learning: With applications in python}}.
\newblock
\APACaddressPublisher{}{Springer Nature}.
\PrintBackRefs{\CurrentBib}

\bibitem [\protect \citeauthoryear {%
Jia%
\ \protect \BOthers {.}}{%
Jia%
\ \protect \BOthers {.}}{%
{\protect \APACyear {2021}}%
}]{%
ALIGN}
\APACinsertmetastar {%
ALIGN}%
\begin{APACrefauthors}%
Jia, C.%
, Yang, Y.%
, Xia, Y.%
, Chen, Y.%
, Parekh, Z.%
, Pham, H.%
\BDBL {}Duerig, T.%
\end{APACrefauthors}%
\unskip\
\newblock
\APACrefYearMonthDay{2021}{}{}.
\newblock
{\BBOQ}\APACrefatitle {{Scaling Up Visual and Vision-Language Representation Learning With Noisy Text Supervision}} {{Scaling Up Visual and Vision-Language Representation Learning With Noisy Text Supervision}}.{\BBCQ}
\newblock
\BIn{} M.~Meila\ \BBA {} T.~Zhang\ (\BEDS), \APACrefbtitle {Proceedings of the 38th International Conference on Machine Learning, {ICML} 2021, 18-24 July 2021, Virtual Event} {Proceedings of the 38th international conference on machine learning, {ICML} 2021, 18-24 july 2021, virtual event}\ (\BVOL~139, \BPGS\ 4904--4916).
\newblock
\APACaddressPublisher{}{{PMLR}}.
\newblock
\begin{APACrefURL} \url{http://proceedings.mlr.press/v139/jia21b.html} \end{APACrefURL}
\PrintBackRefs{\CurrentBib}

\bibitem [\protect \citeauthoryear {%
Jin%
\ \protect \BOthers {.}}{%
Jin%
\ \protect \BOthers {.}}{%
{\protect \APACyear {2019}}%
}]{%
RouteConstrainedKD}
\APACinsertmetastar {%
RouteConstrainedKD}%
\begin{APACrefauthors}%
Jin, X.%
, Peng, B.%
, Wu, Y.%
, Liu, Y.%
, Liu, J.%
, Liang, D.%
\BDBL {}Hu, X.%
\end{APACrefauthors}%
\unskip\
\newblock
\APACrefYearMonthDay{2019}{}{}.
\newblock
{\BBOQ}\APACrefatitle {{Knowledge Distillation via Route Constrained Optimization}} {{Knowledge Distillation via Route Constrained Optimization}}.{\BBCQ}
\newblock
\BIn{} \APACrefbtitle {2019 {IEEE/CVF} International Conference on Computer Vision, {ICCV} 2019, Seoul, Korea (South), October 27 - November 2, 2019} {2019 {IEEE/CVF} international conference on computer vision, {ICCV} 2019, seoul, korea (south), october 27 - november 2, 2019}\ (\BPGS\ 1345--1354).
\newblock
\APACaddressPublisher{}{{IEEE}}.
\newblock
\begin{APACrefURL} \url{https://doi.org/10.1109/ICCV.2019.00143} \end{APACrefURL}
\newblock
\begin{APACrefDOI} \doi{10.1109/ICCV.2019.00143} \end{APACrefDOI}
\PrintBackRefs{\CurrentBib}

\bibitem [\protect \citeauthoryear {%
Jing%
\ \BBA {} Tian%
}{%
Jing%
\ \BBA {} Tian%
}{%
{\protect \APACyear {2021}}%
}]{%
SSSurvey}
\APACinsertmetastar {%
SSSurvey}%
\begin{APACrefauthors}%
Jing, L.%
\BCBT {}\ \BBA {} Tian, Y.%
\end{APACrefauthors}%
\unskip\
\newblock
\APACrefYearMonthDay{2021}{}{}.
\newblock
{\BBOQ}\APACrefatitle {{Self-Supervised Visual Feature Learning With Deep Neural Networks: {A} Survey}} {{Self-Supervised Visual Feature Learning With Deep Neural Networks: {A} Survey}}.{\BBCQ}
\newblock
\APACjournalVolNumPages{{IEEE} Trans. Pattern Anal. Mach. Intell.}{43}{11}{4037--4058}.
\newblock
\begin{APACrefURL} \url{https://doi.org/10.1109/TPAMI.2020.2992393} \end{APACrefURL}
\newblock
\begin{APACrefDOI} \doi{10.1109/TPAMI.2020.2992393} \end{APACrefDOI}
\PrintBackRefs{\CurrentBib}

\bibitem [\protect \citeauthoryear {%
Kage%
\ \BBA {} Andreadis%
}{%
Kage%
\ \BBA {} Andreadis%
}{%
{\protect \APACyear {2021}}%
}]{%
ClassIntrospection}
\APACinsertmetastar {%
ClassIntrospection}%
\begin{APACrefauthors}%
Kage, P.%
\BCBT {}\ \BBA {} Andreadis, P.%
\end{APACrefauthors}%
\unskip\
\newblock
\APACrefYearMonthDay{2021}{}{}.
\newblock
{\BBOQ}\APACrefatitle {Class Introspection: {A} Novel Technique for Detecting Unlabeled Subclasses by Leveraging Classifier Explainability Methods} {Class introspection: {A} novel technique for detecting unlabeled subclasses by leveraging classifier explainability methods}.{\BBCQ}
\newblock
\APACjournalVolNumPages{CoRR}{abs/2107.01657}{}{}.
\newblock
\begin{APACrefURL} \url{https://arxiv.org/abs/2107.01657} \end{APACrefURL}
\PrintBackRefs{\CurrentBib}

\bibitem [\protect \citeauthoryear {%
Kolesnikov%
, Zhai%
\BCBL {}\ \BBA {} Beyer%
}{%
Kolesnikov%
\ \protect \BOthers {.}}{%
{\protect \APACyear {2019}}%
}]{%
kolesnikov_revisiting_2019}
\APACinsertmetastar {%
kolesnikov_revisiting_2019}%
\begin{APACrefauthors}%
Kolesnikov, A.%
, Zhai, X.%
\BCBL {}\ \BBA {} Beyer, L.%
\end{APACrefauthors}%
\unskip\
\newblock
\APACrefYearMonthDay{2019}{}{}.
\newblock
{\BBOQ}\APACrefatitle {{Revisiting Self-Supervised Visual Representation Learning}} {{Revisiting Self-Supervised Visual Representation Learning}}.{\BBCQ}
\newblock
\BIn{} \APACrefbtitle {{IEEE} Conference on Computer Vision and Pattern Recognition, {CVPR} 2019, Long Beach, CA, USA, June 16-20, 2019} {{IEEE} conference on computer vision and pattern recognition, {CVPR} 2019, long beach, ca, usa, june 16-20, 2019}\ (\BPGS\ 1920--1929).
\newblock
\APACaddressPublisher{}{Computer Vision Foundation / {IEEE}}.
\newblock
\begin{APACrefURL} \url{http://openaccess.thecvf.com/content\_CVPR\_2019/html/Kolesnikov\_Revisiting\_Self-Supervised\_Visual\_Representation\_Learning\_CVPR\_2019\_paper.html} \end{APACrefURL}
\newblock
\begin{APACrefDOI} \doi{10.1109/CVPR.2019.00202} \end{APACrefDOI}
\PrintBackRefs{\CurrentBib}

\bibitem [\protect \citeauthoryear {%
Laine%
\ \BBA {} Aila%
}{%
Laine%
\ \BBA {} Aila%
}{%
{\protect \APACyear {2017}}%
}]{%
PiModel}
\APACinsertmetastar {%
PiModel}%
\begin{APACrefauthors}%
Laine, S.%
\BCBT {}\ \BBA {} Aila, T.%
\end{APACrefauthors}%
\unskip\
\newblock
\APACrefYearMonthDay{2017}{}{}.
\newblock
{\BBOQ}\APACrefatitle {{Temporal Ensembling for Semi-Supervised Learning}} {{Temporal Ensembling for Semi-Supervised Learning}}.{\BBCQ}
\newblock
\BIn{} \APACrefbtitle {5th International Conference on Learning Representations, {ICLR} 2017, Toulon, France, April 24-26, 2017, Conference Track Proceedings.} {5th international conference on learning representations, {ICLR} 2017, toulon, france, april 24-26, 2017, conference track proceedings.}
\newblock
\APACaddressPublisher{}{OpenReview.net}.
\newblock
\begin{APACrefURL} \url{https://openreview.net/forum?id=BJ6oOfqge} \end{APACrefURL}
\PrintBackRefs{\CurrentBib}

\bibitem [\protect \citeauthoryear {%
Lakshminarayanan%
, Pritzel%
\BCBL {}\ \BBA {} Blundell%
}{%
Lakshminarayanan%
\ \protect \BOthers {.}}{%
{\protect \APACyear {2017}}%
}]{%
SSUncertainty}
\APACinsertmetastar {%
SSUncertainty}%
\begin{APACrefauthors}%
Lakshminarayanan, B.%
, Pritzel, A.%
\BCBL {}\ \BBA {} Blundell, C.%
\end{APACrefauthors}%
\unskip\
\newblock
\APACrefYearMonthDay{2017}{}{}.
\newblock
{\BBOQ}\APACrefatitle {{Simple and Scalable Predictive Uncertainty Estimation using Deep Ensembles}} {{Simple and Scalable Predictive Uncertainty Estimation using Deep Ensembles}}.{\BBCQ}
\newblock
\BIn{} I.~Guyon\ \BOthers {.}\ (\BEDS), \APACrefbtitle {Advances in Neural Information Processing Systems 30: Annual Conference on Neural Information Processing Systems 2017, December 4-9, 2017, Long Beach, CA, {USA}} {Advances in neural information processing systems 30: Annual conference on neural information processing systems 2017, december 4-9, 2017, long beach, ca, {USA}}\ (\BPGS\ 6402--6413).
\newblock
\begin{APACrefURL} \url{https://proceedings.neurips.cc/paper/2017/hash/9ef2ed4b7fd2c810847ffa5fa85bce38-Abstract.html} \end{APACrefURL}
\PrintBackRefs{\CurrentBib}

\bibitem [\protect \citeauthoryear {%
Lee%
}{%
Lee%
}{%
{\protect \APACyear {2013}}%
}]{%
PL}
\APACinsertmetastar {%
PL}%
\begin{APACrefauthors}%
Lee, D\BHBI H.%
\end{APACrefauthors}%
\unskip\
\newblock
\APACrefYearMonthDay{2013}{{\APACmonth{07}}}{}.
\newblock
{\BBOQ}\APACrefatitle {{Pseudo-Label : {{The}} Simple and Efficient Semi-Supervised Learning Method for Deep Neural Networks}} {{Pseudo-Label : {{The}} Simple and Efficient Semi-Supervised Learning Method for Deep Neural Networks}}.{\BBCQ}
\newblock
\APACjournalVolNumPages{ICML 2013 Workshop : Challenges in Representation Learning (WREPL)}{}{}{}.
\PrintBackRefs{\CurrentBib}

\bibitem [\protect \citeauthoryear {%
C.~Li%
\ \protect \BOthers {.}}{%
C.~Li%
\ \protect \BOthers {.}}{%
{\protect \APACyear {2022}}%
}]{%
EsViT}
\APACinsertmetastar {%
EsViT}%
\begin{APACrefauthors}%
Li, C.%
, Yang, J.%
, Zhang, P.%
, Gao, M.%
, Xiao, B.%
, Dai, X.%
\BDBL {}Gao, J.%
\end{APACrefauthors}%
\unskip\
\newblock
\APACrefYearMonthDay{2022}{}{}.
\newblock
{\BBOQ}\APACrefatitle {{Efficient Self-supervised Vision Transformers for Representation Learning}} {{Efficient Self-supervised Vision Transformers for Representation Learning}}.{\BBCQ}
\newblock
\BIn{} \APACrefbtitle {The Tenth International Conference on Learning Representations, {ICLR} 2022, Virtual Event, April 25-29, 2022.} {The tenth international conference on learning representations, {ICLR} 2022, virtual event, april 25-29, 2022.}
\newblock
\APACaddressPublisher{}{OpenReview.net}.
\newblock
\begin{APACrefURL} \url{https://openreview.net/forum?id=fVu3o-YUGQK} \end{APACrefURL}
\PrintBackRefs{\CurrentBib}

\bibitem [\protect \citeauthoryear {%
Z.~Li%
\ \protect \BOthers {.}}{%
Z.~Li%
\ \protect \BOthers {.}}{%
{\protect \APACyear {2023}}%
}]{%
CurriculumTemperatureKD}
\APACinsertmetastar {%
CurriculumTemperatureKD}%
\begin{APACrefauthors}%
Li, Z.%
, Li, X.%
, Yang, L.%
, Zhao, B.%
, Song, R.%
, Luo, L.%
\BDBL {}Yang, J.%
\end{APACrefauthors}%
\unskip\
\newblock
\APACrefYearMonthDay{2023}{}{}.
\newblock
{\BBOQ}\APACrefatitle {{Curriculum Temperature for Knowledge Distillation}} {{Curriculum Temperature for Knowledge Distillation}}.{\BBCQ}
\newblock
\BIn{} B.~Williams, Y.~Chen\BCBL {}\ \BBA {} J.~Neville\ (\BEDS), \APACrefbtitle {Thirty-Seventh {AAAI} Conference on Artificial Intelligence, {AAAI} 2023, Thirty-Fifth Conference on Innovative Applications of Artificial Intelligence, {IAAI} 2023, Thirteenth Symposium on Educational Advances in Artificial Intelligence, {EAAI} 2023, Washington, DC, USA, February 7-14, 2023} {Thirty-seventh {AAAI} conference on artificial intelligence, {AAAI} 2023, thirty-fifth conference on innovative applications of artificial intelligence, {IAAI} 2023, thirteenth symposium on educational advances in artificial intelligence, {EAAI} 2023, washington, dc, usa, february 7-14, 2023}\ (\BPGS\ 1504--1512).
\newblock
\APACaddressPublisher{}{{AAAI} Press}.
\newblock
\begin{APACrefURL} \url{https://doi.org/10.1609/aaai.v37i2.25236} \end{APACrefURL}
\newblock
\begin{APACrefDOI} \doi{10.1609/AAAI.V37I2.25236} \end{APACrefDOI}
\PrintBackRefs{\CurrentBib}

\bibitem [\protect \citeauthoryear {%
Liu%
, Wang%
, Owens%
\BCBL {}\ \BBA {} Li%
}{%
Liu%
\ \protect \BOthers {.}}{%
{\protect \APACyear {2021}}%
}]{%
FreeEnergy}
\APACinsertmetastar {%
FreeEnergy}%
\begin{APACrefauthors}%
Liu, W.%
, Wang, X.%
, Owens, J\BPBI D.%
\BCBL {}\ \BBA {} Li, Y.%
\end{APACrefauthors}%
\unskip\
\newblock
\APACrefYearMonthDay{2021}{{\APACmonth{04}}}{}.
\newblock
\APACrefbtitle {Energy-Based {{Out-of-distribution Detection}}} {Energy-based {{Out-of-distribution Detection}}}\ (\BNUM\ arXiv:2010.03759).
\newblock
\APACaddressPublisher{}{arXiv}.
\newblock
\begin{APACrefDOI} \doi{10.48550/arXiv.2010.03759} \end{APACrefDOI}
\PrintBackRefs{\CurrentBib}

\bibitem [\protect \citeauthoryear {%
Lopes%
, Fenu%
\BCBL {}\ \BBA {} Starner%
}{%
Lopes%
\ \protect \BOthers {.}}{%
{\protect \APACyear {2017}}%
}]{%
DataFreeKD}
\APACinsertmetastar {%
DataFreeKD}%
\begin{APACrefauthors}%
Lopes, R\BPBI G.%
, Fenu, S.%
\BCBL {}\ \BBA {} Starner, T.%
\end{APACrefauthors}%
\unskip\
\newblock
\APACrefYearMonthDay{2017}{}{}.
\newblock
{\BBOQ}\APACrefatitle {{Data-Free Knowledge Distillation for Deep Neural Networks}} {{Data-Free Knowledge Distillation for Deep Neural Networks}}.{\BBCQ}
\newblock
\APACjournalVolNumPages{CoRR}{abs/1710.07535}{}{}.
\newblock
\begin{APACrefURL} \url{http://arxiv.org/abs/1710.07535} \end{APACrefURL}
\PrintBackRefs{\CurrentBib}

\bibitem [\protect \citeauthoryear {%
Louizos%
\ \BBA {} Welling%
}{%
Louizos%
\ \BBA {} Welling%
}{%
{\protect \APACyear {2016}}%
}]{%
GaussianPosteriors}
\APACinsertmetastar {%
GaussianPosteriors}%
\begin{APACrefauthors}%
Louizos, C.%
\BCBT {}\ \BBA {} Welling, M.%
\end{APACrefauthors}%
\unskip\
\newblock
\APACrefYearMonthDay{2016}{}{}.
\newblock
{\BBOQ}\APACrefatitle {{Structured and Efficient Variational Deep Learning with Matrix Gaussian Posteriors}} {{Structured and Efficient Variational Deep Learning with Matrix Gaussian Posteriors}}.{\BBCQ}
\newblock
\BIn{} M.~Balcan\ \BBA {} K\BPBI Q.~Weinberger\ (\BEDS), \APACrefbtitle {Proceedings of the 33nd International Conference on Machine Learning, {ICML} 2016, New York City, NY, USA, June 19-24, 2016} {Proceedings of the 33nd international conference on machine learning, {ICML} 2016, new york city, ny, usa, june 19-24, 2016}\ (\BVOL~48, \BPGS\ 1708--1716).
\newblock
\APACaddressPublisher{}{JMLR.org}.
\newblock
\begin{APACrefURL} \url{http://proceedings.mlr.press/v48/louizos16.html} \end{APACrefURL}
\PrintBackRefs{\CurrentBib}

\bibitem [\protect \citeauthoryear {%
McLachlan%
}{%
McLachlan%
}{%
{\protect \APACyear {1975}}%
}]{%
SelfTrain}
\APACinsertmetastar {%
SelfTrain}%
\begin{APACrefauthors}%
McLachlan, G\BPBI J.%
\end{APACrefauthors}%
\unskip\
\newblock
\APACrefYearMonthDay{1975}{}{}.
\newblock
{\BBOQ}\APACrefatitle {{Iterative Reclassification Procedure for Constructing an Asymptotically Optimal Rule of Allocation in Discriminant Analysis}} {{Iterative Reclassification Procedure for Constructing an Asymptotically Optimal Rule of Allocation in Discriminant Analysis}}.{\BBCQ}
\newblock
\APACjournalVolNumPages{Journal of the American Statistical Association}{70}{350}{365--369}.
\newblock
\begin{APACrefURL} [{2024-01-31}]\url{http://www.jstor.org/stable/2285824} \end{APACrefURL}
\PrintBackRefs{\CurrentBib}

\bibitem [\protect \citeauthoryear {%
Minderer%
, Gritsenko%
\BCBL {}\ \BBA {} Houlsby%
}{%
Minderer%
\ \protect \BOthers {.}}{%
{\protect \APACyear {2023}}%
}]{%
OVOD}
\APACinsertmetastar {%
OVOD}%
\begin{APACrefauthors}%
Minderer, M.%
, Gritsenko, A\BPBI A.%
\BCBL {}\ \BBA {} Houlsby, N.%
\end{APACrefauthors}%
\unskip\
\newblock
\APACrefYearMonthDay{2023}{}{}.
\newblock
{\BBOQ}\APACrefatitle {{Scaling Open-Vocabulary Object Detection}} {{Scaling Open-Vocabulary Object Detection}}.{\BBCQ}
\newblock
\APACjournalVolNumPages{CoRR}{abs/2306.09683}{}{}.
\newblock
\begin{APACrefURL} \url{https://doi.org/10.48550/arXiv.2306.09683} \end{APACrefURL}
\newblock
\begin{APACrefDOI} \doi{10.48550/ARXIV.2306.09683} \end{APACrefDOI}
\PrintBackRefs{\CurrentBib}

\bibitem [\protect \citeauthoryear {%
Mitchell%
}{%
Mitchell%
}{%
{\protect \APACyear {1997}}%
}]{%
Book:Mitchell}
\APACinsertmetastar {%
Book:Mitchell}%
\begin{APACrefauthors}%
Mitchell, T\BPBI M.%
\end{APACrefauthors}%
\unskip\
\newblock
\APACrefYear{1997}.
\newblock
\APACrefbtitle {{Machine learning, International Edition}} {{Machine learning, International Edition}}.
\newblock
\APACaddressPublisher{}{McGraw-Hill}.
\newblock
\begin{APACrefURL} \url{https://www.worldcat.org/oclc/61321007} \end{APACrefURL}
\PrintBackRefs{\CurrentBib}

\bibitem [\protect \citeauthoryear {%
Miyato%
, Maeda%
, Koyama%
\BCBL {}\ \BBA {} Ishii%
}{%
Miyato%
\ \protect \BOthers {.}}{%
{\protect \APACyear {2019}}%
}]{%
VAT}
\APACinsertmetastar {%
VAT}%
\begin{APACrefauthors}%
Miyato, T.%
, Maeda, S.%
, Koyama, M.%
\BCBL {}\ \BBA {} Ishii, S.%
\end{APACrefauthors}%
\unskip\
\newblock
\APACrefYearMonthDay{2019}{}{}.
\newblock
{\BBOQ}\APACrefatitle {{Virtual Adversarial Training: {A} Regularization Method for Supervised and Semi-Supervised Learning}} {{Virtual Adversarial Training: {A} Regularization Method for Supervised and Semi-Supervised Learning}}.{\BBCQ}
\newblock
\APACjournalVolNumPages{{IEEE} Trans. Pattern Anal. Mach. Intell.}{41}{8}{1979--1993}.
\newblock
\begin{APACrefURL} \url{https://doi.org/10.1109/TPAMI.2018.2858821} \end{APACrefURL}
\newblock
\begin{APACrefDOI} \doi{10.1109/TPAMI.2018.2858821} \end{APACrefDOI}
\PrintBackRefs{\CurrentBib}

\bibitem [\protect \citeauthoryear {%
Ouali%
, Hudelot%
\BCBL {}\ \BBA {} Tami%
}{%
Ouali%
\ \protect \BOthers {.}}{%
{\protect \APACyear {2020}}%
}]{%
ouali_overview_2020}
\APACinsertmetastar {%
ouali_overview_2020}%
\begin{APACrefauthors}%
Ouali, Y.%
, Hudelot, C.%
\BCBL {}\ \BBA {} Tami, M.%
\end{APACrefauthors}%
\unskip\
\newblock
\APACrefYearMonthDay{2020}{}{}.
\newblock
{\BBOQ}\APACrefatitle {{An Overview of Deep Semi-Supervised Learning}} {{An Overview of Deep Semi-Supervised Learning}}.{\BBCQ}
\newblock
\APACjournalVolNumPages{CoRR}{abs/2006.05278}{}{}.
\newblock
\begin{APACrefURL} \url{https://arxiv.org/abs/2006.05278} \end{APACrefURL}
\PrintBackRefs{\CurrentBib}

\bibitem [\protect \citeauthoryear {%
Patrini%
, Rozza%
, Menon%
, Nock%
\BCBL {}\ \BBA {} Qu%
}{%
Patrini%
\ \protect \BOthers {.}}{%
{\protect \APACyear {2017}}%
}]{%
Noise:Label:Inspired:Recent}
\APACinsertmetastar {%
Noise:Label:Inspired:Recent}%
\begin{APACrefauthors}%
Patrini, G.%
, Rozza, A.%
, Menon, A\BPBI K.%
, Nock, R.%
\BCBL {}\ \BBA {} Qu, L.%
\end{APACrefauthors}%
\unskip\
\newblock
\APACrefYearMonthDay{2017}{}{}.
\newblock
{\BBOQ}\APACrefatitle {{Making Deep Neural Networks Robust to Label Noise: {A} Loss Correction Approach}} {{Making Deep Neural Networks Robust to Label Noise: {A} Loss Correction Approach}}.{\BBCQ}
\newblock
\BIn{} \APACrefbtitle {2017 {IEEE} Conference on Computer Vision and Pattern Recognition, {CVPR} 2017, Honolulu, HI, USA, July 21-26, 2017} {2017 {IEEE} conference on computer vision and pattern recognition, {CVPR} 2017, honolulu, hi, usa, july 21-26, 2017}\ (\BPGS\ 2233--2241).
\newblock
\APACaddressPublisher{}{{IEEE} Computer Society}.
\newblock
\begin{APACrefURL} \url{https://doi.org/10.1109/CVPR.2017.240} \end{APACrefURL}
\newblock
\begin{APACrefDOI} \doi{10.1109/CVPR.2017.240} \end{APACrefDOI}
\PrintBackRefs{\CurrentBib}

\bibitem [\protect \citeauthoryear {%
Pham%
, Dai%
, Xie%
\BCBL {}\ \BBA {} Le%
}{%
Pham%
\ \protect \BOthers {.}}{%
{\protect \APACyear {2021}}%
}]{%
MetaPL}
\APACinsertmetastar {%
MetaPL}%
\begin{APACrefauthors}%
Pham, H.%
, Dai, Z.%
, Xie, Q.%
\BCBL {}\ \BBA {} Le, Q\BPBI V.%
\end{APACrefauthors}%
\unskip\
\newblock
\APACrefYearMonthDay{2021}{}{}.
\newblock
{\BBOQ}\APACrefatitle {{Meta Pseudo Labels}} {{Meta Pseudo Labels}}.{\BBCQ}
\newblock
\BIn{} \APACrefbtitle {{IEEE} Conference on Computer Vision and Pattern Recognition, {CVPR} 2021, virtual, June 19-25, 2021} {{IEEE} conference on computer vision and pattern recognition, {CVPR} 2021, virtual, june 19-25, 2021}\ (\BPGS\ 11557--11568).
\newblock
\APACaddressPublisher{}{Computer Vision Foundation / {IEEE}}.
\newblock
\begin{APACrefURL} \url{https://openaccess.thecvf.com/content/CVPR2021/html/Pham\_Meta\_Pseudo\_Labels\_CVPR\_2021\_paper.html} \end{APACrefURL}
\newblock
\begin{APACrefDOI} \doi{10.1109/CVPR46437.2021.01139} \end{APACrefDOI}
\PrintBackRefs{\CurrentBib}

\bibitem [\protect \citeauthoryear {%
Pise%
\ \BBA {} Kulkarni%
}{%
Pise%
\ \BBA {} Kulkarni%
}{%
{\protect \APACyear {2008}}%
}]{%
pise2008survey}
\APACinsertmetastar {%
pise2008survey}%
\begin{APACrefauthors}%
Pise, N\BPBI N.%
\BCBT {}\ \BBA {} Kulkarni, P\BPBI A.%
\end{APACrefauthors}%
\unskip\
\newblock
\APACrefYearMonthDay{2008}{}{}.
\newblock
{\BBOQ}\APACrefatitle {{A Survey of Semi-Supervised Learning Methods}} {{A Survey of Semi-Supervised Learning Methods}}.{\BBCQ}
\newblock
\BIn{} \APACrefbtitle {2008 International Conference on Computational Intelligence and Security, {CIS} 2008, 13-17 December 2008, Suzhou, China, Volume 2, Workshop Papers} {2008 international conference on computational intelligence and security, {CIS} 2008, 13-17 december 2008, suzhou, china, volume 2, workshop papers}\ (\BPGS\ 30--34).
\newblock
\APACaddressPublisher{}{{IEEE} Computer Society}.
\newblock
\begin{APACrefURL} \url{https://doi.org/10.1109/CIS.2008.204} \end{APACrefURL}
\newblock
\begin{APACrefDOI} \doi{10.1109/CIS.2008.204} \end{APACrefDOI}
\PrintBackRefs{\CurrentBib}

\bibitem [\protect \citeauthoryear {%
Prakash%
\ \BBA {} Nithya%
}{%
Prakash%
\ \BBA {} Nithya%
}{%
{\protect \APACyear {2014}}%
}]{%
prakash2014survey}
\APACinsertmetastar {%
prakash2014survey}%
\begin{APACrefauthors}%
Prakash, V\BPBI J.%
\BCBT {}\ \BBA {} Nithya, L\BPBI M.%
\end{APACrefauthors}%
\unskip\
\newblock
\APACrefYearMonthDay{2014}{}{}.
\newblock
{\BBOQ}\APACrefatitle {{A Survey on Semi-Supervised Learning Techniques}} {{A Survey on Semi-Supervised Learning Techniques}}.{\BBCQ}
\newblock
\APACjournalVolNumPages{CoRR}{abs/1402.4645}{}{}.
\newblock
\begin{APACrefURL} \url{http://arxiv.org/abs/1402.4645} \end{APACrefURL}
\PrintBackRefs{\CurrentBib}

\bibitem [\protect \citeauthoryear {%
Qiao%
, Shen%
, Zhang%
, Wang%
\BCBL {}\ \BBA {} Yuille%
}{%
Qiao%
\ \protect \BOthers {.}}{%
{\protect \APACyear {2018}}%
}]{%
deepcotraining}
\APACinsertmetastar {%
deepcotraining}%
\begin{APACrefauthors}%
Qiao, S.%
, Shen, W.%
, Zhang, Z.%
, Wang, B.%
\BCBL {}\ \BBA {} Yuille, A\BPBI L.%
\end{APACrefauthors}%
\unskip\
\newblock
\APACrefYearMonthDay{2018}{}{}.
\newblock
{\BBOQ}\APACrefatitle {Deep Co-Training for Semi-Supervised Image Recognition} {Deep co-training for semi-supervised image recognition}.{\BBCQ}
\newblock
\APACjournalVolNumPages{CoRR}{abs/1803.05984}{}{}.
\newblock
\begin{APACrefURL} \url{http://arxiv.org/abs/1803.05984} \end{APACrefURL}
\PrintBackRefs{\CurrentBib}

\bibitem [\protect \citeauthoryear {%
Radford%
\ \protect \BOthers {.}}{%
Radford%
\ \protect \BOthers {.}}{%
{\protect \APACyear {2021}}%
}]{%
CLIP}
\APACinsertmetastar {%
CLIP}%
\begin{APACrefauthors}%
Radford, A.%
, Kim, J\BPBI W.%
, Hallacy, C.%
, Ramesh, A.%
, Goh, G.%
, Agarwal, S.%
\BDBL {}Sutskever, I.%
\end{APACrefauthors}%
\unskip\
\newblock
\APACrefYearMonthDay{2021}{}{}.
\newblock
{\BBOQ}\APACrefatitle {{Learning Transferable Visual Models From Natural Language Supervision}} {{Learning Transferable Visual Models From Natural Language Supervision}}.{\BBCQ}
\newblock
\BIn{} M.~Meila\ \BBA {} T.~Zhang\ (\BEDS), \APACrefbtitle {Proceedings of the 38th International Conference on Machine Learning, {ICML} 2021, 18-24 July 2021, Virtual Event} {Proceedings of the 38th international conference on machine learning, {ICML} 2021, 18-24 july 2021, virtual event}\ (\BVOL~139, \BPGS\ 8748--8763).
\newblock
\APACaddressPublisher{}{{PMLR}}.
\newblock
\begin{APACrefURL} \url{http://proceedings.mlr.press/v139/radford21a.html} \end{APACrefURL}
\PrintBackRefs{\CurrentBib}

\bibitem [\protect \citeauthoryear {%
Rizve%
, Duarte%
, Rawat%
\BCBL {}\ \BBA {} Shah%
}{%
Rizve%
\ \protect \BOthers {.}}{%
{\protect \APACyear {2021}}%
}]{%
InDefenseOfPL}
\APACinsertmetastar {%
InDefenseOfPL}%
\begin{APACrefauthors}%
Rizve, M\BPBI N.%
, Duarte, K.%
, Rawat, Y\BPBI S.%
\BCBL {}\ \BBA {} Shah, M.%
\end{APACrefauthors}%
\unskip\
\newblock
\APACrefYearMonthDay{2021}{}{}.
\newblock
{\BBOQ}\APACrefatitle {{In Defense of Pseudo-Labeling: An Uncertainty-Aware Pseudo-label Selection Framework for Semi-Supervised Learning}} {{In Defense of Pseudo-Labeling: An Uncertainty-Aware Pseudo-label Selection Framework for Semi-Supervised Learning}}.{\BBCQ}
\newblock
\BIn{} \APACrefbtitle {9th International Conference on Learning Representations, {ICLR} 2021, Virtual Event, Austria, May 3-7, 2021.} {9th international conference on learning representations, {ICLR} 2021, virtual event, austria, may 3-7, 2021.}
\newblock
\APACaddressPublisher{}{OpenReview.net}.
\newblock
\begin{APACrefURL} \url{https://openreview.net/forum?id=-ODN6SbiUU} \end{APACrefURL}
\PrintBackRefs{\CurrentBib}

\bibitem [\protect \citeauthoryear {%
Rothenberger%
\ \BBA {} Diochnos%
}{%
Rothenberger%
\ \BBA {} Diochnos%
}{%
{\protect \APACyear {2023}}%
}]{%
MetaCotraining}
\APACinsertmetastar {%
MetaCotraining}%
\begin{APACrefauthors}%
Rothenberger, J\BPBI C.%
\BCBT {}\ \BBA {} Diochnos, D\BPBI I.%
\end{APACrefauthors}%
\unskip\
\newblock
\APACrefYearMonthDay{2023}{}{}.
\newblock
{\BBOQ}\APACrefatitle {{Meta Co-Training: Two Views are Better than One}} {{Meta Co-Training: Two Views are Better than One}}.{\BBCQ}
\newblock
\APACjournalVolNumPages{CoRR}{abs/2311.18083}{}{}.
\newblock
\begin{APACrefURL} \url{https://doi.org/10.48550/arXiv.2311.18083} \end{APACrefURL}
\newblock
\begin{APACrefDOI} \doi{10.48550/ARXIV.2311.18083} \end{APACrefDOI}
\PrintBackRefs{\CurrentBib}

\bibitem [\protect \citeauthoryear {%
Ruspini%
}{%
Ruspini%
}{%
{\protect \APACyear {1969}}%
}]{%
fuzzypart}
\APACinsertmetastar {%
fuzzypart}%
\begin{APACrefauthors}%
Ruspini, E\BPBI H.%
\end{APACrefauthors}%
\unskip\
\newblock
\APACrefYearMonthDay{1969}{}{}.
\newblock
{\BBOQ}\APACrefatitle {A new approach to clustering} {A new approach to clustering}.{\BBCQ}
\newblock
\APACjournalVolNumPages{Information and Control}{15}{1}{22-32}.
\newblock
\begin{APACrefURL} \url{https://www.sciencedirect.com/science/article/pii/S0019995869905919} \end{APACrefURL}
\newblock
\begin{APACrefDOI} \doi{https://doi.org/10.1016/S0019-9958(69)90591-9} \end{APACrefDOI}
\PrintBackRefs{\CurrentBib}

\bibitem [\protect \citeauthoryear {%
Samragh%
\ \protect \BOthers {.}}{%
Samragh%
\ \protect \BOthers {.}}{%
{\protect \APACyear {2023}}%
}]{%
directInit}
\APACinsertmetastar {%
directInit}%
\begin{APACrefauthors}%
Samragh, M.%
, Farajtabar, M.%
, Mehta, S.%
, Vemulapalli, R.%
, Faghri, F.%
, Naik, D.%
\BDBL {}Rastegari, M.%
\end{APACrefauthors}%
\unskip\
\newblock
\APACrefYearMonthDay{2023}{}{}.
\newblock
{\BBOQ}\APACrefatitle {{Weight subcloning: direct initialization of transformers using larger pretrained ones}} {{Weight subcloning: direct initialization of transformers using larger pretrained ones}}.{\BBCQ}
\newblock
\APACjournalVolNumPages{CoRR}{abs/2312.09299}{}{}.
\newblock
\begin{APACrefURL} \url{https://doi.org/10.48550/arXiv.2312.09299} \end{APACrefURL}
\newblock
\begin{APACrefDOI} \doi{10.48550/ARXIV.2312.09299} \end{APACrefDOI}
\PrintBackRefs{\CurrentBib}

\bibitem [\protect \citeauthoryear {%
Schuhmann%
\ \protect \BOthers {.}}{%
Schuhmann%
\ \protect \BOthers {.}}{%
{\protect \APACyear {2022}}%
}]{%
LAION5B}
\APACinsertmetastar {%
LAION5B}%
\begin{APACrefauthors}%
Schuhmann, C.%
, Beaumont, R.%
, Vencu, R.%
, Gordon, C.%
, Wightman, R.%
, Cherti, M.%
\BDBL {}Jitsev, J.%
\end{APACrefauthors}%
\unskip\
\newblock
\APACrefYearMonthDay{2022}{}{}.
\newblock
{\BBOQ}\APACrefatitle {{{LAION-5B:} An open large-scale dataset for training next generation image-text models}} {{{LAION-5B:} An open large-scale dataset for training next generation image-text models}}.{\BBCQ}
\newblock
\BIn{} S.~Koyejo, S.~Mohamed, A.~Agarwal, D.~Belgrave, K.~Cho\BCBL {}\ \BBA {} A.~Oh\ (\BEDS), \APACrefbtitle {Advances in Neural Information Processing Systems 35: Annual Conference on Neural Information Processing Systems 2022, NeurIPS 2022, New Orleans, LA, USA, November 28 - December 9, 2022.} {Advances in neural information processing systems 35: Annual conference on neural information processing systems 2022, neurips 2022, new orleans, la, usa, november 28 - december 9, 2022.}
\newblock
\begin{APACrefURL} \url{http://papers.nips.cc/paper\_files/paper/2022/hash/a1859debfb3b59d094f3504d5ebb6c25-Abstract-Datasets\_and\_Benchmarks.html} \end{APACrefURL}
\PrintBackRefs{\CurrentBib}

\bibitem [\protect \citeauthoryear {%
Shalev{-}Shwartz%
\ \BBA {} Ben{-}David%
}{%
Shalev{-}Shwartz%
\ \BBA {} Ben{-}David%
}{%
{\protect \APACyear {2014}}%
}]{%
Book:Understanding}
\APACinsertmetastar {%
Book:Understanding}%
\begin{APACrefauthors}%
Shalev{-}Shwartz, S.%
\BCBT {}\ \BBA {} Ben{-}David, S.%
\end{APACrefauthors}%
\unskip\
\newblock
\APACrefYear{2014}.
\newblock
\APACrefbtitle {{Understanding Machine Learning - From Theory to Algorithms}} {{Understanding Machine Learning - From Theory to Algorithms}}.
\newblock
\APACaddressPublisher{}{Cambridge University Press}.
\newblock
\begin{APACrefURL} \url{http://www.cambridge.org/de/academic/subjects/computer-science/pattern-recognition-and-machine-learning/understanding-machine-learning-theory-algorithms} \end{APACrefURL}
\PrintBackRefs{\CurrentBib}

\bibitem [\protect \citeauthoryear {%
Shi%
\ \protect \BOthers {.}}{%
Shi%
\ \protect \BOthers {.}}{%
{\protect \APACyear {2018}}%
}]{%
MMF}
\APACinsertmetastar {%
MMF}%
\begin{APACrefauthors}%
Shi, W.%
, Gong, Y.%
, Ding, C.%
, Ma, Z.%
, Tao, X.%
\BCBL {}\ \BBA {} Zheng, N.%
\end{APACrefauthors}%
\unskip\
\newblock
\APACrefYearMonthDay{2018}{}{}.
\newblock
{\BBOQ}\APACrefatitle {{Transductive Semi-Supervised Deep Learning Using Min-Max Features}} {{Transductive Semi-Supervised Deep Learning Using Min-Max Features}}.{\BBCQ}
\newblock
\BIn{} V.~Ferrari, M.~Hebert, C.~Sminchisescu\BCBL {}\ \BBA {} Y.~Weiss\ (\BEDS), \APACrefbtitle {Computer Vision - {ECCV} 2018 - 15th European Conference, Munich, Germany, September 8-14, 2018, Proceedings, Part {V}} {Computer vision - {ECCV} 2018 - 15th european conference, munich, germany, september 8-14, 2018, proceedings, part {V}}\ (\BVOL\ 11209, \BPGS\ 311--327).
\newblock
\APACaddressPublisher{}{Springer}.
\newblock
\begin{APACrefURL} \url{https://doi.org/10.1007/978-3-030-01228-1\_19} \end{APACrefURL}
\newblock
\begin{APACrefDOI} \doi{10.1007/978-3-030-01228-1\_19} \end{APACrefDOI}
\PrintBackRefs{\CurrentBib}

\bibitem [\protect \citeauthoryear {%
Shorten%
\ \BBA {} Khoshgoftaar%
}{%
Shorten%
\ \BBA {} Khoshgoftaar%
}{%
{\protect \APACyear {2019}}%
}]{%
DataAugSurvey}
\APACinsertmetastar {%
DataAugSurvey}%
\begin{APACrefauthors}%
Shorten, C.%
\BCBT {}\ \BBA {} Khoshgoftaar, T\BPBI M.%
\end{APACrefauthors}%
\unskip\
\newblock
\APACrefYearMonthDay{2019}{}{}.
\newblock
{\BBOQ}\APACrefatitle {{A survey on Image Data Augmentation for Deep Learning}} {{A survey on Image Data Augmentation for Deep Learning}}.{\BBCQ}
\newblock
\APACjournalVolNumPages{J. Big Data}{6}{}{60}.
\newblock
\begin{APACrefURL} \url{https://doi.org/10.1186/s40537-019-0197-0} \end{APACrefURL}
\newblock
\begin{APACrefDOI} \doi{10.1186/S40537-019-0197-0} \end{APACrefDOI}
\PrintBackRefs{\CurrentBib}

\bibitem [\protect \citeauthoryear {%
Sohn%
\ \protect \BOthers {.}}{%
Sohn%
\ \protect \BOthers {.}}{%
{\protect \APACyear {2020}}%
}]{%
FixMatch}
\APACinsertmetastar {%
FixMatch}%
\begin{APACrefauthors}%
Sohn, K.%
, Berthelot, D.%
, Carlini, N.%
, Zhang, Z.%
, Zhang, H.%
, Raffel, C.%
\BDBL {}Li, C.%
\end{APACrefauthors}%
\unskip\
\newblock
\APACrefYearMonthDay{2020}{}{}.
\newblock
{\BBOQ}\APACrefatitle {{FixMatch: Simplifying Semi-Supervised Learning with Consistency and Confidence}} {{FixMatch: Simplifying Semi-Supervised Learning with Consistency and Confidence}}.{\BBCQ}
\newblock
\BIn{} H.~Larochelle, M.~Ranzato, R.~Hadsell, M.~Balcan\BCBL {}\ \BBA {} H.~Lin\ (\BEDS), \APACrefbtitle {Advances in Neural Information Processing Systems 33: Annual Conference on Neural Information Processing Systems 2020, NeurIPS 2020, December 6-12, 2020, virtual.} {Advances in neural information processing systems 33: Annual conference on neural information processing systems 2020, neurips 2020, december 6-12, 2020, virtual.}
\newblock
\begin{APACrefURL} \url{https://proceedings.neurips.cc/paper/2020/hash/06964dce9addb1c5cb5d6e3d9838f733-Abstract.html} \end{APACrefURL}
\PrintBackRefs{\CurrentBib}

\bibitem [\protect \citeauthoryear {%
Song%
, Kim%
, Park%
, Shin%
\BCBL {}\ \BBA {} Lee%
}{%
Song%
\ \protect \BOthers {.}}{%
{\protect \APACyear {2023}}%
}]{%
Noise:Surveys:Label-DNN}
\APACinsertmetastar {%
Noise:Surveys:Label-DNN}%
\begin{APACrefauthors}%
Song, H.%
, Kim, M.%
, Park, D.%
, Shin, Y.%
\BCBL {}\ \BBA {} Lee, J.%
\end{APACrefauthors}%
\unskip\
\newblock
\APACrefYearMonthDay{2023}{}{}.
\newblock
{\BBOQ}\APACrefatitle {{Learning From Noisy Labels With Deep Neural Networks: {A} Survey}} {{Learning From Noisy Labels With Deep Neural Networks: {A} Survey}}.{\BBCQ}
\newblock
\APACjournalVolNumPages{{IEEE} Trans. Neural Networks Learn. Syst.}{34}{11}{8135--8153}.
\newblock
\begin{APACrefURL} \url{https://doi.org/10.1109/TNNLS.2022.3152527} \end{APACrefURL}
\newblock
\begin{APACrefDOI} \doi{10.1109/TNNLS.2022.3152527} \end{APACrefDOI}
\PrintBackRefs{\CurrentBib}

\bibitem [\protect \citeauthoryear {%
Sun%
, Jin%
\BCBL {}\ \BBA {} Tu%
}{%
Sun%
\ \protect \BOthers {.}}{%
{\protect \APACyear {2011}}%
}]{%
viewconstruction}
\APACinsertmetastar {%
viewconstruction}%
\begin{APACrefauthors}%
Sun, S.%
, Jin, F.%
\BCBL {}\ \BBA {} Tu, W.%
\end{APACrefauthors}%
\unskip\
\newblock
\APACrefYearMonthDay{2011}{}{}.
\newblock
{\BBOQ}\APACrefatitle {{View Construction for Multi-view Semi-supervised Learning}} {{View Construction for Multi-view Semi-supervised Learning}}.{\BBCQ}
\newblock
\BIn{} D.~Liu, H.~Zhang, M\BPBI M.~Polycarpou, C.~Alippi\BCBL {}\ \BBA {} H.~He\ (\BEDS), \APACrefbtitle {Advances in Neural Networks - {ISNN} 2011 - 8th International Symposium on Neural Networks, {ISNN} 2011, Guilin, China, May 29-June 1, 2011, Proceedings, Part {I}} {Advances in neural networks - {ISNN} 2011 - 8th international symposium on neural networks, {ISNN} 2011, guilin, china, may 29-june 1, 2011, proceedings, part {I}}\ (\BVOL\ 6675, \BPGS\ 595--601).
\newblock
\APACaddressPublisher{}{Springer}.
\newblock
\begin{APACrefURL} \url{https://doi.org/10.1007/978-3-642-21105-8\_69} \end{APACrefURL}
\newblock
\begin{APACrefDOI} \doi{10.1007/978-3-642-21105-8\_69} \end{APACrefDOI}
\PrintBackRefs{\CurrentBib}

\bibitem [\protect \citeauthoryear {%
Tarvainen%
\ \BBA {} Valpola%
}{%
Tarvainen%
\ \BBA {} Valpola%
}{%
{\protect \APACyear {2017}}%
}]{%
MeanTeachers}
\APACinsertmetastar {%
MeanTeachers}%
\begin{APACrefauthors}%
Tarvainen, A.%
\BCBT {}\ \BBA {} Valpola, H.%
\end{APACrefauthors}%
\unskip\
\newblock
\APACrefYearMonthDay{2017}{}{}.
\newblock
{\BBOQ}\APACrefatitle {{Mean teachers are better role models: Weight-averaged consistency targets improve semi-supervised deep learning results}} {{Mean teachers are better role models: Weight-averaged consistency targets improve semi-supervised deep learning results}}.{\BBCQ}
\newblock
\BIn{} I.~Guyon\ \BOthers {.}\ (\BEDS), \APACrefbtitle {Advances in Neural Information Processing Systems 30: Annual Conference on Neural Information Processing Systems 2017, December 4-9, 2017, Long Beach, CA, {USA}} {Advances in neural information processing systems 30: Annual conference on neural information processing systems 2017, december 4-9, 2017, long beach, ca, {USA}}\ (\BPGS\ 1195--1204).
\newblock
\begin{APACrefURL} \url{https://proceedings.neurips.cc/paper/2017/hash/68053af2923e00204c3ca7c6a3150cf7-Abstract.html} \end{APACrefURL}
\PrintBackRefs{\CurrentBib}

\bibitem [\protect \citeauthoryear {%
{Tong Xiao}%
, {Tian Xia}%
, {Yi Yang}%
, {Chang Huang}%
\BCBL {}\ \BBA {} {Xiaogang Wang}%
}{%
{Tong Xiao}%
\ \protect \BOthers {.}}{%
{\protect \APACyear {2015}}%
}]{%
DLfromNL}
\APACinsertmetastar {%
DLfromNL}%
\begin{APACrefauthors}%
{Tong Xiao}%
, {Tian Xia}%
, {Yi Yang}%
, {Chang Huang}%
\BCBL {}\ \BBA {} {Xiaogang Wang}.%
\end{APACrefauthors}%
\unskip\
\newblock
\APACrefYearMonthDay{2015}{{\APACmonth{06}}}{}.
\newblock
{\BBOQ}\APACrefatitle {{Learning from Massive Noisy Labeled Data for Image Classification}} {{Learning from Massive Noisy Labeled Data for Image Classification}}.{\BBCQ}
\newblock
\BIn{} \APACrefbtitle {2015 {{IEEE Conference}} on {{Computer Vision}} and {{Pattern Recognition}} ({{CVPR}})} {2015 {{IEEE Conference}} on {{Computer Vision}} and {{Pattern Recognition}} ({{CVPR}})}\ (\BPGS\ 2691--2699).
\newblock
\APACaddressPublisher{Boston, MA, USA}{IEEE}.
\newblock
\begin{APACrefDOI} \doi{10.1109/CVPR.2015.7298885} \end{APACrefDOI}
\PrintBackRefs{\CurrentBib}

\bibitem [\protect \citeauthoryear {%
van Engelen%
\ \BBA {} Hoos%
}{%
van Engelen%
\ \BBA {} Hoos%
}{%
{\protect \APACyear {2020}}%
}]{%
van2020survey}
\APACinsertmetastar {%
van2020survey}%
\begin{APACrefauthors}%
van Engelen, J\BPBI E.%
\BCBT {}\ \BBA {} Hoos, H\BPBI H.%
\end{APACrefauthors}%
\unskip\
\newblock
\APACrefYearMonthDay{2020}{}{}.
\newblock
{\BBOQ}\APACrefatitle {{A survey on semi-supervised learning}} {{A survey on semi-supervised learning}}.{\BBCQ}
\newblock
\APACjournalVolNumPages{Mach. Learn.}{109}{2}{373--440}.
\newblock
\begin{APACrefURL} \url{https://doi.org/10.1007/s10994-019-05855-6} \end{APACrefURL}
\newblock
\begin{APACrefDOI} \doi{10.1007/S10994-019-05855-6} \end{APACrefDOI}
\PrintBackRefs{\CurrentBib}

\bibitem [\protect \citeauthoryear {%
Verma%
\ \protect \BOthers {.}}{%
Verma%
\ \protect \BOthers {.}}{%
{\protect \APACyear {2022}}%
}]{%
ICT}
\APACinsertmetastar {%
ICT}%
\begin{APACrefauthors}%
Verma, V.%
, Kawaguchi, K.%
, Lamb, A.%
, Kannala, J.%
, Solin, A.%
, Bengio, Y.%
\BCBL {}\ \BBA {} Lopez-Paz, D.%
\end{APACrefauthors}%
\unskip\
\newblock
\APACrefYearMonthDay{2022}{{\APACmonth{01}}}{}.
\newblock
{\BBOQ}\APACrefatitle {Interpolation consistency training for semi-supervised learning} {Interpolation consistency training for semi-supervised learning}.{\BBCQ}
\newblock
\APACjournalVolNumPages{Neural Networks}{145}{}{90–106}.
\newblock
\begin{APACrefURL} \url{http://dx.doi.org/10.1016/j.neunet.2021.10.008} \end{APACrefURL}
\newblock
\begin{APACrefDOI} \doi{10.1016/j.neunet.2021.10.008} \end{APACrefDOI}
\PrintBackRefs{\CurrentBib}

\bibitem [\protect \citeauthoryear {%
Wallin%
, Svensson%
, Kahl%
\BCBL {}\ \BBA {} Hammarstrand%
}{%
Wallin%
\ \protect \BOthers {.}}{%
{\protect \APACyear {2023}}%
}]{%
SeFOSS}
\APACinsertmetastar {%
SeFOSS}%
\begin{APACrefauthors}%
Wallin, E.%
, Svensson, L.%
, Kahl, F.%
\BCBL {}\ \BBA {} Hammarstrand, L.%
\end{APACrefauthors}%
\unskip\
\newblock
\APACrefYearMonthDay{2023}{}{}.
\newblock
\APACrefbtitle {Improving Open-Set Semi-Supervised Learning with Self-Supervision.} {Improving open-set semi-supervised learning with self-supervision.}
\newblock
\begin{APACrefURL} \url{https://arxiv.org/abs/2301.10127} \end{APACrefURL}
\PrintBackRefs{\CurrentBib}

\bibitem [\protect \citeauthoryear {%
Wallin%
, Svensson%
, Kahl%
\BCBL {}\ \BBA {} Hammarstrand%
}{%
Wallin%
\ \protect \BOthers {.}}{%
{\protect \APACyear {2024}}%
}]{%
ProSub}
\APACinsertmetastar {%
ProSub}%
\begin{APACrefauthors}%
Wallin, E.%
, Svensson, L.%
, Kahl, F.%
\BCBL {}\ \BBA {} Hammarstrand, L.%
\end{APACrefauthors}%
\unskip\
\newblock
\APACrefYearMonthDay{2024}{}{}.
\newblock
\APACrefbtitle {ProSub: Probabilistic Open-Set Semi-Supervised Learning with Subspace-Based Out-of-Distribution Detection.} {Prosub: Probabilistic open-set semi-supervised learning with subspace-based out-of-distribution detection.}
\newblock
\begin{APACrefURL} \url{https://arxiv.org/abs/2407.11735} \end{APACrefURL}
\PrintBackRefs{\CurrentBib}

\bibitem [\protect \citeauthoryear {%
J.~Wang%
, Luo%
\BCBL {}\ \BBA {} Zeng%
}{%
J.~Wang%
\ \protect \BOthers {.}}{%
{\protect \APACyear {2008}}%
}]{%
RASCO}
\APACinsertmetastar {%
RASCO}%
\begin{APACrefauthors}%
Wang, J.%
, Luo, S.%
\BCBL {}\ \BBA {} Zeng, X.%
\end{APACrefauthors}%
\unskip\
\newblock
\APACrefYearMonthDay{2008}{}{}.
\newblock
{\BBOQ}\APACrefatitle {{A random subspace method for co-training}} {{A random subspace method for co-training}}.{\BBCQ}
\newblock
\BIn{} \APACrefbtitle {Proceedings of the International Joint Conference on Neural Networks, {IJCNN} 2008, part of the {IEEE} World Congress on Computational Intelligence, {WCCI} 2008, Hong Kong, China, June 1-6, 2008} {Proceedings of the international joint conference on neural networks, {IJCNN} 2008, part of the {IEEE} world congress on computational intelligence, {WCCI} 2008, hong kong, china, june 1-6, 2008}\ (\BPGS\ 195--200).
\newblock
\APACaddressPublisher{}{{IEEE}}.
\newblock
\begin{APACrefURL} \url{https://doi.org/10.1109/IJCNN.2008.4633789} \end{APACrefURL}
\newblock
\begin{APACrefDOI} \doi{10.1109/IJCNN.2008.4633789} \end{APACrefDOI}
\PrintBackRefs{\CurrentBib}

\bibitem [\protect \citeauthoryear {%
Q.~Wang%
\ \protect \BOthers {.}}{%
Q.~Wang%
\ \protect \BOthers {.}}{%
{\protect \APACyear {2021}}%
}]{%
Noise:Label:InstanceDependent:UPM}
\APACinsertmetastar {%
Noise:Label:InstanceDependent:UPM}%
\begin{APACrefauthors}%
Wang, Q.%
, Han, B.%
, Liu, T.%
, Niu, G.%
, Yang, J.%
\BCBL {}\ \BBA {} Gong, C.%
\end{APACrefauthors}%
\unskip\
\newblock
\APACrefYearMonthDay{2021}{}{}.
\newblock
{\BBOQ}\APACrefatitle {{Tackling Instance-Dependent Label Noise via a Universal Probabilistic Model}} {{Tackling Instance-Dependent Label Noise via a Universal Probabilistic Model}}.{\BBCQ}
\newblock
\BIn{} \APACrefbtitle {{Thirty-Fifth {AAAI} Conference on Artificial Intelligence, {AAAI} 2021, Thirty-Third Conference on Innovative Applications of Artificial Intelligence, {IAAI} 2021, The Eleventh Symposium on Educational Advances in Artificial Intelligence, {EAAI} 2021, Virtual Event, February 2-9, 2021}} {{Thirty-Fifth {AAAI} Conference on Artificial Intelligence, {AAAI} 2021, Thirty-Third Conference on Innovative Applications of Artificial Intelligence, {IAAI} 2021, The Eleventh Symposium on Educational Advances in Artificial Intelligence, {EAAI} 2021, Virtual Event, February 2-9, 2021}}\ (\BPGS\ 10183--10191).
\newblock
\APACaddressPublisher{}{{AAAI} Press}.
\newblock
\begin{APACrefURL} \url{https://doi.org/10.1609/aaai.v35i11.17221} \end{APACrefURL}
\newblock
\begin{APACrefDOI} \doi{10.1609/AAAI.V35I11.17221} \end{APACrefDOI}
\PrintBackRefs{\CurrentBib}

\bibitem [\protect \citeauthoryear {%
Q.~Wang%
\ \protect \BOthers {.}}{%
Q.~Wang%
\ \protect \BOthers {.}}{%
{\protect \APACyear {2022}}%
}]{%
InstanceDependentNoise}
\APACinsertmetastar {%
InstanceDependentNoise}%
\begin{APACrefauthors}%
Wang, Q.%
, Han, B.%
, Liu, T.%
, Niu, G.%
, Yang, J.%
\BCBL {}\ \BBA {} Gong, C.%
\end{APACrefauthors}%
\unskip\
\newblock
\APACrefYearMonthDay{2022}{{\APACmonth{03}}}{}.
\newblock
\APACrefbtitle {Tackling {{Instance-Dependent Label Noise}} via a {{Universal Probabilistic Model}}} {Tackling {{Instance-Dependent Label Noise}} via a {{Universal Probabilistic Model}}}\ (\BNUM\ arXiv:2101.05467).
\newblock
\APACaddressPublisher{}{arXiv}.
\newblock
\begin{APACrefDOI} \doi{10.48550/arXiv.2101.05467} \end{APACrefDOI}
\PrintBackRefs{\CurrentBib}

\bibitem [\protect \citeauthoryear {%
W.~Wang%
\ \BBA {} Zhou%
}{%
W.~Wang%
\ \BBA {} Zhou%
}{%
{\protect \APACyear {2013}}%
}]{%
insufficientviews}
\APACinsertmetastar {%
insufficientviews}%
\begin{APACrefauthors}%
Wang, W.%
\BCBT {}\ \BBA {} Zhou, Z.%
\end{APACrefauthors}%
\unskip\
\newblock
\APACrefYearMonthDay{2013}{}{}.
\newblock
{\BBOQ}\APACrefatitle {{Co-Training with Insufficient Views}} {{Co-Training with Insufficient Views}}.{\BBCQ}
\newblock
\BIn{} C\BPBI S.~Ong\ \BBA {} T\BPBI B.~Ho\ (\BEDS), \APACrefbtitle {Asian Conference on Machine Learning, {ACML} 2013, Canberra, ACT, Australia, November 13-15, 2013} {Asian conference on machine learning, {ACML} 2013, canberra, act, australia, november 13-15, 2013}\ (\BVOL~29, \BPGS\ 467--482).
\newblock
\APACaddressPublisher{}{JMLR.org}.
\newblock
\begin{APACrefURL} \url{http://proceedings.mlr.press/v29/Wang13b.html} \end{APACrefURL}
\PrintBackRefs{\CurrentBib}

\bibitem [\protect \citeauthoryear {%
Y.~Wang%
, Guo%
, Song%
\BCBL {}\ \BBA {} Huang%
}{%
Y.~Wang%
\ \protect \BOthers {.}}{%
{\protect \APACyear {2020}}%
}]{%
Meta-Semi}
\APACinsertmetastar {%
Meta-Semi}%
\begin{APACrefauthors}%
Wang, Y.%
, Guo, J.%
, Song, S.%
\BCBL {}\ \BBA {} Huang, G.%
\end{APACrefauthors}%
\unskip\
\newblock
\APACrefYearMonthDay{2020}{}{}.
\newblock
{\BBOQ}\APACrefatitle {Meta-Semi: {A} Meta-learning Approach for Semi-supervised Learning} {Meta-semi: {A} meta-learning approach for semi-supervised learning}.{\BBCQ}
\newblock
\APACjournalVolNumPages{CoRR}{abs/2007.02394}{}{}.
\newblock
\begin{APACrefURL} \url{https://arxiv.org/abs/2007.02394} \end{APACrefURL}
\PrintBackRefs{\CurrentBib}

\bibitem [\protect \citeauthoryear {%
Wen%
, Lai%
\BCBL {}\ \BBA {} Qian%
}{%
Wen%
\ \protect \BOthers {.}}{%
{\protect \APACyear {2021}}%
}]{%
PreparingKD}
\APACinsertmetastar {%
PreparingKD}%
\begin{APACrefauthors}%
Wen, T.%
, Lai, S.%
\BCBL {}\ \BBA {} Qian, X.%
\end{APACrefauthors}%
\unskip\
\newblock
\APACrefYearMonthDay{2021}{}{}.
\newblock
{\BBOQ}\APACrefatitle {{Preparing lessons: Improve knowledge distillation with better supervision}} {{Preparing lessons: Improve knowledge distillation with better supervision}}.{\BBCQ}
\newblock
\APACjournalVolNumPages{Neurocomputing}{454}{}{25--33}.
\newblock
\begin{APACrefURL} \url{https://doi.org/10.1016/j.neucom.2021.04.102} \end{APACrefURL}
\newblock
\begin{APACrefDOI} \doi{10.1016/J.NEUCOM.2021.04.102} \end{APACrefDOI}
\PrintBackRefs{\CurrentBib}

\bibitem [\protect \citeauthoryear {%
Xia%
\ \protect \BOthers {.}}{%
Xia%
\ \protect \BOthers {.}}{%
{\protect \APACyear {2019}}%
}]{%
Noise:Label:New:Anchors}
\APACinsertmetastar {%
Noise:Label:New:Anchors}%
\begin{APACrefauthors}%
Xia, X.%
, Liu, T.%
, Wang, N.%
, Han, B.%
, Gong, C.%
, Niu, G.%
\BCBL {}\ \BBA {} Sugiyama, M.%
\end{APACrefauthors}%
\unskip\
\newblock
\APACrefYearMonthDay{2019}{}{}.
\newblock
{\BBOQ}\APACrefatitle {{Are Anchor Points Really Indispensable in Label-Noise Learning?}} {{Are Anchor Points Really Indispensable in Label-Noise Learning?}}{\BBCQ}
\newblock
\BIn{} H\BPBI M.~Wallach, H.~Larochelle, A.~Beygelzimer, F.~d'Alch{\'{e}}{-}Buc, E\BPBI B.~Fox\BCBL {}\ \BBA {} R.~Garnett\ (\BEDS), \APACrefbtitle {{Advances in Neural Information Processing Systems 32: Annual Conference on Neural Information Processing Systems 2019, NeurIPS 2019, December 8-14, 2019, Vancouver, BC, Canada}} {{Advances in Neural Information Processing Systems 32: Annual Conference on Neural Information Processing Systems 2019, NeurIPS 2019, December 8-14, 2019, Vancouver, BC, Canada}}\ (\BPGS\ 6835--6846).
\newblock
\begin{APACrefURL} \url{https://proceedings.neurips.cc/paper/2019/hash/9308b0d6e5898366a4a986bc33f3d3e7-Abstract.html} \end{APACrefURL}
\PrintBackRefs{\CurrentBib}

\bibitem [\protect \citeauthoryear {%
Xiao%
\ \protect \BOthers {.}}{%
Xiao%
\ \protect \BOthers {.}}{%
{\protect \APACyear {2023}}%
}]{%
HALOC}
\APACinsertmetastar {%
HALOC}%
\begin{APACrefauthors}%
Xiao, J.%
, Zhang, C.%
, Gong, Y.%
, Yin, M.%
, Sui, Y.%
, Xiang, L.%
\BDBL {}Yuan, B.%
\end{APACrefauthors}%
\unskip\
\newblock
\APACrefYearMonthDay{2023}{}{}.
\newblock
{\BBOQ}\APACrefatitle {{{HALOC:} Hardware-Aware Automatic Low-Rank Compression for Compact Neural Networks}} {{{HALOC:} Hardware-Aware Automatic Low-Rank Compression for Compact Neural Networks}}.{\BBCQ}
\newblock
\BIn{} B.~Williams, Y.~Chen\BCBL {}\ \BBA {} J.~Neville\ (\BEDS), \APACrefbtitle {Thirty-Seventh {AAAI} Conference on Artificial Intelligence, {AAAI} 2023, Thirty-Fifth Conference on Innovative Applications of Artificial Intelligence, {IAAI} 2023, Thirteenth Symposium on Educational Advances in Artificial Intelligence, {EAAI} 2023, Washington, DC, USA, February 7-14, 2023} {Thirty-seventh {AAAI} conference on artificial intelligence, {AAAI} 2023, thirty-fifth conference on innovative applications of artificial intelligence, {IAAI} 2023, thirteenth symposium on educational advances in artificial intelligence, {EAAI} 2023, washington, dc, usa, february 7-14, 2023}\ (\BPGS\ 10464--10472).
\newblock
\APACaddressPublisher{}{{AAAI} Press}.
\newblock
\begin{APACrefURL} \url{https://doi.org/10.1609/aaai.v37i9.26244} \end{APACrefURL}
\newblock
\begin{APACrefDOI} \doi{10.1609/AAAI.V37I9.26244} \end{APACrefDOI}
\PrintBackRefs{\CurrentBib}

\bibitem [\protect \citeauthoryear {%
Xie%
, Dai%
, Hovy%
, Luong%
\BCBL {}\ \BBA {} Le%
}{%
Xie%
\ \protect \BOthers {.}}{%
{\protect \APACyear {2019}}%
}]{%
UDA}
\APACinsertmetastar {%
UDA}%
\begin{APACrefauthors}%
Xie, Q.%
, Dai, Z.%
, Hovy, E\BPBI H.%
, Luong, M.%
\BCBL {}\ \BBA {} Le, Q\BPBI V.%
\end{APACrefauthors}%
\unskip\
\newblock
\APACrefYearMonthDay{2019}{}{}.
\newblock
{\BBOQ}\APACrefatitle {{Unsupervised Data Augmentation}} {{Unsupervised Data Augmentation}}.{\BBCQ}
\newblock
\APACjournalVolNumPages{CoRR}{abs/1904.12848}{}{}.
\newblock
\begin{APACrefURL} \url{http://arxiv.org/abs/1904.12848} \end{APACrefURL}
\PrintBackRefs{\CurrentBib}

\bibitem [\protect \citeauthoryear {%
Xie%
, Luong%
, Hovy%
\BCBL {}\ \BBA {} Le%
}{%
Xie%
\ \protect \BOthers {.}}{%
{\protect \APACyear {2020}}%
}]{%
NST}
\APACinsertmetastar {%
NST}%
\begin{APACrefauthors}%
Xie, Q.%
, Luong, M.%
, Hovy, E\BPBI H.%
\BCBL {}\ \BBA {} Le, Q\BPBI V.%
\end{APACrefauthors}%
\unskip\
\newblock
\APACrefYearMonthDay{2020}{}{}.
\newblock
{\BBOQ}\APACrefatitle {{Self-Training With Noisy Student Improves ImageNet Classification}} {{Self-Training With Noisy Student Improves ImageNet Classification}}.{\BBCQ}
\newblock
\BIn{} \APACrefbtitle {2020 {IEEE/CVF} Conference on Computer Vision and Pattern Recognition, {CVPR} 2020, Seattle, WA, USA, June 13-19, 2020} {2020 {IEEE/CVF} conference on computer vision and pattern recognition, {CVPR} 2020, seattle, wa, usa, june 13-19, 2020}\ (\BPGS\ 10684--10695).
\newblock
\APACaddressPublisher{}{Computer Vision Foundation / {IEEE}}.
\newblock
\begin{APACrefURL} \url{https://openaccess.thecvf.com/content\_CVPR\_2020/html/Xie\_Self-Training\_With\_Noisy\_Student\_Improves\_ImageNet\_Classification\_CVPR\_2020\_paper.html} \end{APACrefURL}
\newblock
\begin{APACrefDOI} \doi{10.1109/CVPR42600.2020.01070} \end{APACrefDOI}
\PrintBackRefs{\CurrentBib}

\bibitem [\protect \citeauthoryear {%
Xu%
\ \protect \BOthers {.}}{%
Xu%
\ \protect \BOthers {.}}{%
{\protect \APACyear {2023}}%
}]{%
demystifyingCLIP}
\APACinsertmetastar {%
demystifyingCLIP}%
\begin{APACrefauthors}%
Xu, H.%
, Xie, S.%
, Tan, X\BPBI E.%
, Huang, P.%
, Howes, R.%
, Sharma, V.%
\BDBL {}Feichtenhofer, C.%
\end{APACrefauthors}%
\unskip\
\newblock
\APACrefYearMonthDay{2023}{}{}.
\newblock
{\BBOQ}\APACrefatitle {{Demystifying {CLIP} Data}} {{Demystifying {CLIP} Data}}.{\BBCQ}
\newblock
\APACjournalVolNumPages{CoRR}{abs/2309.16671}{}{}.
\newblock
\begin{APACrefURL} \url{https://doi.org/10.48550/arXiv.2309.16671} \end{APACrefURL}
\newblock
\begin{APACrefDOI} \doi{10.48550/ARXIV.2309.16671} \end{APACrefDOI}
\PrintBackRefs{\CurrentBib}

\bibitem [\protect \citeauthoryear {%
Yan%
, Rosales%
, Fung%
, Ramanathan%
\BCBL {}\ \BBA {} Dy%
}{%
Yan%
\ \protect \BOthers {.}}{%
{\protect \APACyear {2014}}%
}]{%
Noise:Label:New:Annotators}
\APACinsertmetastar {%
Noise:Label:New:Annotators}%
\begin{APACrefauthors}%
Yan, Y.%
, Rosales, R.%
, Fung, G.%
, Ramanathan, S.%
\BCBL {}\ \BBA {} Dy, J\BPBI G.%
\end{APACrefauthors}%
\unskip\
\newblock
\APACrefYearMonthDay{2014}{}{}.
\newblock
{\BBOQ}\APACrefatitle {{Learning from multiple annotators with varying expertise}} {{Learning from multiple annotators with varying expertise}}.{\BBCQ}
\newblock
\APACjournalVolNumPages{Mach. Learn.}{95}{3}{291--327}.
\newblock
\begin{APACrefURL} \url{https://doi.org/10.1007/s10994-013-5412-1} \end{APACrefURL}
\newblock
\begin{APACrefDOI} \doi{10.1007/S10994-013-5412-1} \end{APACrefDOI}
\PrintBackRefs{\CurrentBib}

\bibitem [\protect \citeauthoryear {%
Yan%
, Xu%
, Tsang%
, Long%
\BCBL {}\ \BBA {} Yang%
}{%
Yan%
\ \protect \BOthers {.}}{%
{\protect \APACyear {2016}}%
}]{%
ROSSEL}
\APACinsertmetastar {%
ROSSEL}%
\begin{APACrefauthors}%
Yan, Y.%
, Xu, Z.%
, Tsang, I\BPBI W.%
, Long, G.%
\BCBL {}\ \BBA {} Yang, Y.%
\end{APACrefauthors}%
\unskip\
\newblock
\APACrefYearMonthDay{2016}{}{}.
\newblock
{\BBOQ}\APACrefatitle {{Robust Semi-Supervised Learning through Label Aggregation}} {{Robust Semi-Supervised Learning through Label Aggregation}}.{\BBCQ}
\newblock
\BIn{} D.~Schuurmans\ \BBA {} M\BPBI P.~Wellman\ (\BEDS), \APACrefbtitle {Proceedings of the Thirtieth {AAAI} Conference on Artificial Intelligence, February 12-17, 2016, Phoenix, Arizona, {USA}} {Proceedings of the thirtieth {AAAI} conference on artificial intelligence, february 12-17, 2016, phoenix, arizona, {USA}}\ (\BPGS\ 2244--2250).
\newblock
\APACaddressPublisher{}{{AAAI} Press}.
\newblock
\begin{APACrefURL} \url{https://doi.org/10.1609/aaai.v30i1.10276} \end{APACrefURL}
\newblock
\begin{APACrefDOI} \doi{10.1609/AAAI.V30I1.10276} \end{APACrefDOI}
\PrintBackRefs{\CurrentBib}

\bibitem [\protect \citeauthoryear {%
Yang%
, Song%
, King%
\BCBL {}\ \BBA {} Xu%
}{%
Yang%
\ \protect \BOthers {.}}{%
{\protect \APACyear {2023}}%
}]{%
ASurveyDeepSemi}
\APACinsertmetastar {%
ASurveyDeepSemi}%
\begin{APACrefauthors}%
Yang, X.%
, Song, Z.%
, King, I.%
\BCBL {}\ \BBA {} Xu, Z.%
\end{APACrefauthors}%
\unskip\
\newblock
\APACrefYearMonthDay{2023}{}{}.
\newblock
{\BBOQ}\APACrefatitle {{A Survey on Deep Semi-Supervised Learning}} {{A Survey on Deep Semi-Supervised Learning}}.{\BBCQ}
\newblock
\APACjournalVolNumPages{{IEEE} Trans. Knowl. Data Eng.}{35}{9}{8934--8954}.
\newblock
\begin{APACrefURL} \url{https://doi.org/10.1109/TKDE.2022.3220219} \end{APACrefURL}
\newblock
\begin{APACrefDOI} \doi{10.1109/TKDE.2022.3220219} \end{APACrefDOI}
\PrintBackRefs{\CurrentBib}

\bibitem [\protect \citeauthoryear {%
Yi%
, Wang%
\BCBL {}\ \BBA {} Wu%
}{%
Yi%
\ \protect \BOthers {.}}{%
{\protect \APACyear {2022}}%
}]{%
PENCIL}
\APACinsertmetastar {%
PENCIL}%
\begin{APACrefauthors}%
Yi, K.%
, Wang, G\BHBI H.%
\BCBL {}\ \BBA {} Wu, J.%
\end{APACrefauthors}%
\unskip\
\newblock
\APACrefYearMonthDay{2022}{{\APACmonth{02}}}{}.
\newblock
\APACrefbtitle {{{PENCIL}}: {{Deep Learning}} with {{Noisy Labels}}} {{{PENCIL}}: {{Deep Learning}} with {{Noisy Labels}}}\ (\BNUM\ arXiv:2202.08436).
\newblock
\APACaddressPublisher{}{arXiv}.
\newblock
\begin{APACrefDOI} \doi{10.48550/arXiv.2202.08436} \end{APACrefDOI}
\PrintBackRefs{\CurrentBib}

\bibitem [\protect \citeauthoryear {%
J.~Yu%
\ \protect \BOthers {.}}{%
J.~Yu%
\ \protect \BOthers {.}}{%
{\protect \APACyear {2021}}%
}]{%
BetterTriTraining}
\APACinsertmetastar {%
BetterTriTraining}%
\begin{APACrefauthors}%
Yu, J.%
, Yin, H.%
, Gao, M.%
, Xia, X.%
, Zhang, X.%
\BCBL {}\ \BBA {} Hung, N\BPBI Q\BPBI V.%
\end{APACrefauthors}%
\unskip\
\newblock
\APACrefYearMonthDay{2021}{}{}.
\newblock
{\BBOQ}\APACrefatitle {{Socially-Aware Self-Supervised Tri-Training for Recommendation}} {{Socially-Aware Self-Supervised Tri-Training for Recommendation}}.{\BBCQ}
\newblock
\BIn{} F.~Zhu, B\BPBI C.~Ooi\BCBL {}\ \BBA {} C.~Miao\ (\BEDS), \APACrefbtitle {{KDD} '21: The 27th {ACM} {SIGKDD} Conference on Knowledge Discovery and Data Mining, Virtual Event, Singapore, August 14-18, 2021} {{KDD} '21: The 27th {ACM} {SIGKDD} conference on knowledge discovery and data mining, virtual event, singapore, august 14-18, 2021}\ (\BPGS\ 2084--2092).
\newblock
\APACaddressPublisher{}{{ACM}}.
\newblock
\begin{APACrefURL} \url{https://doi.org/10.1145/3447548.3467340} \end{APACrefURL}
\newblock
\begin{APACrefDOI} \doi{10.1145/3447548.3467340} \end{APACrefDOI}
\PrintBackRefs{\CurrentBib}

\bibitem [\protect \citeauthoryear {%
X.~Yu%
, Liu%
, Wang%
\BCBL {}\ \BBA {} Tao%
}{%
X.~Yu%
\ \protect \BOthers {.}}{%
{\protect \APACyear {2017}}%
}]{%
lowrank}
\APACinsertmetastar {%
lowrank}%
\begin{APACrefauthors}%
Yu, X.%
, Liu, T.%
, Wang, X.%
\BCBL {}\ \BBA {} Tao, D.%
\end{APACrefauthors}%
\unskip\
\newblock
\APACrefYearMonthDay{2017}{}{}.
\newblock
{\BBOQ}\APACrefatitle {{On Compressing Deep Models by Low Rank and Sparse Decomposition}} {{On Compressing Deep Models by Low Rank and Sparse Decomposition}}.{\BBCQ}
\newblock
\BIn{} \APACrefbtitle {2017 {IEEE} Conference on Computer Vision and Pattern Recognition, {CVPR} 2017, Honolulu, HI, USA, July 21-26, 2017} {2017 {IEEE} conference on computer vision and pattern recognition, {CVPR} 2017, honolulu, hi, usa, july 21-26, 2017}\ (\BPGS\ 67--76).
\newblock
\APACaddressPublisher{}{{IEEE} Computer Society}.
\newblock
\begin{APACrefURL} \url{https://doi.org/10.1109/CVPR.2017.15} \end{APACrefURL}
\newblock
\begin{APACrefDOI} \doi{10.1109/CVPR.2017.15} \end{APACrefDOI}
\PrintBackRefs{\CurrentBib}

\bibitem [\protect \citeauthoryear {%
Yuan%
, Chen%
, Zhang%
, Tai%
\BCBL {}\ \BBA {} McMains%
}{%
Yuan%
\ \protect \BOthers {.}}{%
{\protect \APACyear {2018}}%
}]{%
ICL}
\APACinsertmetastar {%
ICL}%
\begin{APACrefauthors}%
Yuan, B.%
, Chen, J.%
, Zhang, W.%
, Tai, H.%
\BCBL {}\ \BBA {} McMains, S.%
\end{APACrefauthors}%
\unskip\
\newblock
\APACrefYearMonthDay{2018}{}{}.
\newblock
{\BBOQ}\APACrefatitle {{Iterative Cross Learning on Noisy Labels}} {{Iterative Cross Learning on Noisy Labels}}.{\BBCQ}
\newblock
\BIn{} \APACrefbtitle {2018 {IEEE} Winter Conference on Applications of Computer Vision, {WACV} 2018, Lake Tahoe, NV, USA, March 12-15, 2018} {2018 {IEEE} winter conference on applications of computer vision, {WACV} 2018, lake tahoe, nv, usa, march 12-15, 2018}\ (\BPGS\ 757--765).
\newblock
\APACaddressPublisher{}{{IEEE} Computer Society}.
\newblock
\begin{APACrefURL} \url{https://doi.org/10.1109/WACV.2018.00088} \end{APACrefURL}
\newblock
\begin{APACrefDOI} \doi{10.1109/WACV.2018.00088} \end{APACrefDOI}
\PrintBackRefs{\CurrentBib}

\bibitem [\protect \citeauthoryear {%
Zadeh%
}{%
Zadeh%
}{%
{\protect \APACyear {1965}}%
}]{%
Fuzzy:Zadeh}
\APACinsertmetastar {%
Fuzzy:Zadeh}%
\begin{APACrefauthors}%
Zadeh, L.%
\end{APACrefauthors}%
\unskip\
\newblock
\APACrefYearMonthDay{1965}{}{}.
\newblock
{\BBOQ}\APACrefatitle {{Fuzzy sets}} {{Fuzzy sets}}.{\BBCQ}
\newblock
\APACjournalVolNumPages{Information and Control}{8}{3}{338-353}.
\newblock
\begin{APACrefURL} \url{https://www.sciencedirect.com/science/article/pii/S001999586590241X} \end{APACrefURL}
\newblock
\begin{APACrefDOI} \doi{https://doi.org/10.1016/S0019-9958(65)90241-X} \end{APACrefDOI}
\PrintBackRefs{\CurrentBib}

\bibitem [\protect \citeauthoryear {%
Zhai%
, Mustafa%
, Kolesnikov%
\BCBL {}\ \BBA {} Beyer%
}{%
Zhai%
\ \protect \BOthers {.}}{%
{\protect \APACyear {2023}}%
}]{%
SigLIP}
\APACinsertmetastar {%
SigLIP}%
\begin{APACrefauthors}%
Zhai, X.%
, Mustafa, B.%
, Kolesnikov, A.%
\BCBL {}\ \BBA {} Beyer, L.%
\end{APACrefauthors}%
\unskip\
\newblock
\APACrefYearMonthDay{2023}{}{}.
\newblock
{\BBOQ}\APACrefatitle {{Sigmoid Loss for Language Image Pre-Training}} {{Sigmoid Loss for Language Image Pre-Training}}.{\BBCQ}
\newblock
\BIn{} \APACrefbtitle {{IEEE/CVF} International Conference on Computer Vision, {ICCV} 2023, Paris, France, October 1-6, 2023} {{IEEE/CVF} international conference on computer vision, {ICCV} 2023, paris, france, october 1-6, 2023}\ (\BPGS\ 11941--11952).
\newblock
\APACaddressPublisher{}{{IEEE}}.
\newblock
\begin{APACrefURL} \url{https://doi.org/10.1109/ICCV51070.2023.01100} \end{APACrefURL}
\newblock
\begin{APACrefDOI} \doi{10.1109/ICCV51070.2023.01100} \end{APACrefDOI}
\PrintBackRefs{\CurrentBib}

\bibitem [\protect \citeauthoryear {%
B.~Zhang%
\ \protect \BOthers {.}}{%
B.~Zhang%
\ \protect \BOthers {.}}{%
{\protect \APACyear {2021}}%
}]{%
FlexMatch}
\APACinsertmetastar {%
FlexMatch}%
\begin{APACrefauthors}%
Zhang, B.%
, Wang, Y.%
, Hou, W.%
, Wu, H.%
, Wang, J.%
, Okumura, M.%
\BCBL {}\ \BBA {} Shinozaki, T.%
\end{APACrefauthors}%
\unskip\
\newblock
\APACrefYearMonthDay{2021}{}{}.
\newblock
{\BBOQ}\APACrefatitle {{FlexMatch: Boosting Semi-Supervised Learning with Curriculum Pseudo Labeling}} {{FlexMatch: Boosting Semi-Supervised Learning with Curriculum Pseudo Labeling}}.{\BBCQ}
\newblock
\BIn{} M.~Ranzato, A.~Beygelzimer, Y\BPBI N.~Dauphin, P.~Liang\BCBL {}\ \BBA {} J\BPBI W.~Vaughan\ (\BEDS), \APACrefbtitle {Advances in Neural Information Processing Systems 34: Annual Conference on Neural Information Processing Systems 2021, NeurIPS 2021, December 6-14, 2021, virtual} {Advances in neural information processing systems 34: Annual conference on neural information processing systems 2021, neurips 2021, december 6-14, 2021, virtual}\ (\BPGS\ 18408--18419).
\newblock
\begin{APACrefURL} \url{https://proceedings.neurips.cc/paper/2021/hash/995693c15f439e3d189b06e89d145dd5-Abstract.html} \end{APACrefURL}
\PrintBackRefs{\CurrentBib}

\bibitem [\protect \citeauthoryear {%
H.~Zhang%
, Ciss{\'{e}}%
, Dauphin%
\BCBL {}\ \BBA {} Lopez{-}Paz%
}{%
H.~Zhang%
\ \protect \BOthers {.}}{%
{\protect \APACyear {2018}}%
}]{%
MixUp}
\APACinsertmetastar {%
MixUp}%
\begin{APACrefauthors}%
Zhang, H.%
, Ciss{\'{e}}, M.%
, Dauphin, Y\BPBI N.%
\BCBL {}\ \BBA {} Lopez{-}Paz, D.%
\end{APACrefauthors}%
\unskip\
\newblock
\APACrefYearMonthDay{2018}{}{}.
\newblock
{\BBOQ}\APACrefatitle {{mixup: Beyond Empirical Risk Minimization}} {{mixup: Beyond Empirical Risk Minimization}}.{\BBCQ}
\newblock
\BIn{} \APACrefbtitle {6th International Conference on Learning Representations, {ICLR} 2018, Vancouver, BC, Canada, April 30 - May 3, 2018, Conference Track Proceedings.} {6th international conference on learning representations, {ICLR} 2018, vancouver, bc, canada, april 30 - may 3, 2018, conference track proceedings.}
\newblock
\APACaddressPublisher{}{OpenReview.net}.
\newblock
\begin{APACrefURL} \url{https://openreview.net/forum?id=r1Ddp1-Rb} \end{APACrefURL}
\PrintBackRefs{\CurrentBib}

\bibitem [\protect \citeauthoryear {%
L.~Zhang%
\ \protect \BOthers {.}}{%
L.~Zhang%
\ \protect \BOthers {.}}{%
{\protect \APACyear {2019}}%
}]{%
KD-self-distill}
\APACinsertmetastar {%
KD-self-distill}%
\begin{APACrefauthors}%
Zhang, L.%
, Song, J.%
, Gao, A.%
, Chen, J.%
, Bao, C.%
\BCBL {}\ \BBA {} Ma, K.%
\end{APACrefauthors}%
\unskip\
\newblock
\APACrefYearMonthDay{2019}{}{}.
\newblock
{\BBOQ}\APACrefatitle {{Be Your Own Teacher: Improve the Performance of Convolutional Neural Networks via Self Distillation}} {{Be Your Own Teacher: Improve the Performance of Convolutional Neural Networks via Self Distillation}}.{\BBCQ}
\newblock
\BIn{} \APACrefbtitle {2019 {IEEE/CVF} International Conference on Computer Vision, {ICCV} 2019, Seoul, Korea (South), October 27 - November 2, 2019} {2019 {IEEE/CVF} international conference on computer vision, {ICCV} 2019, seoul, korea (south), october 27 - november 2, 2019}\ (\BPGS\ 3712--3721).
\newblock
\APACaddressPublisher{}{{IEEE}}.
\newblock
\begin{APACrefURL} \url{https://doi.org/10.1109/ICCV.2019.00381} \end{APACrefURL}
\newblock
\begin{APACrefDOI} \doi{10.1109/ICCV.2019.00381} \end{APACrefDOI}
\PrintBackRefs{\CurrentBib}

\bibitem [\protect \citeauthoryear {%
T.~Zhou%
, Wang%
\BCBL {}\ \BBA {} Bilmes%
}{%
T.~Zhou%
\ \protect \BOthers {.}}{%
{\protect \APACyear {2020}}%
}]{%
TCSSL}
\APACinsertmetastar {%
TCSSL}%
\begin{APACrefauthors}%
Zhou, T.%
, Wang, S.%
\BCBL {}\ \BBA {} Bilmes, J\BPBI A.%
\end{APACrefauthors}%
\unskip\
\newblock
\APACrefYearMonthDay{2020}{}{}.
\newblock
{\BBOQ}\APACrefatitle {{Time-Consistent Self-Supervision for Semi-Supervised Learning}} {{Time-Consistent Self-Supervision for Semi-Supervised Learning}}.{\BBCQ}
\newblock
\BIn{} \APACrefbtitle {Proceedings of the 37th International Conference on Machine Learning, {ICML} 2020, 13-18 July 2020, Virtual Event} {Proceedings of the 37th international conference on machine learning, {ICML} 2020, 13-18 july 2020, virtual event}\ (\BVOL~119, \BPGS\ 11523--11533).
\newblock
\APACaddressPublisher{}{{PMLR}}.
\newblock
\begin{APACrefURL} \url{http://proceedings.mlr.press/v119/zhou20d.html} \end{APACrefURL}
\PrintBackRefs{\CurrentBib}

\bibitem [\protect \citeauthoryear {%
Y.~Zhou%
\ \BBA {} Goldman%
}{%
Y.~Zhou%
\ \BBA {} Goldman%
}{%
{\protect \APACyear {2004}}%
}]{%
democraticcolearning}
\APACinsertmetastar {%
democraticcolearning}%
\begin{APACrefauthors}%
Zhou, Y.%
\BCBT {}\ \BBA {} Goldman, S\BPBI A.%
\end{APACrefauthors}%
\unskip\
\newblock
\APACrefYearMonthDay{2004}{}{}.
\newblock
{\BBOQ}\APACrefatitle {{Democratic Co-Learning}} {{Democratic Co-Learning}}.{\BBCQ}
\newblock
\BIn{} \APACrefbtitle {16th {IEEE} International Conference on Tools with Artificial Intelligence {(ICTAI} 2004), 15-17 November 2004, Boca Raton, FL, {USA}} {16th {IEEE} international conference on tools with artificial intelligence {(ICTAI} 2004), 15-17 november 2004, boca raton, fl, {USA}}\ (\BPGS\ 594--602).
\newblock
\APACaddressPublisher{}{{IEEE} Computer Society}.
\newblock
\begin{APACrefURL} \url{https://doi.org/10.1109/ICTAI.2004.48} \end{APACrefURL}
\newblock
\begin{APACrefDOI} \doi{10.1109/ICTAI.2004.48} \end{APACrefDOI}
\PrintBackRefs{\CurrentBib}

\bibitem [\protect \citeauthoryear {%
Z.~Zhou%
\ \BBA {} Li%
}{%
Z.~Zhou%
\ \BBA {} Li%
}{%
{\protect \APACyear {2005}}%
}]{%
TriTraining}
\APACinsertmetastar {%
TriTraining}%
\begin{APACrefauthors}%
Zhou, Z.%
\BCBT {}\ \BBA {} Li, M.%
\end{APACrefauthors}%
\unskip\
\newblock
\APACrefYearMonthDay{2005}{}{}.
\newblock
{\BBOQ}\APACrefatitle {{Tri-Training: Exploiting Unlabeled Data Using Three Classifiers}} {{Tri-Training: Exploiting Unlabeled Data Using Three Classifiers}}.{\BBCQ}
\newblock
\APACjournalVolNumPages{{IEEE} Trans. Knowl. Data Eng.}{17}{11}{1529--1541}.
\newblock
\begin{APACrefURL} \url{https://doi.org/10.1109/TKDE.2005.186} \end{APACrefURL}
\newblock
\begin{APACrefDOI} \doi{10.1109/TKDE.2005.186} \end{APACrefDOI}
\PrintBackRefs{\CurrentBib}

\bibitem [\protect \citeauthoryear {%
W.~Zhu%
, Liu%
, Fernandez{-}Granda%
\BCBL {}\ \BBA {} Razavian%
}{%
W.~Zhu%
\ \protect \BOthers {.}}{%
{\protect \APACyear {2023}}%
}]{%
FairSSL}
\APACinsertmetastar {%
FairSSL}%
\begin{APACrefauthors}%
Zhu, W.%
, Liu, S.%
, Fernandez{-}Granda, C.%
\BCBL {}\ \BBA {} Razavian, N.%
\end{APACrefauthors}%
\unskip\
\newblock
\APACrefYearMonthDay{2023}{}{}.
\newblock
{\BBOQ}\APACrefatitle {{Making Self-supervised Learning Robust to Spurious Correlation via Learning-speed Aware Sampling}} {{Making Self-supervised Learning Robust to Spurious Correlation via Learning-speed Aware Sampling}}.{\BBCQ}
\newblock
\APACjournalVolNumPages{CoRR}{abs/2311.16361}{}{}.
\newblock
\begin{APACrefURL} \url{https://doi.org/10.48550/arXiv.2311.16361} \end{APACrefURL}
\newblock
\begin{APACrefDOI} \doi{10.48550/ARXIV.2311.16361} \end{APACrefDOI}
\PrintBackRefs{\CurrentBib}

\bibitem [\protect \citeauthoryear {%
X.~Zhu%
\ \BBA {} Ghahramani%
}{%
X.~Zhu%
\ \BBA {} Ghahramani%
}{%
{\protect \APACyear {2002}}%
}]{%
LabelPropagation}
\APACinsertmetastar {%
LabelPropagation}%
\begin{APACrefauthors}%
Zhu, X.%
\BCBT {}\ \BBA {} Ghahramani, Z.%
\end{APACrefauthors}%
\unskip\
\newblock
\APACrefYearMonthDay{2002}{}{}.
\newblock
{\BBOQ}\APACrefatitle {{Learning from Labeled and Unlabeled Data with Label Propagation}} {{Learning from Labeled and Unlabeled Data with Label Propagation}}.{\BBCQ}
\newblock
\APACjournalVolNumPages{ProQuest Number: INFORMATION TO ALL USERS}{}{}{}.
\PrintBackRefs{\CurrentBib}

\bibitem [\protect \citeauthoryear {%
X.~Zhu%
\ \BBA {} Goldberg%
}{%
X.~Zhu%
\ \BBA {} Goldberg%
}{%
{\protect \APACyear {2009}}%
}]{%
zhu_introduction_2009}
\APACinsertmetastar {%
zhu_introduction_2009}%
\begin{APACrefauthors}%
Zhu, X.%
\BCBT {}\ \BBA {} Goldberg, A\BPBI B.%
\end{APACrefauthors}%
\unskip\
\newblock
\APACrefYear{2009}.
\newblock
\APACrefbtitle {{Introduction to Semi-Supervised Learning}} {{Introduction to Semi-Supervised Learning}}.
\newblock
\APACaddressPublisher{}{Morgan {\&} Claypool Publishers}.
\newblock
\begin{APACrefURL} \url{https://doi.org/10.2200/S00196ED1V01Y200906AIM006} \end{APACrefURL}
\newblock
\begin{APACrefDOI} \doi{10.2200/S00196ED1V01Y200906AIM006} \end{APACrefDOI}
\PrintBackRefs{\CurrentBib}

\end{thebibliography}

\end{document}